\AtBeginDocument{%
	\paperwidth=\dimexpr
	1in + \oddsidemargin
	+ \textwidth
	+ 1in + \oddsidemargin
	\relax
	\paperheight=\dimexpr
	1in + \topmargin
	+ \headheight + \headsep
	+ \textheight
	+ 1in + \topmargin
	\relax
	\usepackage[pass]{geometry}\relax
}

\documentclass[smallextended]{svjour3}
\DeclareMathAlphabet{\mathbbcal}{OMS}{cmsy}{m}{n}
\usepackage{amsmath}
\usepackage{times}
\usepackage{newtxtext, newtxmath}
\usepackage{dutchcal}

\usepackage{multicol}        %
\usepackage[bottom]{footmisc}%
\usepackage{bm}
\usepackage{overpic}
\usepackage{algorithm}
\usepackage{algpseudocode}
\usepackage{color}
\usepackage{enumitem}   
\usepackage{mathtools}
\usepackage{multirow}
\usepackage{booktabs}%
\usepackage{tikz-cd}
\usepackage[colorlinks=true,allcolors=blue]{hyperref}
\usepackage{anyfontsize}
\usepackage{graphbox}
\usepackage[export]{adjustbox}
\usepackage[retainorgcmds]{IEEEtrantools}
\usepackage{optidef}

\newcommand{\R}{\mathbb R}
\renewcommand{\S}{\mathbb S}

\newcommand{\Prob}{\mathbb P}

\def\E{{\mathbb E}}
\newcommand{\Cov}{\mathbb{C}\textmd{ov}}

\newcommand{\norm}[1]{\left\lVert#1\right\rVert}

\ifx \undefined \varvec    \def \varvec    #1{\text{\boldmath$#1$}} \fi
\journalname{Mathematical Geoscience}
\usepackage[authoryear]{natbib}
\bibpunct{(}{)}{}{a}{}{;}

\begin{document}
	
\title{Dual Random Fields and their Application to Mineral Potential Mapping}

\titlerunning{Dual Random Fields and their Application in Mineral Potential Mapping}        %

\author{\'Alvaro I. Riquelme \textsuperscript{1,2}}

\authorrunning{\'Alvaro I. Riquelme} %

\institute{\'Alvaro I. Riquelme \at
\textsuperscript{1}  CSIRO, Canberra, ACT, Australia \at
\textsuperscript{2}  Mineral Exploration Cooperative Research Centre (MinEx CRC), Kensington, WA, Australia \at
\email{riq004@csiro.au}         %
}

\maketitle

\begin{abstract}
In various geosciences branches, including mineral exploration, geometallurgical characterization on established mining operations, and remote sensing, the regionalized input variables are spatially well-sampled across the domain of interest, limiting the scope of spatial uncertainty quantification procedures. In turn, response outcomes such as the mineral potential in a given region, mining throughput, metallurgical recovery, or in-situ estimations from remote satellite imagery, are usually modeled from a much-restricted subset of testing samples, collected at certain locations due to accessibility restrictions and the high acquisition costs. 
Our limited understanding of these functions, in terms of the multi-dimensional complexity of causalities and unnoticed dependencies on inaccessible inputs, may lead to observing changes in such functions based on their geographical location. Pooling together different response functions across the domain is critical to correctly predict outcome responses, the uncertainty associated with these inferred values, and the significance of inputs in such predictions at unexplored areas. This paper introduces the notion of a dual random field (dRF), where the response function itself is considered a regionalized variable. In this way, different established response models across the geographic domain can be considered as observations of a dRF realization, enabling the spatial inference and uncertainty assessment of both response models and their predictions. We explain how dRFs inherit all the properties from classical random fields, allowing the use of standard Gaussian simulation procedures to simulate them. Additionally, we illustrate the application of dRFs in a mineral potential mapping case study in which different local binary response models are calibrated on the domain by using support vector classification. These models are combined to obtain a mineral potential response, providing an example of how to rigorously integrate machine learning approaches with geostatistics.

\end{abstract}

\section{Introduction}
\label{sec:intro}

The drop in grades of current mining projects, the gradual decline in the discovery rate of new deposits (\citealt{blain2000fifty}; \citealt{ali2017mineral}; \citealt{castillo2021defining}) and the growing global demand for different minerals (\citealt{elshkaki2018resource}; \citealt{watari2021major}; \citealt{singer2023long}) puts the mining industry under increasing pressure for addressing the shortfall in resources. To meet global demand, mining-related stakeholders rely on expanding current projects when economic conditions and diverse mining constraints (geotechnical and environmental) allow it, but fore-mostly on discovering new resources.

Mineral exploration is among the activities facing the problem of shortage in resources, with not only private entities playing an active role but also government agencies through the collection of initial (pre-competitive) geological, geochemical, and geophysical data on strategic geographical zones, with the purpose of triggering future private exploration initiatives (\citealt{witt2013regional}; \citealt{riganti2015125}; \citealt{ford2019translating}). To highlight areas of interest for prospectivity and mitigate the high risk, intrinsic to the exploration activity \citep{singer1996application}, mineral potential mapping (MPM) methods have been increasingly established since the mid-1960s and early 1970s (\citealt{harris1965application}; \citealt{agterberg1974automatic}) as a set of analytical tools adapted for narrowing down the exploration search space. 

A typical MPM method aims to fit either a real-valued function, $ \mathcal{m}: \R^p  \to \R$ or a categorical function (usually binary) $ \mathcal{m}: \R^p  \to \{0,1\}$, taking a set of $p$ input variables, denoted by the $p$-vector $ {\textbf{\textit{z}}}= [{z}_1,{z}_2,\dots,{z}_i,\dots, {z}_p]^T $ with entries indexed by $ i \in \{1,\dots,p\} $, representing different proxy features related to a given mineral system under study, and mapping them into either a real scalar response variable $ y $, usually interpreted as a potential, probability or any transitive estimate (grade or tonnage), or rather a binary response, indicating the presence of mineral, respectively. Such fitting relies on the availability of sampling data over the geographical domain under analysis. Hereafter, the terms ``response function'' and ``response model'' will be used interchangeably to denote this function. If we consider a training set of $ N $ observations $ \{\textbf{\textit{z}}_\alpha, {\textit{y}}_\alpha\}^N_{\alpha=1}  \in \R^p\times \{0,1\} $, with  $ {{{\textbf{\textit{z}}}}}_\alpha = [{z}_{1\alpha},\dots,{z}_{p\alpha}]^T $, then $ \mathcal{m} $ is required to satisfy $ {\textit{y}}_\alpha = \mathcal{m}(\textbf{\textit{z}}_\alpha) $ incurring in the less possible misfitting, for all $ \alpha \in \{1,\dots,N\} $.

Under this setting, many statistical- and machine-learning methods have been employed for MPM purposes. Among the main ones, we find: weights of evidence (\citealt{good1950probability}; \citealt{agterberg1990statistical}; \citealt{bonham1990application}; \citealt{schaeben2016quest}; \citealt{baddeley2021optimal}); logistic  regression (\citealt{agterberg1974automatic};  \citealt{chung1980regression};  \citealt{carranza2001logistic}; 
\citealt{oh2008regional}; \citealt{zhang2018improved};  \citealt{lin2020mineral}); decision trees (\citealt{chen2014method} \citealt{rodriguez2015machine}) and random forest (\citealt{harris2015data};  \citealt{carranza2015data}; \citealt{porwal2015introduction}; \citealt{carranza2016data}; \citealt{rodriguez2015machine}; \citealt{ford2020practical}); support vector machine (\citealt{zuo2011support}; \citealt{abedi2012support}; \citealt{granek2015data}; \citealt{chen2017application}); neural networks  (\citealt{singer1996application}; \citealt{brown2003use}; \citealt{Rigol-Sanchez2003}; \citealt{brown2003artificial}; \citealt{porwal2003artificial}; \citealt{corsini2009weight}; \citealt{wang2010probabilistic}; \citealt{rodriguez2015machine}); and deep learning techniques (\citealt{zuo2019deep}; \citealt{luo2020recognition}; \citealt{li2020applications}; \citealt{sun2020data}; \citealt{zhang2021detection}; \citealt{yin2022mineral}; \citealt{xiong2022physically}; \citealt{wang2022lithological}). An explanatory review of these methods and the key aspects considered when applied to MPM problems can be found in \citealt{zuo2020geodata} and \citealt{liu2022developments}.

Although the features $ {z}_1,{z}_2,\dots, {z}_p $ can be considered as regionalized variables defined on the geographical domain $D$ under study, ${z}_i=\{{z}_i(\mathbf{u}): \mathbf{u} \in D \subseteq \R^n\}$, most previously mentioned statistical- and machine-learning methods omit this fact, missing (\textit{i}) the spatial dependencies between predictor variables themselves \citep{schaeben2014targeting} and (\textit{ii}) the spatial relation between inputs and outcomes. Geostatistical-aided MPM methodologies (\citeauthor{wang2020monte} \citeyear{wang2020monte}; \citeauthor{talebi2022truly} \citeyear{talebi2022truly}; \citeauthor{yang2023quantification} \citeyear{yang2023quantification}; \citeauthor{sadeghi2023decision} \citeyear{sadeghi2023decision}) exploit such relationships by stochastically filling in input features at points with only mineralization data, thereby updating global response models with newly augmented data. Geostatistical-aided techniques are especially beneficial for sparse spatial data, enabling us to measure the uncertainty of responses and to identify those zones with persistent high mineral potential across multiple valid statistical scenarios. 

In regions with high exploration maturity, however, input features have a dense coverage. As MPM workflows typically involve working with features acquired near the ground level (hereafter, we set $n=2$), inputs like geological maps and geophysical data might cover extensive areas, including those beyond known mineralized zones. As the interest in unlocking new resources intensifies with time, surface data acquisition not only expands existing datasets but also increases their spatial density, limiting the scope of geostatistical uncertainty quantification procedures previously discussed. 

A major challenge for MPM methods is the limited availability of mineralized zones, often spatially grouped towards well-established exploration regions. The uneven spatial disposal of mineralized locations introduces biases in potential maps, which tend to overestimate the mineral potential value at known areas and underestimate it at the undiscovered resources (\citeauthor{coolbaugh2007assessment} \citeyear{coolbaugh2007assessment};  \citeauthor{PORWAL2015839} \citeyear{PORWAL2015839};  \citeauthor{HRONSKY2019647} \citeyear{HRONSKY2019647};   \citeauthor{YOUSEFI2021106839} \citeyear{YOUSEFI2021106839}). Although this bias is commonly attributed to weighting schemes that favor input layers showing signatures related to mineralized locations, it is also important to consider that the use of one single, globally calibrated response model on the domain also has a decisive impact.

\begin{figure}[htbp]
\centering
\includegraphics[width=1\textwidth,trim={0 0cm 0cm 0cm},clip]{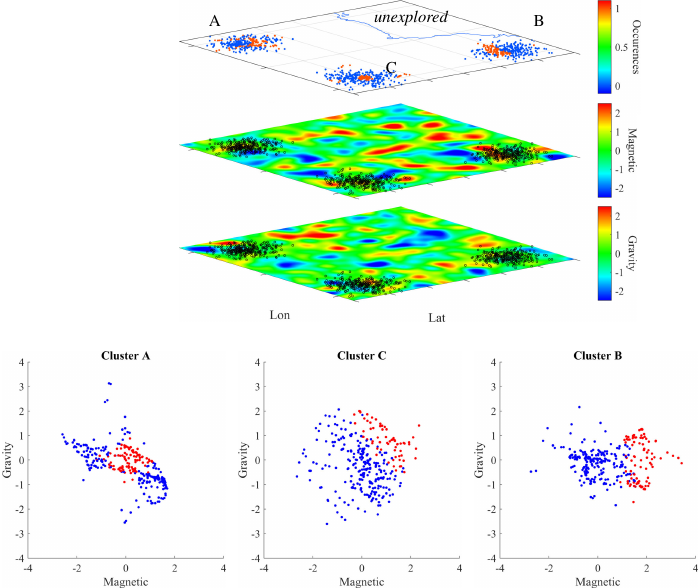}
\caption{General setting in mineral prospectivity mapping illustration. Relations among input features (gravity and magnetic intensity) and mineralized response (colored bullet in red and blue) may change at different locations. Simple local relations between inputs and the response will be overlooked due to the overlapping of datasets if a global model is considered in the domain. How do we extract the information from calibrated models at A, B, and C in a coherent fashion in order to make inferences at the unexplored location?}
\label{expprob}
\end{figure}

Causality in MPM is inherently linked to spatial factors, and what might explain the process of mineral formation in one area could vary significantly and not necessarily translate across large enough regions. Therefore, using a uniform model with unchanging parameters across an entire region may prove to be overly rigid when there is an incomplete comprehension of causality factors. Since MPM models typically apply a general calibration to the geographic area without adjusting parameters for spatial variations, even minor differences in causal factors can result in the failure to establish a universal rule on the domain (\citealt{fouedjio2020exact}). Figure \ref{expprob} illustrates the general MPM setting and the possible geographical change in a binary response model when considering two input features.

\begin{figure}[htbp]
\raggedleft
\includegraphics[width=0.3\textwidth,trim={0 0cm 0cm 0cm},clip]{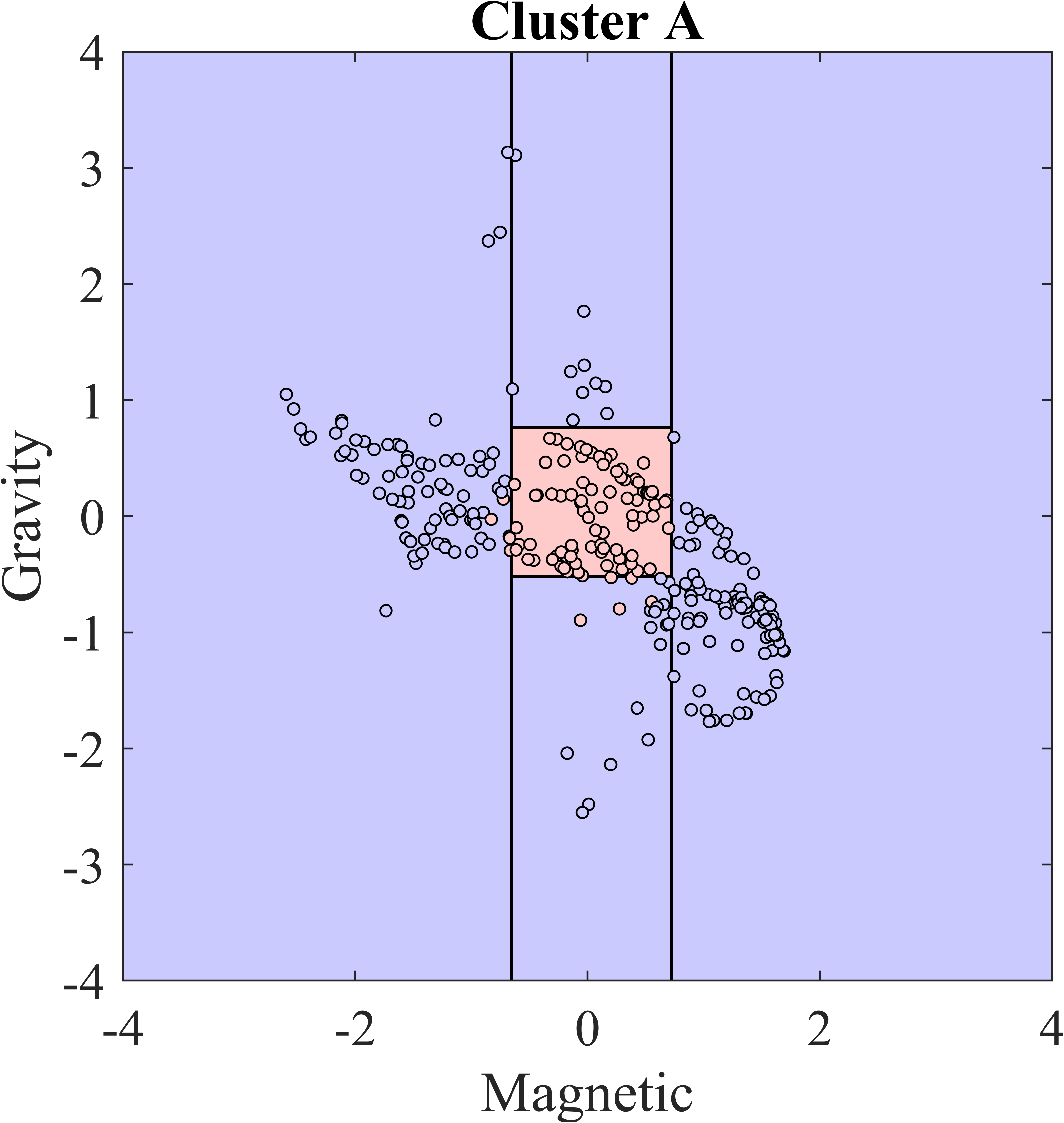}
\includegraphics[width=0.6\textwidth,trim={1cm 0.5cm 1cm 0.5cm},clip]{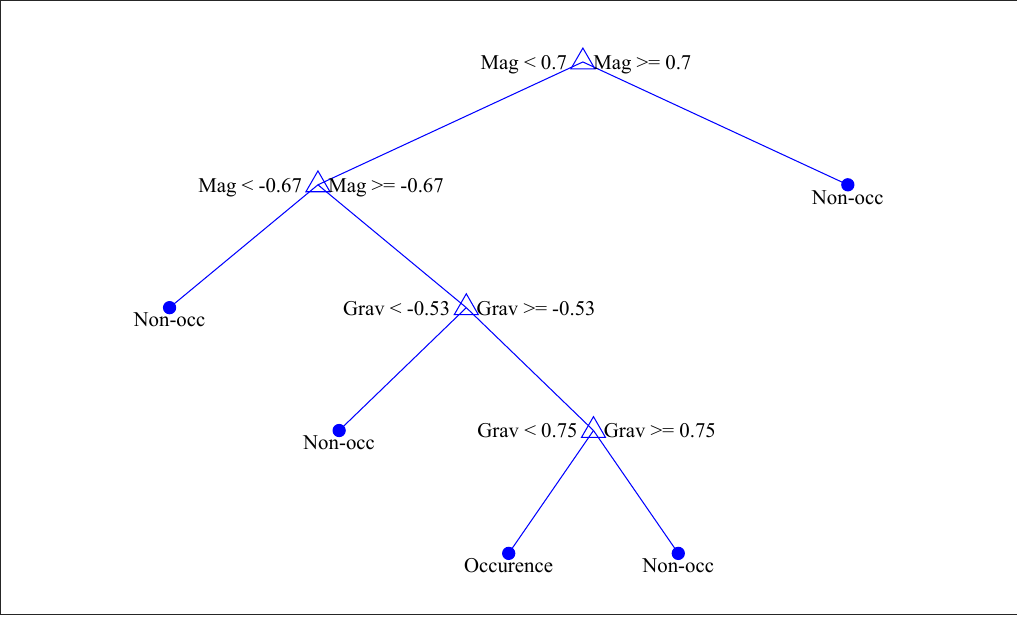}\\
\raggedleft
\includegraphics[width=0.3\textwidth,trim={0 0cm 0cm 0cm},clip]{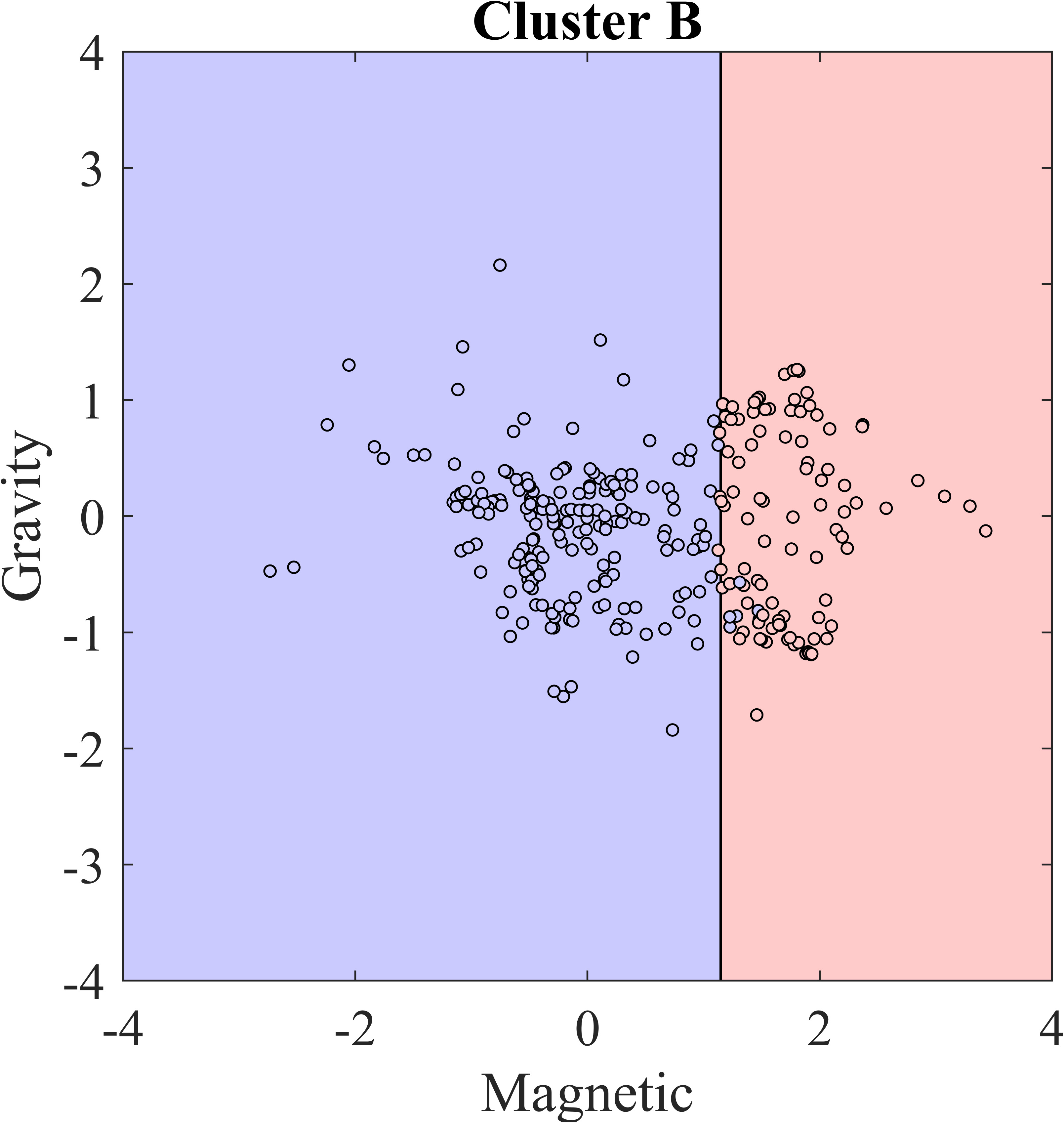}
\includegraphics[width=0.6\textwidth,trim={1cm 0.5cm 1cm 1cm},clip]{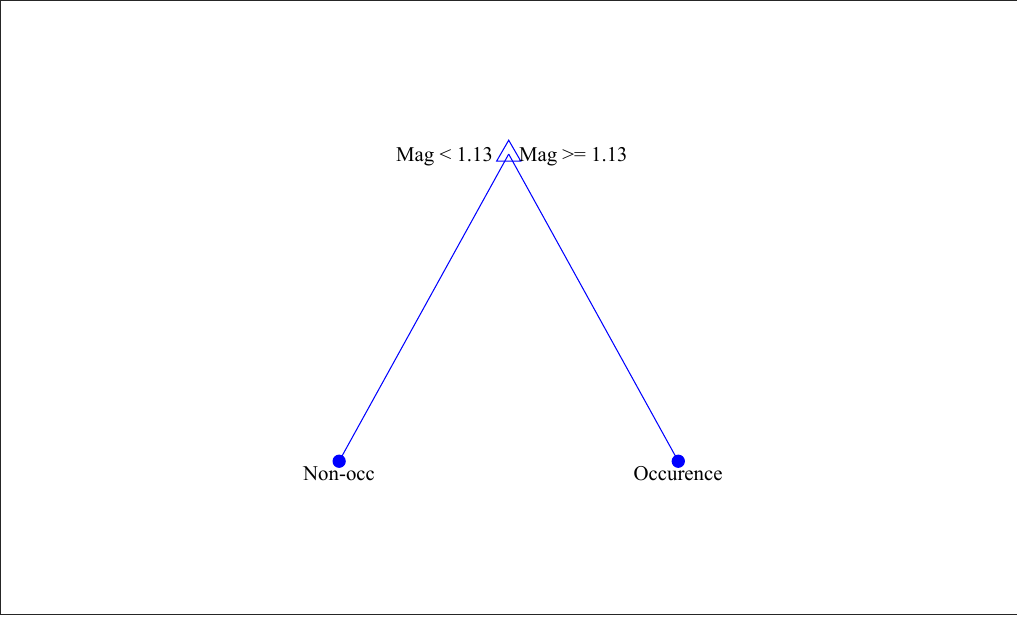}\\
\raggedleft
\includegraphics[width=0.3\textwidth,trim={0 0cm 0cm 0cm},clip]{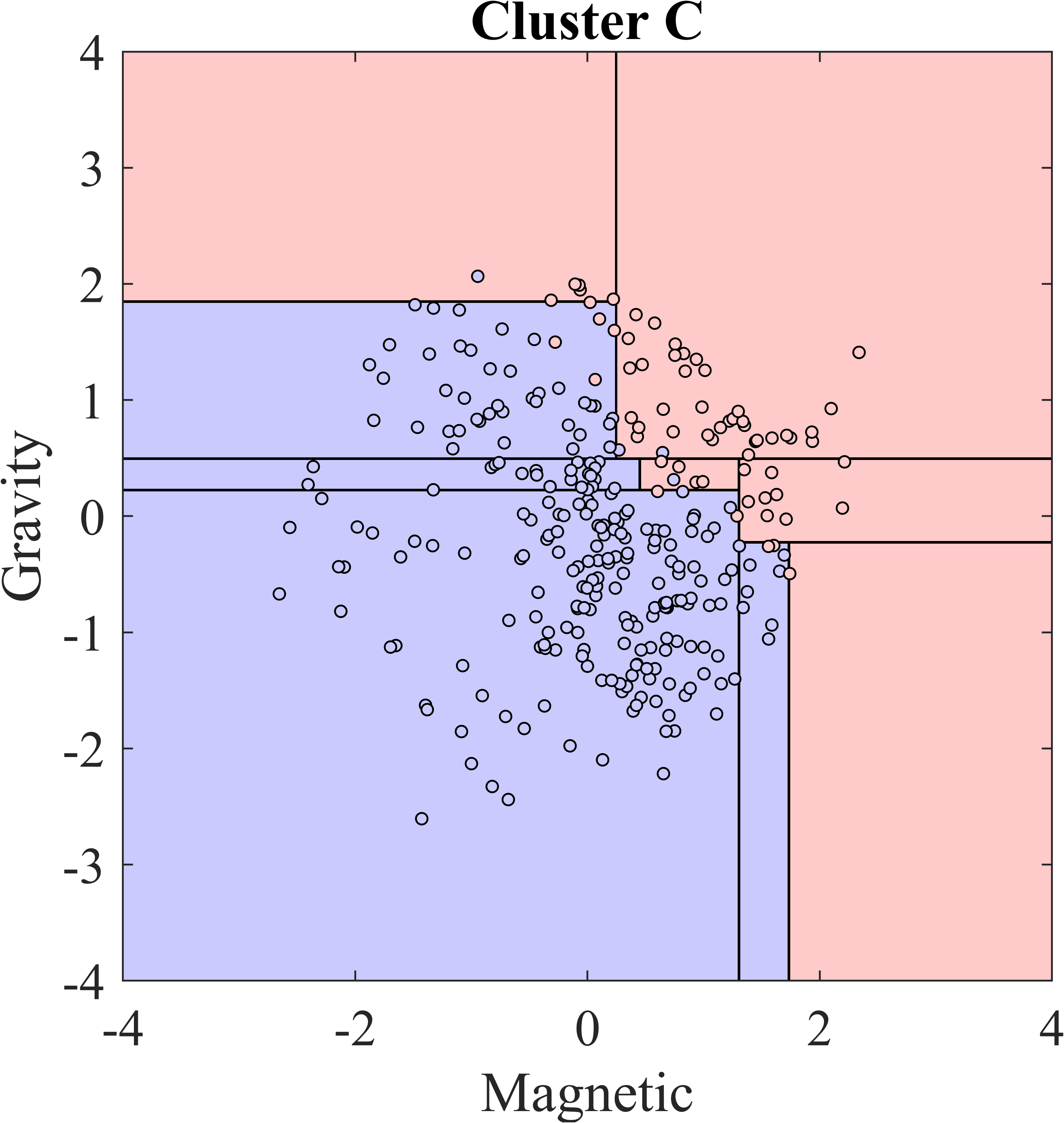}
\includegraphics[width=0.6\textwidth,trim={1cm 0.5cm 1cm 0.1cm},clip]{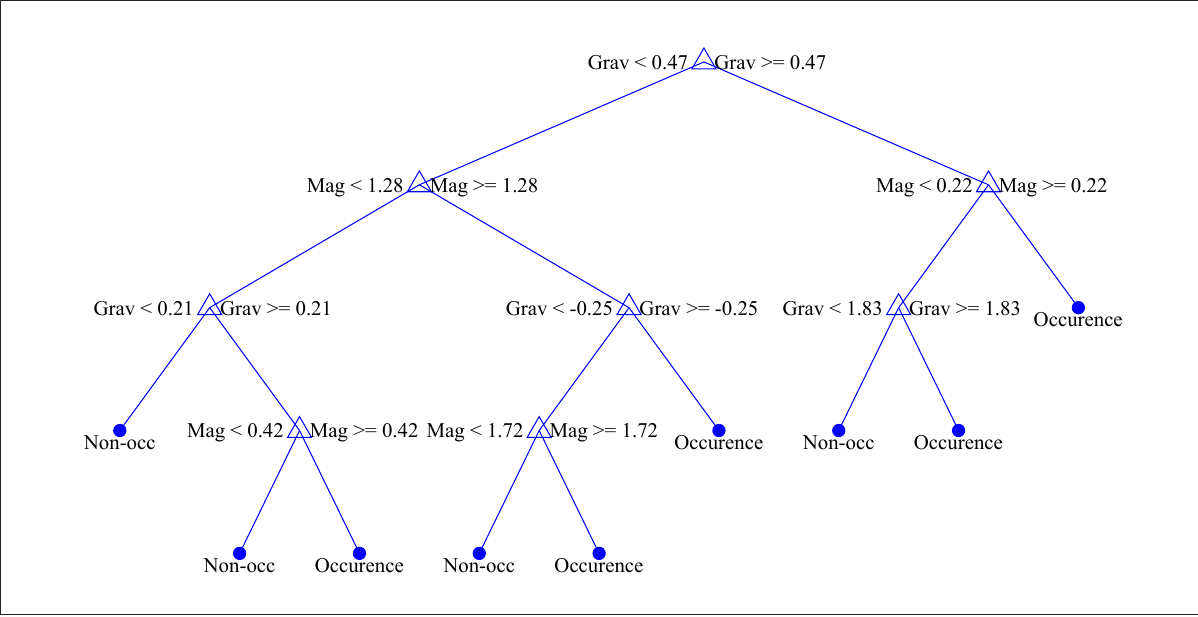}
	\caption{Decision tree modeling of mineralization at A, B, and C. Although each model is highly interpretative by itself, the graph-structure representation of the response model conditions the spatial problem in the sense of allowing the inference of a best-estimated model at unexplored locations.}
	\label{tree}
\end{figure}

\begin{figure}[htbp]
	\centering
	\includegraphics[width=0.74\textwidth,trim={0 0cm 0.0cm 0cm},clip]{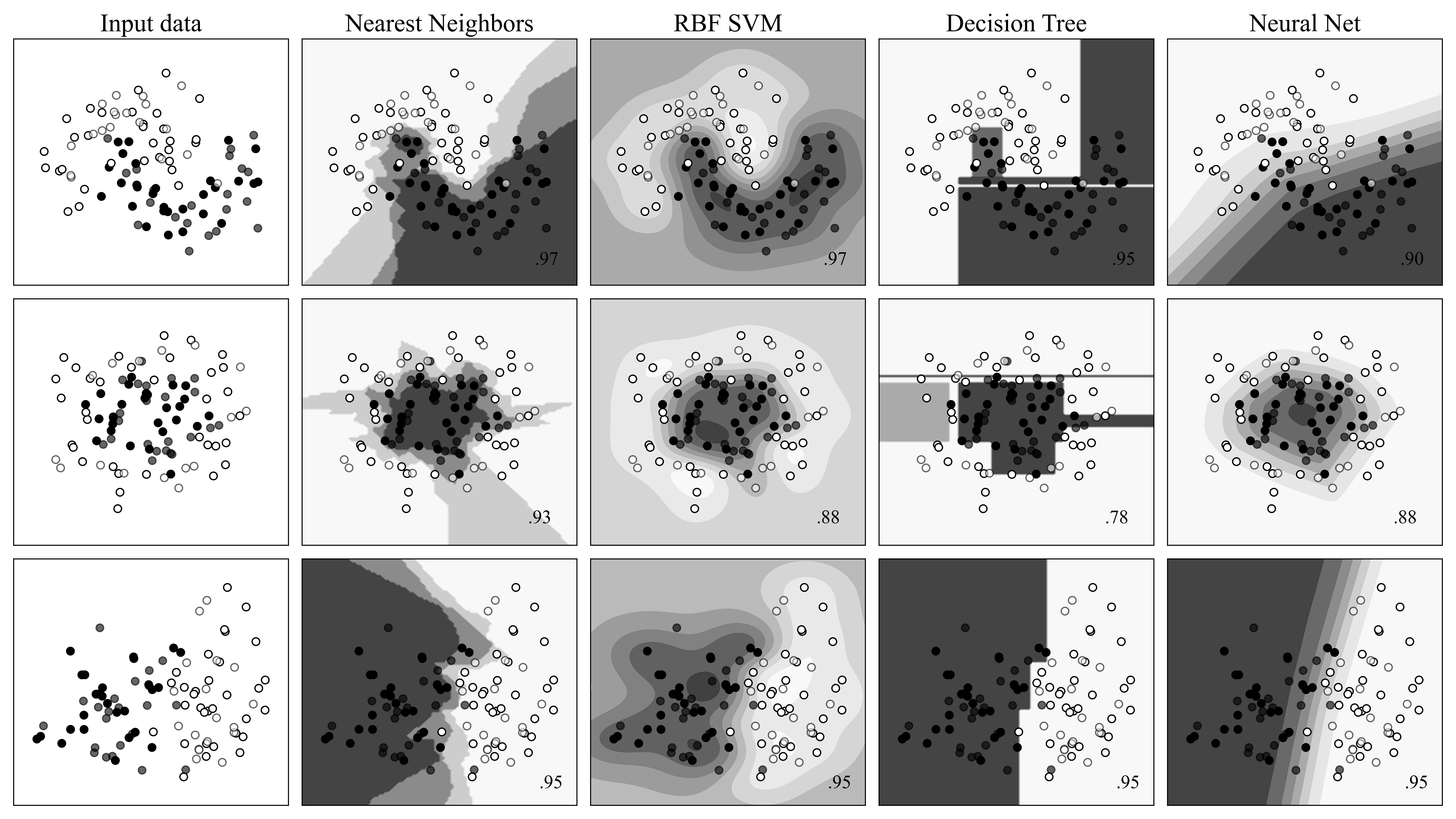}
	\includegraphics[width=0.25\textwidth,trim={0 0cm 0.0cm 0cm},clip]{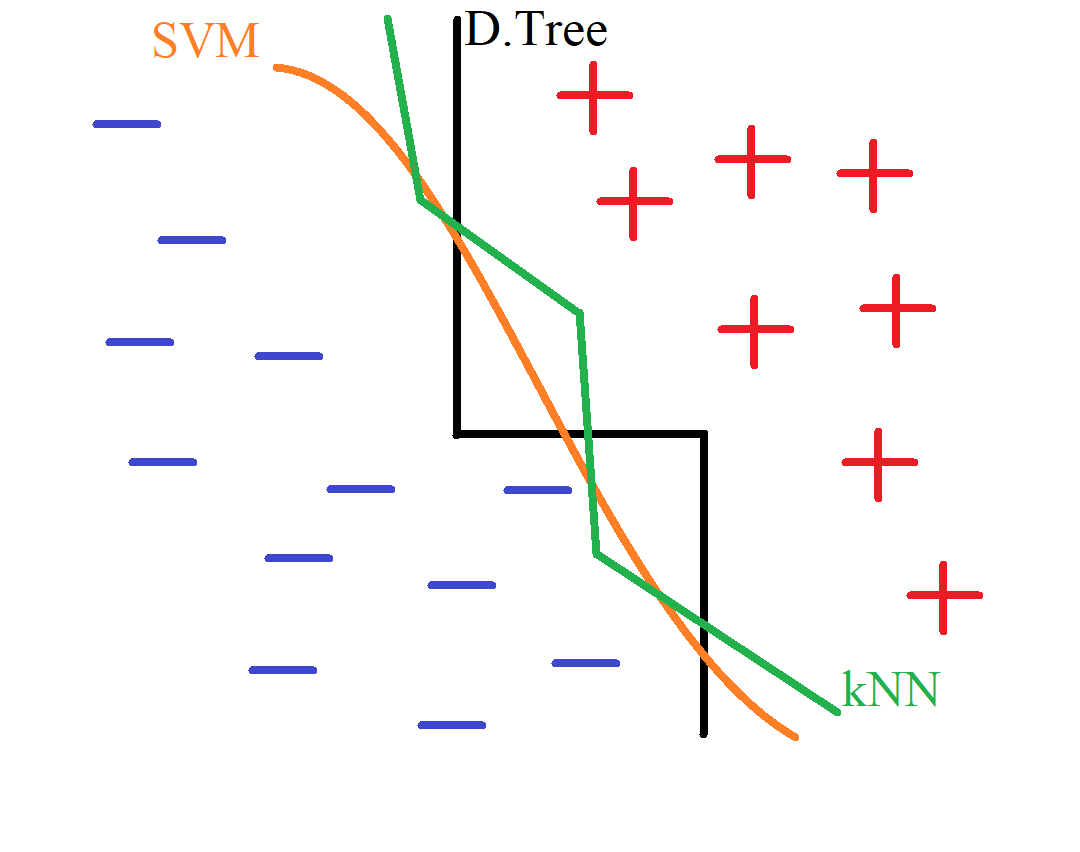}
	\caption{At the left, a comparison of classifiers: input data, nearest neighbors, support vector machine (using a radial basis function), decision tree, and neural net. This example is meant to demonstrate how different classifiers establish their decision boundaries. The diagrams display the training data in solid hues and the testing data in a translucent form. Classification accuracy on the test set is displayed in the lower right corner \citep{scikit-learn}. Right: illustrative summary of the boundary situation when comparing these classifiers (after \citeauthor{domingos2012few} \citeyear{domingos2012few}).}
	\label{fig8}
\end{figure}

Shifting to an adaptive approach that incorporates various models calibrated to specific regions instead of a single-fits-all approach requires addressing a critical issue: the selection of a model type. Different models have distinct conceptual frameworks, which influence how calibrations can be compared within the same kind of model. For example, models that are structured like graphs, including neural networks or decision trees, might pose challenges in comparison and aggregation due to alterations in the graph configuration with changing response functions. Conversely, models with vector-based structures, such as logistic regression or support vector machines, generally avoid these issues. Figure \ref{tree} illustrates how the predictive model structure in the decision tree method changes with different patterns. Disquisitions about the efficacy of varied approaches may seem trivial when selecting a predictive method since, in many geoscientific situations, significant distinctions between data-driven methods become evident primarily in the extrapolation region of the attribute space (Fig. \ref{fig8}) where there is no data to support the input-response relationship (\citeauthor{RODRIGUEZGALIANO2015804} \citeyear{RODRIGUEZGALIANO2015804};  \citeauthor{sun2020data} \citeyear{sun2020data};   \citeauthor{MAEPA2021103968} \citeyear{MAEPA2021103968};   \citeauthor{YOUSEFI2024105930} \citeyear{YOUSEFI2024105930}). 

In this paper, an ad-hoc mathematical framework is presented on which different response models $ \mathcal{m}_{\textbf{u}_\alpha} $, established at different $ \textbf{u}_\alpha $ locations on $ D $, can be pooled together, enabling the statistical inference globally and locally on the geographical domain. The framework assumes that the response model itself is a regionalized variable, which can be regarded as one among many possible realizations of a dual random field (dRF) $ \mathbbcal{M}(\textbf{u}) $, that is, a RF attaining values in the space of (real-valued) linear functions on $ \R^p $. As we shall see, by restricting ourselves to work with linear functions, the problem is reduced to study of a single RF $ \textbf{S}(\textbf{u}) $ attaining values on the unit
hyper-sphere in $ p $ dimensions, $ \textbf{S} : D \to \S^{p-1} $, with $ \S^{p-1} = \{ \textbf{x} \in \R^p : \norm{\textbf{x}} = 1\}$, with $ \norm{\cdot} $ denoting the Euclidean norm. This type of process can be understood as a regionalized orientation. (This type of RF must not be confused with the ones on which the geographical domain has a spherical shape, $ \textbf{Y} : \S^2 \to  \R$, typically named RFs ``on the sphere''(\citealt{van2002kriging}).)

A case study showcases the application of the dRF approach in MPM, categorizing the domain according to its mineral potential. Different local classification models $ \mathcal{m}_{\textbf{u}_\alpha} $ are first inferred by using a support vector classifier (or support vector machine, SVM). The different models can be pooled together easily as any SVM is defined by its decision boundary's normal vector in $ \R^p $ (orientation) and an origin offset. Traditional spatial inference and uncertainty assessment methods can then be employed in the set of orientations and offsets to obtain response models at unsampled/unexplored locations. 
Spatial SVMs have been employed in the past for geographical datasets (\citeauthor{andris2013support} \citeyear{andris2013support}); however, the treatment has been done in the response outcome rather than in the models. Still, in this study, we fill some gaps with respect to the geometric aspects of linear response functions required to enable both the variographic analysis (spatial continuity structure of models) and the spatial uncertainty assessment of inferred responses. %

The structure of this paper is the following: in Sect. \ref{Argument}, a brief argument is given to motivate the use of a local approach in modeling response functions. Section  \ref{Dual random} presents the dual random field formalism and its relationship to the classical random field. In Sect. \ref{Grassmanian}, we explain how to build a random field compatible with different support vector classification models, adjusted at different geographical locations. A key step shown in this section is the reduction of the problem to the study of random fields on the hyper-sphere. 
Section  \ref{Statmodcon} presents a collection of results that enables the simulation of random fields on the hyper-sphere, honoring different statistical aspects of input data such as their spatial continuity and their probability distribution in the hyper-sphere. Finally, a real case study is presented in Sect. \ref{Real}, on which an MPM is built aiming to discover Nickel deposits at the Yilgarn Craton, Western Australia, using different geophysical input signatures and shallowly acquired Nickel geochemistry data. The conclusions of the work are presented in Sect. \ref{Conclusions}.

\section{Argument for spatially indexed response models}
\label{Argument}

Although a spatially indexed response model can be considered for granted model in the MPM practice, it can still be motivated from a deterministic perspective. Assume that there is a global deterministic equation, $ \mathcal{m} $, governing the occurrence of mineral deposits, $ y $, depending of $ q $ input variables ($ p < q $), $ {z}_1,\dots, {z}_p,\dots, {z}_q $, that is
\[y=\mathcal{m}\big({z}_1,\dots, {z}_p,\dots, {z}_q\big).\]
As we shall see, it is enough to face a situation where limited access to a subset of these input layers is presented, say the first $ p $ of them, to obtain a spatially indexed response model. As the input variables are regionalized variables, the corresponding outcomes of the global response model $ \mathcal{m} $, at two different locations $ \textbf{u}_1 $ and $ \textbf{u}_2 $, are given by
\[y_1=\mathcal{m}\big({z}_1(\textbf{u}_1),\dots, {z}_p(\textbf{u}_1),\dots, {z}_q(\textbf{u}_1)\big) \]
and
\[ \quad  y_2=\mathcal{m}\big({z}_1(\textbf{u}_2),\dots, {z}_p(\textbf{u}_2),\dots, {z}_q(\textbf{u}_2)\big),\]
respectively.

Therefore, if the situation is such that at least one of last $ q-p $ inaccessible variables differ according to the location, that is, if $ {z}_i(\textbf{u}_1) \neq {z}_i(\textbf{u}_2) $ for any $ i \in \{p+1,\dots,q\} $, then it is no longer possible to fully infer $ \mathcal{m} $, forcing us then to consider models conditioned to the last $ q-p $ variables $ \mathcal{m}_{z_{p+1},\dots, {z}_q} $. At $ \textbf{u}_1 $ and $ \textbf{u}_2 $ the situation can then be described through the models
\[y_1=\mathcal{m}_{z_{p+1}(\textbf{u}_1),\dots, {z}_q(\textbf{u}_1)}\big({z}_1(\textbf{u}_1),\dots, {z}_p(\textbf{u}_1)\big) \]
and
\[ y_2=\mathcal{m}_{z_{p+1}(\textbf{u}_2),\dots, {z}_q(\textbf{u}_2)}\big({z}_1(\textbf{u}_2),\dots, {z}_p(\textbf{u}_2)\big),\]
respectively. In particular, the dependence to the $ q-p $ inaccessible regionalized variables can be reduced just to the particular location of application of $ \mathcal{m} $, leaving us with the spatially indexed response models
\[y_1=\mathcal{m}_{\textbf{u}_1}\big({z}_1,\dots, {z}_p\big) \quad \textmd{and} \quad y_2=\mathcal{m}_{\textbf{u}_2}\big({z}_1,\dots, {z}_p\big).\]

The use of a spatially indexed response model $ \mathcal{m}_{\textbf{u}} $ can be justified, in summary, due to both the multi-dimensional complexity and our limited understanding of global response functions, which can hide further (possibly inaccessible) dependencies to additional variables. As such, we are likely to observe changes in input-outcome relations according to the spatial location under analysis. 

\section{Dual random field formalism}
\label{Dual random}

Let us recall that a $ \R^p$-valued spatial RF $ \textbf{Z} $ defined over a geographical domain $ D \subset \R^n $ and the probability space $ (\upOmega,\mathbbcal{F},\Prob) $ is a $ \R^p$-valued
function $ \textbf{Z} : D\times\upOmega \to \R^p $ that is  $ \mathbbcal{F} $-measurable on its second variable, namely, satisfying the property that, at any $ \textbf{u} \in D$,
\begin{equation}
\textbf{Z}_\textbf{u}^{-1}(I):=\{ \omega \in \upOmega : \textbf{Z}(\textbf{u},\omega) \in I\} \in \mathbbcal{F} \label{rf}
\end{equation}
for each half-open interval set in $ \R^p $  of the form
\begin{equation}
I = (-\infty,t_1]\times(-\infty,t_2]\times\dots\times(-\infty,t_p]. \nonumber
\end{equation}
(We refer to the book by \citeauthor{adler2010geometry} \citeyear{adler2010geometry} for a complete description of this formalism.) As usual, the set $ \upOmega $ is omitted from this definition to get the simpler but less exact definition of a RF, as the function
\begin{IEEEeqnarray*}{rrCl}
\textbf{Z} : &\, D & \to & \R^p\\
& \textbf{u} & \mapsto & \textbf{Z}(\textbf{u}) = \textbf{\textit{z}}_\textbf{u},
\end{IEEEeqnarray*}
with $ \textbf{Z}(\textbf{u}) $ a $ \R^p$-valued random variable (RV) completely described by the set of distribution functions
\begin{equation}
F(\textbf{u}_1,\dots,\textbf{u}_k;I_1,\dots,I_k) = \Prob\big(\textbf{Z}(\textbf{u}_1) \in I_1,\dots,\textbf{Z}(\textbf{u}_k)\in I_k\big),\nonumber
\end{equation}
for all positive integers $ k $, all configurations of points $ \textbf{u}_1,\dots,\textbf{u}_k $, and any sequence $ I_1,\dots,I_k $ of half-open intervals in $ \R^p $.

In the same spirit, we seek to define a spatially indexed mapping $\mathbbcal{M}$ into the set of responses $ \{\mathcal{m} : \R^p \to \R\} $ in a RF fashion
\begin{IEEEeqnarray*}{rrCl}
\mathbbcal{M} : &\, D & \to & \{\mathcal{m} : \R^p \to \R\}\\
& \textbf{u} & \mapsto & \mathbbcal{M}(\textbf{u}) = \mathcal{m}_\textbf{u},
\end{IEEEeqnarray*}
such that, when considering any $\R^p$-valued RF $ \textbf{Z}(\textbf{u}) $ or any $\R^p$-valued regionalized variable $ \textbf{\textit{z}}(\textbf{u}) $ for the evaluation of $\mathbbcal{M}$ at $\textbf{u}$, both $ \mathbbcal{M}(\textbf{u})[\textbf{Z}(\textbf{u})] $ and $ \mathbbcal{M}(\textbf{u})[\textbf{\textit{z}}(\textbf{u})] $ attain random values in $\R$. In the first case, the randomness at \textbf{u} is given by $ \mathbbcal{M}(\textbf{u}) $ and $ \textbf{Z}(\textbf{u}) $ simultaneously and, in the second case, only by $ \mathbbcal{M}(\textbf{u}) $.

Since the set of functions $ \{\mathcal{m} : \R^p \to \R\} $ can be quite general, we restrict this set to the case of linear functions. This allows us to keep control in the modeling of usual geostatistical notions such as mathematical expectation, $ \E\{\mathbbcal{M}(\textbf{u})\} $, covariance given a separation vector $ \textbf{h} $, $ \Cov\{\mathbbcal{M}(\textbf{u}),\mathbbcal{M}(\textbf{u}+\textbf{h})\} $,  (non-)stationarity and anisotropy, as we shall see. In addition, this linear imposed simplification enables the spatial simulation of scenarios and, therefore, the uncertainty analysis of MPM models.

For the purpose of having a better understanding of linear functions, it is enough to introduce the notion of dual (or conjugate) space (\citeauthor{kolmogorov1975introductory} \citeyear{kolmogorov1975introductory}; \citeauthor{Tu} \citeyear{Tu}) as follows. Let $ V $ and $ W $ be two real vector spaces. We denote by $ \textmd{L}(V,W) $ the vector space of all linear maps $ f : V \to W $. Then, the dual space $ V^{*} $ of $ V $ is defined as the vector space of all real-valued linear functions on $ V $ \[V^{*} = \textmd{L}(V,\R). \]

In what follows, the reader can have in mind that $ V = \R^p $. Any element in $ V^{*}  $ is called covector simply because the usual notion of basis $ e_1,\dots,e_p $ has its dual version on $ V^{*}$, meaning that any linear function $ g \in V^{*} $ can be decomposed in an orthonormal basis of linear functions $ e^{*}_1,\dots,e^{*}_p $. If $ v=\sum_{i=1}^p a_ie_i \in V $ with $ a_i \in \R $, then $ e^{*}_i : V \to \R $ is a linear function that returns the $ i $th projection (into $ e_i $) of $ v $, that is, $ e^{*}_i(v)= a_i $. Consequently, if $ v $ has a zero component in the $ i $-th direction, there is no value to be projected, and if $ v $ has a unit component in the $ i $-th direction, then $ e^{*}_i(v)= 1 $. In particular,
\[e^{*}_i(e_j)=\begin{cases}
			1, & \text{for $i=j$}\\
            0, & \text{for $i\neq j$.}
		 \end{cases}\]
Any linear function $ f $ is then uniquely determined by how it acts on the unit vectors $ e_1,\dots,e_p $ and, if $ b_i=f(e_i) $, $ f $ can be decomposed as $ f(\,\cdot\,) =\sum_{i=1}^p b^{}_ie^{*}_i(\,\cdot\,) $.

In order to define a notion of dual RF, it is key to note that $ V $ is isomorphic to $ V^{*}$, or loosely speaking, that $ V^{*}$ is a copy of $ V $. Any element $ v \in V $ can be ``transformed'' into a linear function, for example, by the use of the inner product map in $ V $, $ \langle \cdot,\cdot \rangle $,  which gives a mapping from V into its dual space via the linear map $ T $,
\begin{IEEEeqnarray*}{rrCl}
T : &\, V & \to & V^{*}\\
& v & \mapsto & Tv = \langle v ,\cdot \rangle.
\end{IEEEeqnarray*}
We can then define a $ {\R^p}^*$-valued spatial dual random field (dRF) $ \mathbbcal{M} $ over a geographical domain $ D \subset \R^n $ and the probability space $ (\upOmega,\mathbbcal{F},\Prob) $ simply as the composition of a RF $ \textbf{M}$ with the linear mapping $ T $, that is, $ \mathbbcal{M} = T \circ \textbf{M} $ such that any pair $ (\textbf{u},\omega) \in D\times \upOmega$ is mapped into a random linear function attaining values in $ \R $ through the diagram
\[
\begin{tikzcd}[column sep=-0.5em]
 & \textbf{M}(\textbf{u},\omega) \arrow{dr}{T}\\
(\textbf{u},\omega) \arrow{ur}{\textbf{M}} \arrow{rr}{\mathbbcal{M}} && \mathbbcal{M}(\textbf{u},\omega)[\,\cdot\,]  & {}=  \langle \textbf{M}(\textbf{u},\omega) ,\cdot \rangle.
\end{tikzcd}
\]
This definition allows us to study any dRF using the same tools employed in the study of RFs.

\section{SVM random fields and the Grassmanian manifold}
\label{Grassmanian}

In what follows, we attempt to infer a set of local linear models $ \{ \mathcal{m}_\alpha : \R^p \to \R\}^N_{\alpha=1} $ from a set of $ N $ observations $ \{\textbf{\textit{z}}_\alpha, {\textit{y}}_\alpha\}^N_{\alpha=1} \in \R^p\times \{0,1\} $. This procedure is carried out using SVMs as classifiers \citep{vapnik1995nature}.

\subsection{Support vector machine}

In the two-group classification problem, counting with a set of $ L $ training pairs (or calibration samples), $ \{\textbf{\textit{z}}_\alpha, {\textit{y}}_\alpha\}^L_{\alpha=1} \in \R^p\times \{-1,1\} $, a linear function $ \mathcal{m}(\textbf{\textit{z}}) =\left\langle\textbf{v},\textbf{\textit{z}}\right\rangle+{b}$ (in our case, $\left\langle\textbf{v},\textbf{\textit{z}}\right\rangle=\textbf{v}^T\cdot\textbf{\textit{z}}$) associated to the classifier $ \textmd{Class}(\textbf{\textit{z}}) = \textmd{sign}\big(\mathcal{m}(\textbf{\textit{z}})\big) $, with $ (\textbf{v},{b}) \in \R^p\times\R $, is to be
estimated by solving the optimization problem
\begin{mini*}|v|
{{\textbf{v}},{b}}{\frac{1}{2}\norm{{\textbf{v}}}^2}
{}{}
\addConstraint{y_{\alpha}\cdot\big(\left\langle{\textbf{v}},\textbf{\textit{z}}_\alpha\right\rangle + {b}\big) \ge 1,\qquad \alpha = 1,\dots ,L.}
{}
\end{mini*}
Let $ \hat{\beta} \in \R^p  $ be an offset ( or shifting) vector of length $ b $ in the direction of \textbf{v}. Then, the set given by
\[\{\textbf{\textit{z}}\in\R^p\,| \, m (\textbf{\textit{z}}) = 0\} = \textmd{ker}({\textbf{v}}^T) + \hat{\beta}\]
is called the ``maximum-margin hyperplane'' between the two classes \citep{lim2021grassmannian}. Notice that once the vector $ \textbf{s} $ is found, the subspace $ \textmd{ker}({\textbf{v}}^T) + \hat{\beta} $ is invariant by a simultaneous re-scaling of $ \textbf{v} $ and $\hat{\beta}$, allowing us to consider a unit vector $\textbf{s} = {\textbf{v}}^T/\norm{\textbf{v}} $ instead to define the same subspace, $ \textmd{ker}({\textbf{v}}^T/\norm{\textbf{v}}) + \norm{\textbf{v}}\cdot\hat{\beta} = \textmd{ker}({\textbf{s}}^T) + \hat{\beta}' $.

Slack variables defined by
\[\xi_\alpha = \textmd{max}\big(0,1-{\textit{y}}_\alpha\cdot\mathcal{m}(\textbf{\textit{z}}_\alpha)\big),\]
associated to each sample $ \{\textbf{\textit{z}}_\alpha, {\textit{y}}_\alpha\}\in \R^p\times \{-1,1\} $ are introduced when it is not possible to fully separate the classes, allowing points to
be on the wrong side of their margin. A cost term $C$ defined by the user is introduced within the objective function in order to penalize whenever a sample lies on the wrong side of the margin. The optimization problem then becomes
\begin{mini*}|v|
{{\textbf{v}},{b}}{\frac{1}{2}\norm{\textbf{v}}^2+C\sum^{L}_{\alpha=1}\xi_\alpha}
{}{}
\addConstraint{y_{\alpha}\cdot\big(\left\langle{{\textbf{v}}},\textbf{\textit{z}}_\alpha\right\rangle + {b}\big) \ge 1 - \xi_\alpha,\qquad \alpha = 1,\dots ,L.}
{}
\end{mini*}
The larger $C$ is, the higher is the
penalty to misclassification. The solution to this minimization is usually found by setting a Lagrangian formulation of the problem. We refer to the extended literature in SVMs for more details (\citealt{cristianini2000introduction}), and the examples of application in MPM (\citealt{zuo2011support}; \citealt{abedi2012support}; \citealt{granek2015data}; \citealt{chen2017application}).

A set of SVM separating hyperplanes $ \{ \mathcal{m}_\alpha : \R^p \to \R\}^N_{\alpha=1} $ are obtained at each data location by restricting the calibration to a vicinity of $  \textbf{u}_\alpha $, $ \mathbbcal{V}(\textbf{u}_\alpha) $, defined either by fixing a search radius around \textbf{u} or by fixing the cardinality $ |\mathbbcal{V}(\textbf{u}_\alpha)|=l_\alpha $ of the closest samples to be considered for the adjustment. The number of samples in the neighborhood must be chosen through an iterative process that needs to consider different factors, such as data spacing and its density on the domain, and must be validated through cross-validation procedures (\citealt{deutsch1992geostatistical}).

Once the models $ \{ \mathcal{m}_\alpha \}^N_{\alpha=1} $ are calibrated and the pairs $(\textbf{s}_\alpha,{b}_\alpha) \in \S^{p-1}\times\R$, $\alpha \in \{1,\dots,N\}$  characterizing the separation hyperplanes are found, we seek for studying in detail on the next section the following hyperplane process, denoted as $ \mathbbcal{M}(\textbf{u}) $, composed of a dRF $ \mathbbcal{S}(\textbf{u}) $ and an offset real-valued  RF $ B(\textbf{u}) $
such that, given a vector $ \textbf{z} \in \R^p $,
\[\mathbbcal{M}(\textbf{u})[\textbf{z}]=\mathbbcal{S}(\textbf{u})[\textbf{\textit{z}}]+B(\textbf{u}).\]
A corresponding SVM RF, denoted by $ \mathbbcal{SVM}(\textbf{u}) $, associated to $ \mathbbcal{M}(\textbf{u}) $ is then built simply as
\[\mathbbcal{SVM}(\textbf{u})=\textmd{sign}\big(\mathbbcal{M}(\textbf{u})[\textbf{\textit{z}}]\big).\]
On a conditional version of $ \mathbbcal{M}(\textbf{u}) $, we request that the corresponding RF $ \textbf{S}(\textbf{u}) $ associated to $ \mathbbcal{S}(\textbf{u})=\left\langle\textbf{S}(\textbf{u}),\cdot\right\rangle $ and $ B(\textbf{u}) $ attain the values
\[\textbf{S}(\textbf{u}_\alpha)=\textbf{s}_\alpha\qquad \textmd{and}\qquad B(\textbf{u}_\alpha)={b}_\alpha\]
respectively, at $ \textbf{u}_\alpha, \alpha\in \{1,\dots,N\} $.

\subsection{Grassmannian geometry of support vector machines}

We have seen that any separating hyperplane associated with an SVM is a hyperplane passing through the origin translated by an offset (or displacement) vector. If we focus on the hyperplane part of the problem, we notice that this is a particular case of a much general object of study, corresponding to the set of $ k $-dimensional subspaces of $ \R^{p} $ passing through the origin, denoted by $ \textmd{G}(k,p) $ \citep{lim2021grassmannian}. In general, $ \textmd{G}(k,p) $ is a manifold called the Grassmaniann manifold of $ k $-dimensional subspace of $ \R^{p} $, and has a varying geometry according with $ k $ \citep{bendokat2011grassmann}.

In the case of SVMs, it is also important to consider the orientation of the hyperplane, as it divides $ \R^{p} $ into two half-spaces labeled in a different manner. We denote by $ \tilde{\textmd{G}}(k,p) $ set of $ k $-dimensional oriented subspaces of $ \R^{p} $. 

In our case, it is enough to consider the case $ k=p-1$, $ \tilde{\textmd{G}}(p-1,p) $, which coincides with the unit hyper-sphere $ \S^{p-1} $. This can be seen by assigning to each hyperplane its corresponding unit normal (\citealt{hoffman1980geometry}; \citealt{kobayashi1996foundations}).

As a final remark, we note that the geometry of $ \tilde{\textmd{G}}(k,p) $ is different from the one of $ {\textmd{G}}(k,p) $. For example, $ \tilde{\textmd{G}}(2, 3) $  is equal to the 2-sphere, but $ {\textmd{G}}(2,3) $ is a projective space (or half-sphere), useful in the mapping of faults in structural geology.

\section{Statistical modeling considerations on \texorpdfstring{$ \S^{p-1} $}{Lg}}
\label{Statmodcon}

We have been able to reduce the problem of working with a linear function random process to the study of a single RF $ \textbf{S}(\textbf{u}) $, attaining values on the unit hyper-sphere in $ p $ dimensions, $ \S^{p-1} $. The literature dealing with this type of process, also known as regionalized orientations, does not find major difficulties on inference procedures (\citeauthor{fisher1995statistical} \citeyear{fisher1995statistical}; \citeauthor{ley2017modern} \citeyear{ley2017modern}) nor spatial estimation (\citeauthor{young1987indicator} \citeyear{young1987indicator}, \citeyear{young1987random}; \citeauthor{van2002kriging} \citeyear{van2002kriging}) in the case of arbitrary dimensions. The spatial simulation of these fields, however, has been rather limited to the case $ p = 2 $ or angular case (\citeauthor{jona2012spatial} \citeyear{jona2012spatial}; \citeauthor{wang2013directional} \citeyear{wang2013directional}), and prevented from going beyond in dimensionality due to difficulties
to obtain analytical results in the spatial correlation analysis. In the angular case, it is possible to use a complex random variable (\citeauthor{wackernagel2013multivariate} \citeyear{wackernagel2013multivariate}; \citeauthor{de2022new} \citeyear{de2022new}), simplifying the treatment of variography to the real and the imaginary part of the random variable by separate. However, this complexification approach can not be carried out into arbitrary higher dimensions. Here, we generalize the study of regionalized orientations to any dimension.  

\subsection{Variography on \texorpdfstring{$ \S^{p-1} $}{Lg}}

In order to measure the spatial continuity of any $ \S^{p-1} $-RF \textbf{S}, we need to introduce an ad-hoc definition of covariance. Let $ \textbf{v} $ and $ \textbf{w} $ two random unit vectors drawn on $ \S^{p-1} $. Then, their covariance is defined as
\begin{align}
\Cov^{}_\S(\textbf{v},\textbf{w}) =\E\big(\langle \textbf{v},\textbf{w} \rangle\big).
\end{align}
(We include the sub-index $\S$ to distinguish this definition from the usual one given in real-valued RFs.)
This definition attains a zero value when both $ \textbf{v} $ and $ \textbf{w} $ are independent vectors following a uniform distribution on $ \S^{p-1} $, as the average inner product $ \langle \textbf{v},\textbf{w} \rangle $ is zero (half of the times $ \textbf{v} $ and $ \textbf{w} $ are in the same hemisphere and the other half in opposite hemispheres). As usual, we say that $ \textbf{S} $ is second-order stationary if the covariance is just a function of the distance $ \textbf{h} $ between $ \textbf{u}_1 $ and $ \textbf{u}_2 $, $ \Cov^{}_\S\big(\textbf{S}(\textbf{u}_1),\textbf{S}(\textbf{u}_2)\big) = \Cov^{}_\S(\textbf{u}_1-\textbf{u}_2)  = C^{}_\S(\textbf{h})$. 

This definition of covariance is related to the one of variogram introduced by \cite{young1987indicator,young1987random} \[ \gamma^{}_\S(\textbf{u}_1-\textbf{u}_2) =\frac{1}{2}\E\big(\norm{\textbf{S}(\textbf{u}_1)-\textbf{S}(\textbf{u}_2)}^2\big),\] through the identity $ \gamma^{}_\S(\textbf{h}) = C^{}_\S(\textbf{0}) -C^{}_\S(\textbf{h})= 1 - C^{}_\S(\textbf{h}) $, which can be checked by the reader using that $\norm{\textbf{v}-\textbf{w}}^2 = \langle{\textbf{v}-\textbf{w}},{\textbf{v}-\textbf{w}}\rangle = \norm{\textbf{v}}^2 + \norm{\textbf{w}}^2 - 2\langle \textbf{v},\textbf{w} \rangle $.

\subsection{Non-conditional simulation of a uniform \texorpdfstring{$ \S^{p-1} $}{Lg}-RF}

In order to build a $ \S^{p-1} $-RF \textbf{S}, we start with $ p $ independent real-valued Gaussian RFs $ X_1(\textbf{u}),\dots, X_p(\textbf{u})$, $ \textbf{u} \in D \subset \R^n $. We assume that each $ X_i $ has zero mean and the same covariance function, $ C(\textbf{h}) $, with variance 
$ C(\textbf{0}) = 1 $. From these we define a process $ \textbf{S} : D \to \S^{p-1} $ by setting first the vector-RF $ \textbf{X}(\textbf{u}) = \big(X_1(\textbf{u}),\dots, X_p(\textbf{u})\big)^T $ and then computing
\[\textbf{S}(\textbf{u}) = \frac{\textbf{X}(\textbf{u})}{\norm{\textbf{X}(\textbf{u})}},\]
where $ \norm{\textbf{X}(\textbf{u})} = \sqrt{[X_1(\textbf{u})]^2+\dots+[X_p(\textbf{u})]^2} $. It is straightforward to check that the probability density function (PDF) for $ \textbf{S}(\textbf{u}) $ is the uniform distribution (\citeauthor{muller1959note} \citeyear{muller1959note}), assigning equal probabilities to all the points of the sphere, $ \textbf{x} \in \S^{p-1} $
\[ f_{\textmd{Uniform}^{}_{\S^{p-1}}}(\textbf{x}) = \frac{1}{A_{p-1}}=\displaystyle\frac{\upGamma\big(\frac{p}{2}\big)}{2\pi^{\frac{p}{2}}},\]
with $ A_{p-1} $ the surface area of $\S^{p-1} $ and $ \upGamma(\,\cdot\,) $ the standard gamma function, defined recursively by
\[\upGamma(p) = (p - 1)\upGamma(p - 1), \qquad \upGamma(1) = 1, \qquad \upGamma\bigg(\frac{1}{2}\bigg) = \sqrt{\pi}.\]

The covariance structure of a $ \S^{p-1} $-RF is a consequence of an observation made by \cite{hotelling1953new} in the context of relating the theoretical correlation in a set of drawn pairs of samples following a bivariate Gaussian distribution with correlation coefficient $ \rho $, and the Pearson correlation coefficient inferred from such set. For a set of $ L $ samples, the inferred Pearson correlation coefficient is itself a random variable $ R \in (-1,1) $ with density function $ f_{R}(r) $ following the involved but analytical formula (\citealt{fisher1915frequency})

\small
$$ {\displaystyle f_R(r)={\frac {(L-2)\,{\upGamma } (L-1)(1-\rho ^{2})^{\frac {L-1}{2}}(1-r^{2})^{\frac {L-4}{2}}}{{\sqrt {2\pi }}\,{\upGamma } (L-{\tfrac {1}{2}})(1-\rho r)^{L-{\frac {3}{2}}}}}\cdot{}_{2}{F} _{1}{\bigl (}{\tfrac {1}{2}},{\tfrac {1}{2}};{\tfrac {1}{2}}(2L-1);{\tfrac {1}{2}}(\rho r+1){\bigr )}}, $$
\normalsize
with $ {\displaystyle {}_{2} {F} _{1}(a,b;c;z)} $ the Gaussian hyper-geometric function, defined for $ |z| < 1 $ by the power series
$$ {\displaystyle {}_{2}F_{1}(a,b;c;z)=\sum _{n=0}^{\infty }{\frac {(a)_{n}(b)_{n}}{(c)_{n}}}{\frac {z^{n}}{n!}}=1+{\frac {ab}{c}}{\frac {z}{1!}}+{\frac {a(a+1)b(b+1)}{c(c+1)}}{\frac {z^{2}}{2!}}+\cdots.}$$
We reinterpret this formula as follows. Let $ \textbf{S}(\textbf{u}_1) = \textbf{s}_1 $  and $ \textbf{S}(\textbf{u}_2) = \textbf{s}_2 $ a realization of $ \textbf{S}  $ at two locations $ \textbf{u}_1 $ and $ \textbf{u}_2 $, obtained respectively from the vectors \[ \textbf{X}(\textbf{u}_1) = \textbf{x}_1 = [{x}_{11},\dots,{x}_{p1}]^T,\textmd{ and} \qquad\textbf{X}(\textbf{u}_2) = \textbf{x}_2 =[{x}_{12},\dots,{x}_{p2}]^T.\]
Then, their inner product
$$ {\displaystyle \langle \textbf{s}_1,\textbf{s}_2 \rangle=\left\langle \frac{\textbf{x}_1}{\norm{\textbf{x}_1}},\frac{\textbf{x}_2}{\norm{\textbf{x}_2}} \right\rangle= \frac{1}{\norm{\textbf{x}_1}\norm{\textbf{x}_2}} \left\langle {\begin{pmatrix}
x_{11} \\ \vdots \\ x_{p1}
\end{pmatrix},\begin{pmatrix}
x_{12} \\ \vdots \\ x_{p2}
\end{pmatrix}} \right\rangle ={\frac {\sum _{i=1}^{p}x_{i1}x_{i2}}{{\sqrt {\sum _{i=1}^{p}x_{i1}^{2}}}{\sqrt {\sum _{i=1}^{p}x_{i2}^{2}}}}}} $$
coincides with the definition of the Pearson correlation coefficient $ R $ for a set of $ L = p +1 $ pairs of samples
on which an ad-hoc orthogonal transformation can be applied to decrease the degree of freedom by one and make both the sample mean and the sample variance disappear from the usual formulation (\citealt{hotelling1953new}).

In our re-interpretation, the pairs $ (x_{i1},x_{i2}) $, $ i\in \{1,\dots,p\} $, follow a bivariate Gaussian distribution with theoretical $ \rho = C(\textbf{h}) $. Therefore, it follows that the inner product $ \langle \textbf{S}(\textbf{u}),\textbf{S}(\textbf{u}+\textbf{h}) \rangle $ have a PDF given by
\scriptsize
$$ {\displaystyle f^{}_{\langle \textbf{S}(\textbf{u}),\textbf{S}(\textbf{u}+\textbf{h}) \rangle}(r)={\frac {(p-1)\,{\upGamma } (p)\big(1-C(\textbf{h}) ^{2}\big)^{\frac {p}{2}}\big(1-r^{2}\big)^{\frac {p-3}{2}}}{{\sqrt {2\pi }}\,{\upGamma } \big(p+{\tfrac {1}{2}}\big)\big(1-C(\textbf{h}) r\big)^{p-{\frac {1}{2}}}}}\cdot{}_{2}{F}_{1}{\bigl (}{\tfrac {1}{2}},{\tfrac {1}{2}};{\tfrac {1}{2}}(2p+1);{\tfrac {1}{2}}(rC(\textbf{h})+1){\bigr)}}.$$
\normalsize

The expectation of this PDF (\citeauthor{ghosh1966asymptotic} \citeyear{ghosh1966asymptotic}; \citeauthor{shieh2010estimation} \citeyear{shieh2010estimation}; \citeauthor{humphreys2019underestimation} \citeyear{humphreys2019underestimation}) coincides with our definition of covariance
\begin{align}
C^{}_\S(\textbf{h}) = \frac{2}{p}\left\lparen\frac{\upGamma\big(\frac{p+1}{2}\big)}{\upGamma\big(\frac{p}{2}\big)}\right\rparen^2\cdot C(\textbf{h})\cdot{}_{2}{F}_{1}\left\lparen\frac{1}{2},\frac{1}{2};\frac{p+2}{2};C(\textbf{h})^2\right\rparen\nonumber.
\end{align}

In particular, this result contains the case $p = 1$ $\big( \S^0 = \{-1,1\}\big) $ which coincides with the well-known result for the indicator covariance of the thresholded Gaussian random function $ X(\textbf{u}) $ at level $ 0 $ (\citealt{xiao1985description}), verified by using that $ 1_{X(\textbf{u})>0}=I(\textbf{u})=\frac{1}{2}\big(\textbf{S}(\textbf{u})+1\big) $, $ \E\big(\textbf{S}(\textbf{u})\big)=0 $ and $ \arcsin t = t\cdot {}_{2}{F}_{1} (1/2,1/2;3/2; t^2) $. Then
\begin{align}
\Cov(I(\textbf{u}),I(\textbf{u}+\textbf{h})) &= \E\big(I(\textbf{u})I(\textbf{u}+\textbf{h})\big)-\E\big(I(\textbf{u})\big)\E\big(I(\textbf{u}+\textbf{h})\big)\nonumber\\
&= \E\bigg(\Big(\frac{1}{2}\textbf{S}(\textbf{u})+\frac{1}{2}\Big)\Big(\frac{1}{2}\textbf{S}(\textbf{u}+\textbf{h})+\frac{1}{2}\Big)\bigg)-\frac{1}{4}\nonumber\\
&=\frac{1}{4} \Cov^{}_\S\big(\textbf{S}(\textbf{u}),\textbf{S}(\textbf{u}+\textbf{h})\big)+\frac{1}{4}-\frac{1}{4}\nonumber\\
&= \frac{1}{2\pi}\arcsin C(\textbf{h}).\nonumber
\end{align}

\subsection{Geometrical aspects of \texorpdfstring{$ \S^{p-1} $}{Lg}}

We still have left to build a conditional simulation procedure and deal with arbitrary distributions on $ \S^{p-1} $-RF besides the uniform one. In order to do this, we are required to introduce some general notions of geometry on the unit hyper-sphere. We restrict this exposition only to those main geometric quantities exclusively needed to achieve these purposes. 

\begin{description}[leftmargin=0cm,itemsep=4pt]

\item[\textbf{Geodesic Distance}$ \quad $] Given any two points $ \textbf{s}_1, \textbf{s}_2 \in \S^{p-1}$, the shortest path connecting them is along the great circle, with
geodesic distance $ d(\textbf{s}_1, \textbf{s}_2) $ given by
\begin{IEEEeqnarray*}{rrCl}
d : &\, \S^{p-1}\times\S^{p-1} & \to & [0,\pi]\\
& (\textbf{s}_1, \textbf{s}_2) & \mapsto &  d(\textbf{s}_1, \textbf{s}_2) = \arccos(\left\langle\textbf{s}_1, \textbf{s}_2\right\rangle).
\end{IEEEeqnarray*}

\item[\textbf{Tangent space}$ \quad $] It will be useful to work with a linearized version of the sphere, especially when working with probability distributions on the sphere. This linearization takes place on the tangent space at a specific $ \textbf{s} \in \S^{p-1} $, characterized as
\[T_\textbf{s}\S^{p-1} = \{\textbf{v} \in \R^p : \left\langle\textbf{v}, \textbf{s}\right\rangle = 0\}.\]
which is just the hyperplane
tangent to the sphere at $ \textbf{s} $ with origin at $ \textbf{s}$.

\item[\textbf{Logarithmic map}$ \quad $] For the linearization of the sphere, we consider here an alternative to the more common stereographic projection, called the logarithmic map. A point $ \textbf{s} \in \S^{p-1} $ is mapped to $ T_\varvec{\mu}\S^{p-1} $ via the logarithmic map
\begin{align*}
\textmd{Log}_\varvec{\mu}(\textbf{s}) &= \big(\textbf{s}-\varvec{\mu}\left\langle\varvec{\mu}, \textbf{s}\right\rangle\big)\frac{\theta}{\sin{\theta}}, \\
\theta&= \arccos\left\langle\varvec{\mu}, \textbf{s}\right\rangle.
\end{align*}
As the maximal distance from $ \varvec{\mu} $ to any point $ \textbf{s} $ is $ \pi $, any tangent vectors $ \textbf{v} \in T_\varvec{\mu}\S^{p-1} $ has a length $ \norm{\textbf{v}} \leq \pi  $.

\item[\textbf{Exponential map}$ \quad $] The inverse logarithmic mapping, known as the exponential map, sends a tangent vector $ \textbf{v} $ back to the sphere via
\begin{align*}
\textmd{Exp}_\varvec{\mu}(\textbf{v}) &= \cos(\norm{\textbf{v}})\varvec{\mu} +  \sin(\norm{\textbf{v}})\frac{{\textbf{v}}}{\norm{\textbf{v}}}.
\end{align*}

\item[\textbf{Mean and weighted mean}$ \quad $] Given a set of manifold-data values $\textbf{s}_1 ,\dots,\textbf{s}_N \in \S^{p-1}$, we denote the {{geometric}} (or intrinsic) mean by $ \varvec{\mu} $,  defined by
\begin{equation}
	\varvec{\mu} = \underset{\substack{\bar{\varvec{\mu}} \in \S^{p-1}}}{\textmd{argmin}} \sum_{\alpha=1}^N d^2(\bar{\varvec{\mu}},\textbf{s}_i).
	\label{freme}
\end{equation}

Similarly, it is possible to define a weighted geometric mean version, considering a set of weights $\lambda_1 ,\dots,\lambda_N \in \R$, as
\begin{equation}
	\varvec{\mu} = \underset{\substack{\bar{\varvec{\mu}} \in \S^{p-1}}}{\textmd{argmin}} \sum_{\alpha=1}^N \lambda_\alpha d^2(\bar{\varvec{\mu}},\textbf{s}_\alpha),\qquad \sum_{\alpha=1}^N\lambda_\alpha=1.\nonumber
	\label{fremean}
\end{equation}

Algorithm 1 (\citealt{pennec2006intrinsic}) is simple and converges rapidly when solving this minimization problem.

\begin{algorithm}
	\caption{Weighted mean of $ N $ orientations }\label{alg:mean}
	\begin{algorithmic}[1]
		\Require a set of $ N $ orientations  $ \textbf{s}_1,\dots,\textbf{s}_N \in \S^{p-1} $  and $ \epsilon> 0$.
		\State Initialize ${\varvec{\mu}}^{(1)} = \textbf{s}_1$
		\Repeat
		\State ${\varvec{\bar v}} = \sum_{\alpha=1}^{n}\lambda_\alpha\textmd{Log}_{{\varvec{\mu}}^{(t)}}(\textbf{s}_\alpha)$ \Comment{Arithmetic mean in the tangent space}
		\State ${\varvec{\mu}}^{(t+1)}=\textmd{Exp}_{{\varvec{\mu}}^{(t)}}({\varvec{\bar v}})$
		\Comment{Projecting back to the sphere}
		\Until{$ \norm{{\varvec{\bar v}}} < \epsilon$}\\
		\Return ${\varvec{\mu}}^{(t+1)}$	
	\end{algorithmic}
\end{algorithm}

\item[\textbf{Parallel transport}$ \quad $] A tangent vector $ \textbf{v} \in T_{\textbf{s}^{}_1}\S^{p-1} $ can be translated into the tangent space of a second point ${\textbf{s}^{}_2} \in \S^{p-1}$ via parallel transport, denoted by $\upGamma_{\textbf{s}^{}_1\to\textbf{s}^{}_2}(\textbf{v})$, obtaining a new vector $ \textbf{w}  \in T_{\textbf{s}^{}_2}\S^{p-1} $ as  (\citealt{absil2008optimization})
\begin{eqnarray}
\textbf{w} &=& \upGamma_{\textbf{s}^{}_1\to\textbf{s}^{}_2}(\textbf{v}).\nonumber\\
&=&\bigg(\textbf{v} - \textbf{p}\left\lparen\frac{\left\langle\textbf{p}, \textbf{v}\right\rangle}{\norm{\textbf{p}}^2}\right\rparen\bigg)+\frac{\left\langle\textbf{p}, \textbf{v}\right\rangle}{\norm{\textbf{p}}^2}\bigg(\textbf{s}^{}_1\big(-\sin(\norm{\textbf{p}})\norm{\textbf{p}}\big)+\textbf{p}\cos(\norm{\textbf{p}})\bigg),
\label{partras}
\end{eqnarray}
with $ \textbf{p}=\textmd{Log}_{\textbf{s}^{}_1}(\textbf{s}^{}_2) $.
\end{description}

\begin{figure}[hbtp]
\centering
\includegraphics[width=0.6\textwidth,trim={0 0cm 0cm 0cm},clip]{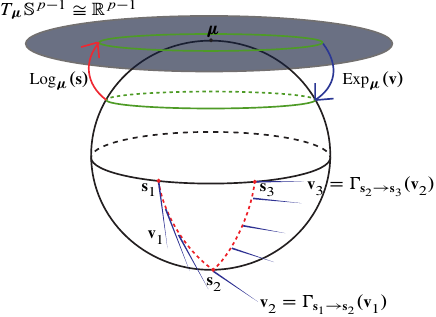}
\caption{The hyper-sphere $ \S^{p-1} $ and a scheme of the most relevant operations on it.}
\label{sphere_sque}
\end{figure}

Figure \ref{sphere_sque} illustrates all previous concepts.

\subsection{Conditional simulation of a uniform \texorpdfstring{$ \S^{p-1} $}{Lg}-RF}
\label{condi}

Given a non-conditional $ \S^{p-1} $-RF, $ \textbf{S}_\textmd{S}(\textbf{u}) $ and a set of observations $\textbf{s}(\textbf{u}_\alpha)=\textbf{s}_\alpha $ at locations $ \textbf{u}_\alpha, \alpha\in \{1,\dots,N\} $, we would like to produce a conditional RF $ \textbf{S}_\textmd{CS}(\textbf{u}) $ such that $ \textbf{S}_\textmd{CS}(\textbf{u}_\alpha) = \textbf{s}_\alpha$.

We recall that this procedure is achieved in the real-valued case, given a RF $ X(\textbf{u}) $ and a set of observations $x(\textbf{u}_\alpha)=x_\alpha $, through
\begin{align}
X_\textmd{CS}(\textbf{u})&=X_\textmd{S}(\textbf{u})+X^\textmd{SK}_\textmd{Error}(\textbf{u})\nonumber\\
&=X_\textmd{S}(\textbf{u}) + [X^\textmd{SK}_\textmd{Obs}(\textbf{u})-X^\textmd{SK}_\textmd{SimObs}(\textbf{u})]\nonumber\\
&=X_\textmd{S}(\textbf{u}) +\sum_{\alpha=1}^{N}\lambda^\textmd{SK}_\alpha(\textbf{u})[x_\alpha- X_\textmd{S}(\textbf{u}_\alpha)]\nonumber
\end{align}
with $\{\lambda^\textmd{SK}_\alpha\}^N_{\alpha=1}$ the set of simple kriging weights at $\textbf{u}$, $ X_\textmd{S}(\textbf{u}) $ the non-conditional simulation, and $ X^\textmd{SK}_\textmd{Error}(\textbf{u}) $ the interpolated field via simple kriging of the  errors $x_\textmd{Error}(\textbf{u}_\alpha)=x_\alpha- X_\textmd{S}(\textbf{u}_\alpha), \alpha\in \{1,\dots,N\}$ (\citeauthor{Chiles} \citeyear{Chiles}; \citeauthor{emery2007conditioning} \citeyear{emery2007conditioning}).
We note that an equivalent construction for $ X^\textmd{SK}_\textmd{Error}(\textbf{u}) $ corresponds to
\[X^\textmd{SK}_\textmd{Error}(\textbf{u})=X^\textmd{SK}_\textmd{Obs}(\textbf{u})-X^\textmd{SK}_\textmd{SimObs}(\textbf{u}),\]
with $ X^\textmd{SK}_\textmd{Obs}(\textbf{u}) $ the interpolated field using the set $\{x_\alpha\}^N_{\alpha=1}$, and $ X^\textmd{SK}_\textmd{SimObs}(\textbf{u}) $ the interpolated field using the set $\{X_\textmd{S}(\textbf{u}_\alpha)\}^N_{\alpha=1}$. However, this conditioning procedure can not be carried out directly to the sphere case as sum operations of vectors do not lie on it.

Notice that the following alternative, weighted average view, can be considered for the conditioning procedure
\[X_\textmd{CS}(\textbf{u})=\sum_{{\bar{\alpha}}=1}^{2N+1}\bar{\lambda}_{\bar{\alpha}}(\textbf{u})\bar{X}_{\bar{\alpha}}(\textbf{u}),\]
with $ \bar{\lambda}_{\bar{\alpha}}(\textbf{u}) $ and $ \bar{X}_{\bar{\alpha}}(\textbf{u}) $ the entries of the vectors
\[\bar{\varvec\lambda}(\textbf{u})=
\begin{bmatrix}
1\\
\lambda_{1}^{\textmd{SK}}(\textbf{u})\\
\lambda_{2}^{\textmd{SK}}(\textbf{u})\\
\vdots\\
\lambda_{N}^{\textmd{SK}}(\textbf{u})\\
-\lambda_{1}^{\textmd{SK}}(\textbf{u})\\
-\lambda_{2}^{\textmd{SK}}(\textbf{u})\\
\vdots\\
-\lambda_{N}^{\textmd{SK}}(\textbf{u})\\
\end{bmatrix}\textmd{ and}\qquad
\bar{\textbf{X}}(\textbf{u})=
\begin{bmatrix}
X_\textmd{S}(\textbf{u})\\
x_1\\
x_2\\
\vdots\\
x_N\\
X_\textmd{S}(\textbf{u}_1)\\
X_\textmd{S}(\textbf{u}_2)\\
\vdots\\
X_\textmd{S}(\textbf{u}_N)\\
\end{bmatrix},
\]
respectively. We take this viewpoint to build, from the set of true observations $\{\textbf{s}_\alpha\}^N_{\alpha=1}$ and simulated values $\{\textbf{S}_\textmd{S}(\textbf{u}_\alpha)\}^N_{\alpha=1}$, the conditional field in the spherical case, $ \textbf{S}_\textmd{CS}(\textbf{u}) $,
\begin{equation}
	\textbf{S}_\textmd{CS}(\textbf{u}) = \underset{\substack{{\varvec{\mu}} \in \S^{p-1}}}{\textmd{argmin}} \sum_{\bar{\alpha}=1}^{2N+1} \bar{\lambda}_{\bar{\alpha}}(\textbf{u}) d^2\big({\varvec{\mu}},\bar{\textbf{S}}_{\bar{\alpha}}(\textbf{u})\big),\nonumber
	\label{fremeankrig}
\end{equation}
with $ \bar{\textbf{S}}_{\bar{\alpha}}(\textbf{u}) $ the columns of the array
\[
\bar{\textbf{S}}(\textbf{u})=
\begin{bmatrix}
\textbf{S}_\textmd{S}(\textbf{u})\quad\textbf{s}_1\quad
\textbf{s}_2\quad\cdots\quad\textbf{s}_N\quad \textbf{S}_\textmd{S}(\textbf{u}_1)\quad \textbf{S}_\textmd{S}(\textbf{u}_2)\quad\cdots\quad
\textbf{S}_\textmd{S}(\textbf{u}_N)
\end{bmatrix}.
\]

\begin{figure}[htbp]
\centering
(a)\includegraphics[width=0.65\textwidth,trim={0.0cm 0cm 0cm 0cm},clip,valign=m]{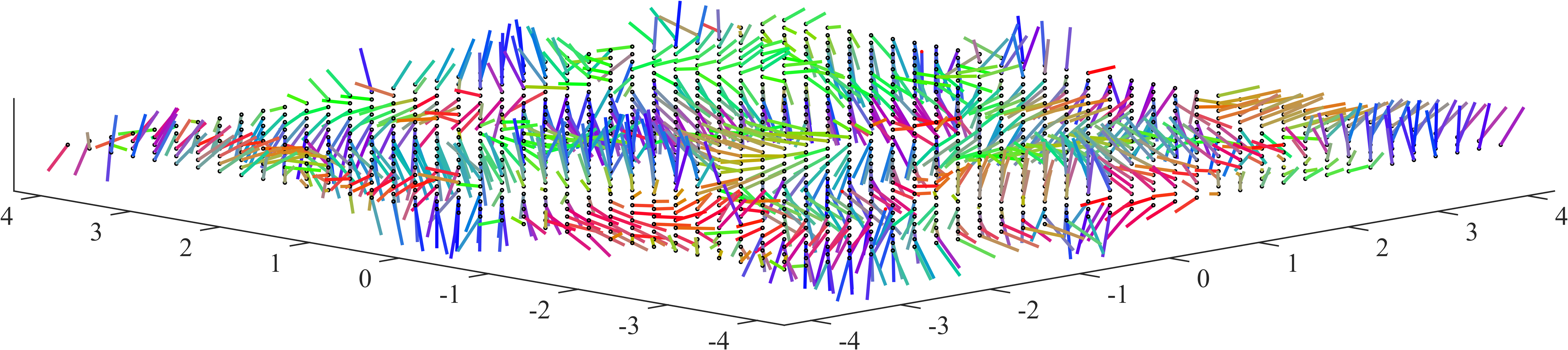}
\includegraphics[width=0.25\textwidth,trim={0.0cm 0cm 0cm 0cm},clip,valign=m]{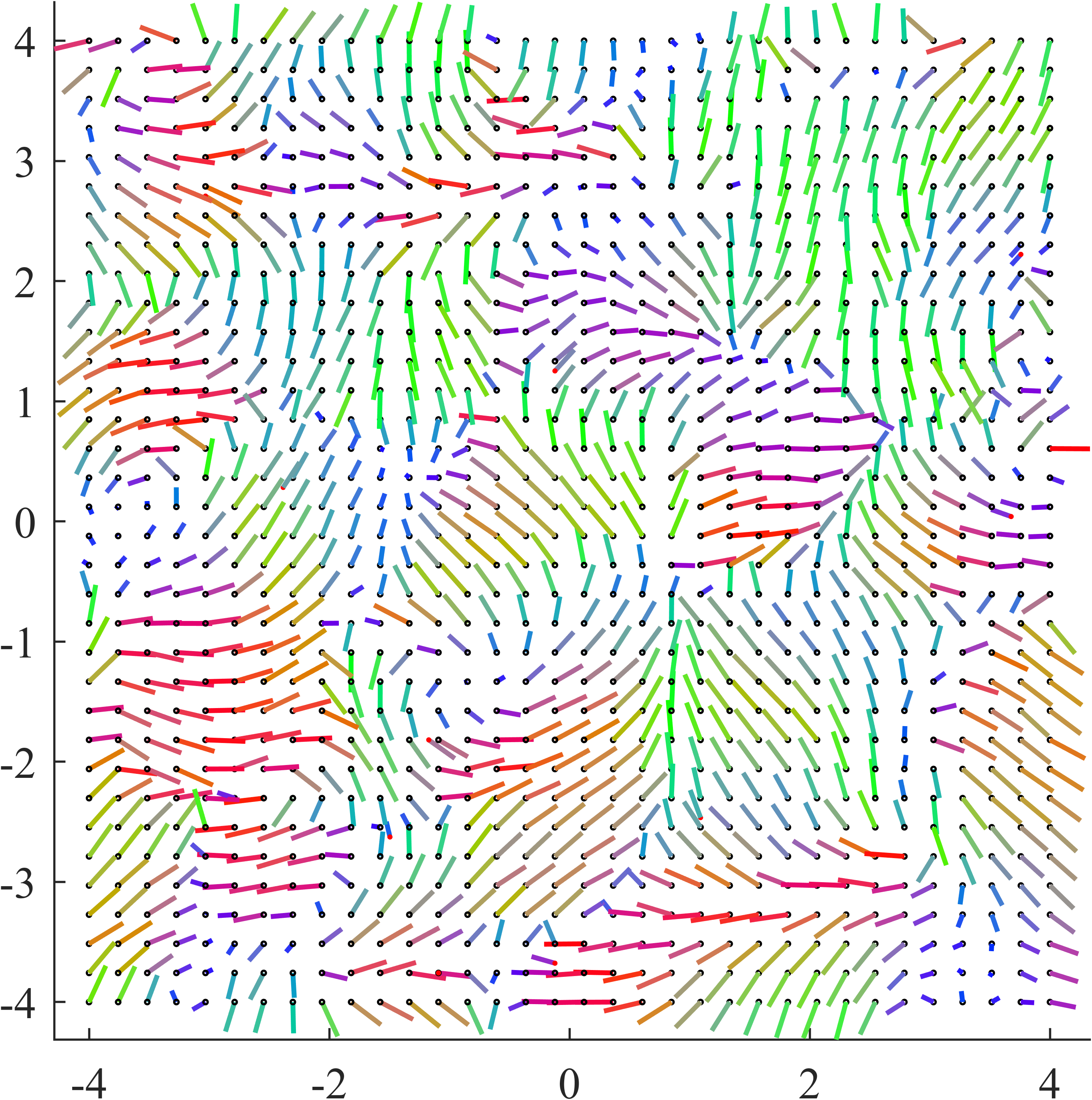}
(b)\includegraphics[width=0.65\textwidth,trim={0.0cm 0cm 0cm 0cm},clip,valign=m]{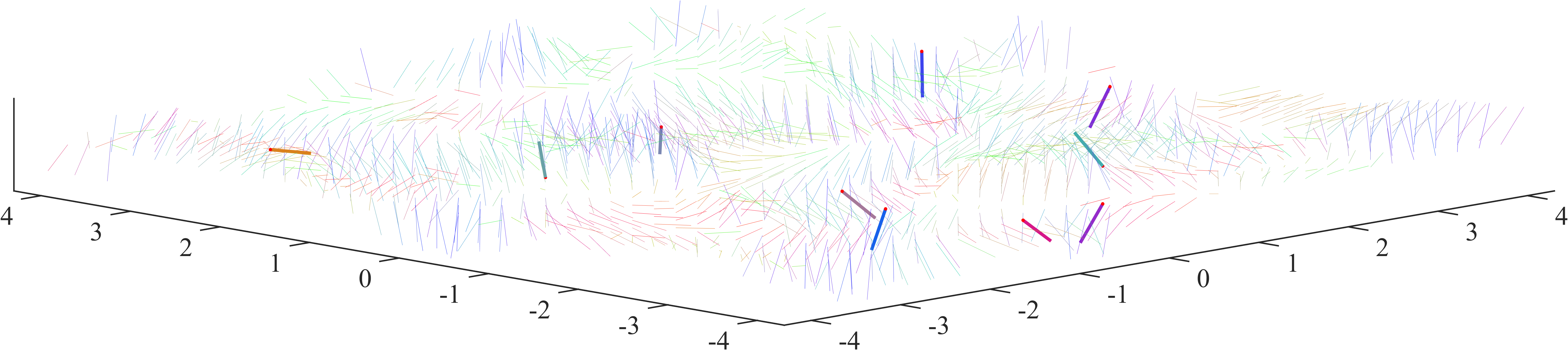}
\includegraphics[width=0.25\textwidth,trim={0.0cm 0cm 0cm 0cm},clip,valign=m]{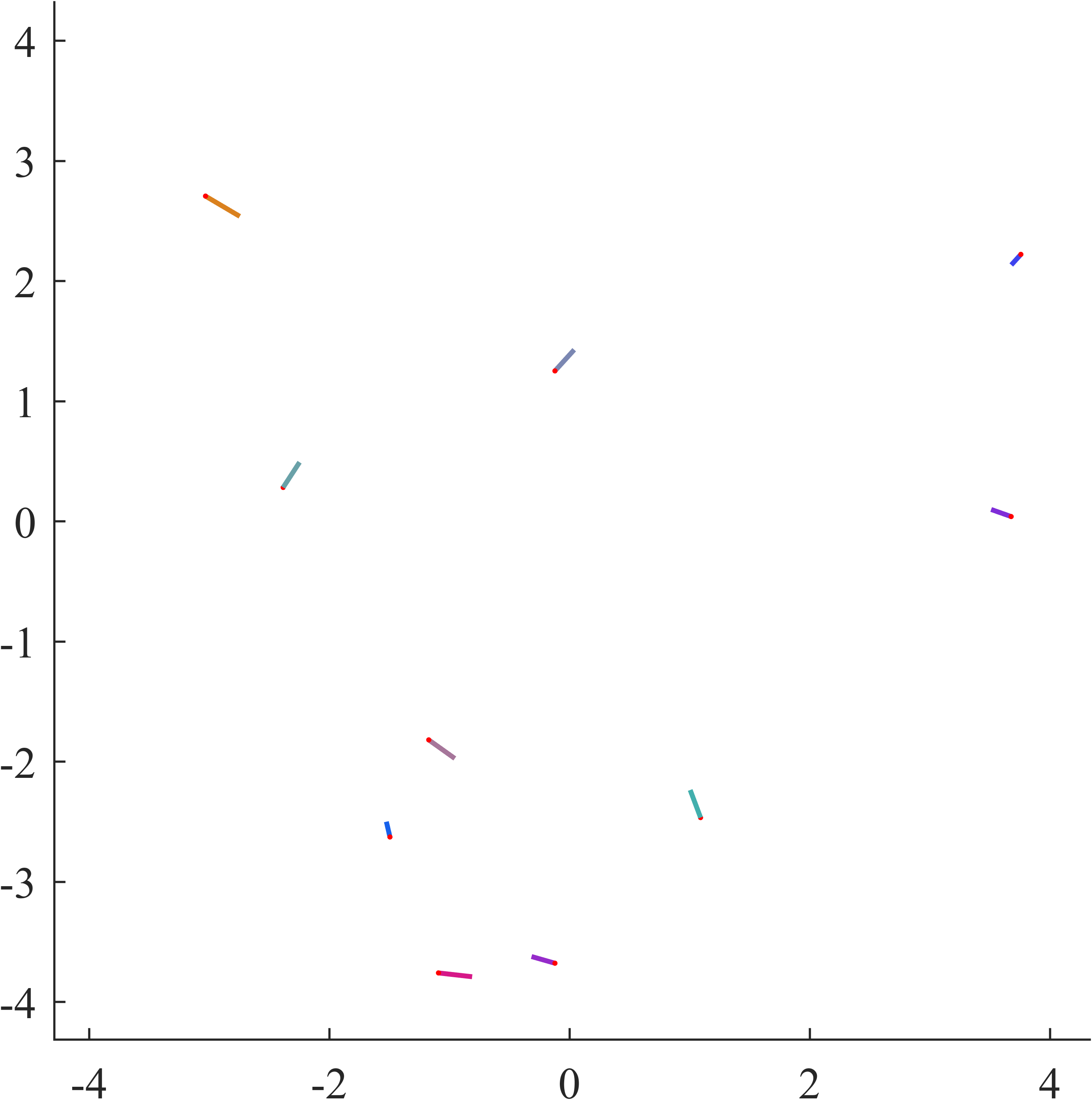}
(c)\includegraphics[width=0.65\textwidth,trim={0.0cm 0cm 0cm 0cm},clip,valign=m]{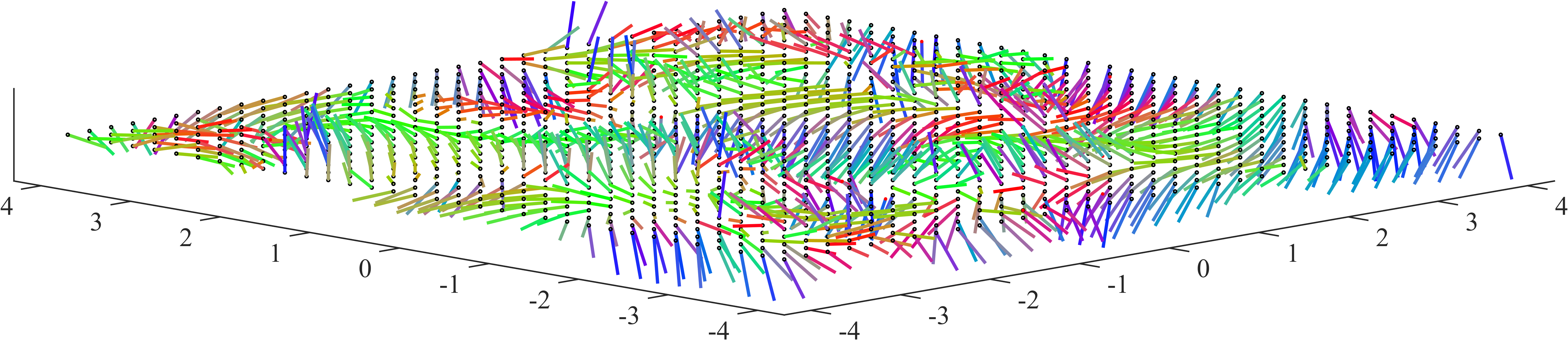}
\includegraphics[width=0.25\textwidth,trim={0.0cm 0cm 0cm 0cm},clip,valign=m]{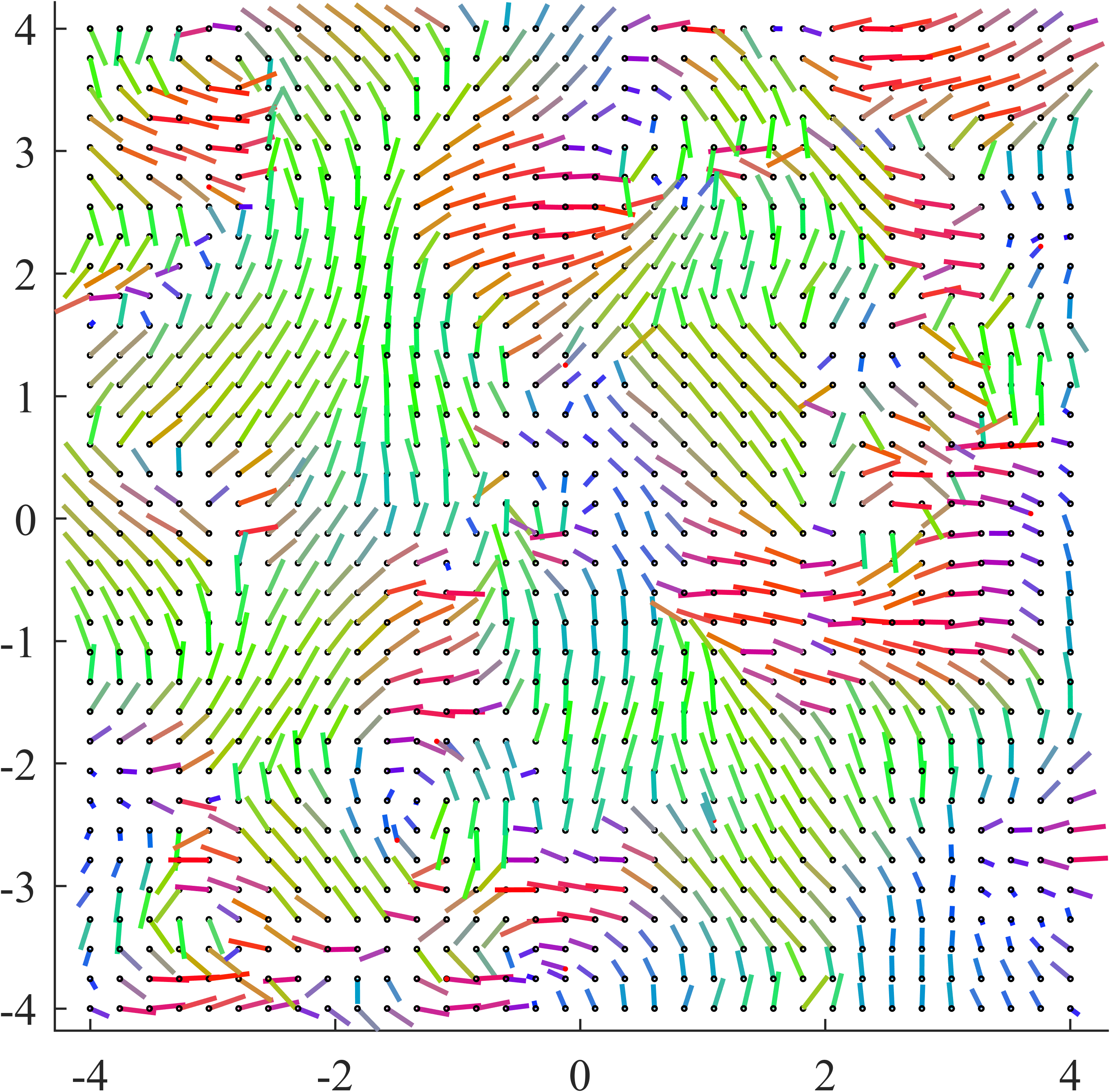}
(d)\includegraphics[width=0.65\textwidth,trim={0.0cm 0cm 0cm 0cm},clip,valign=m]{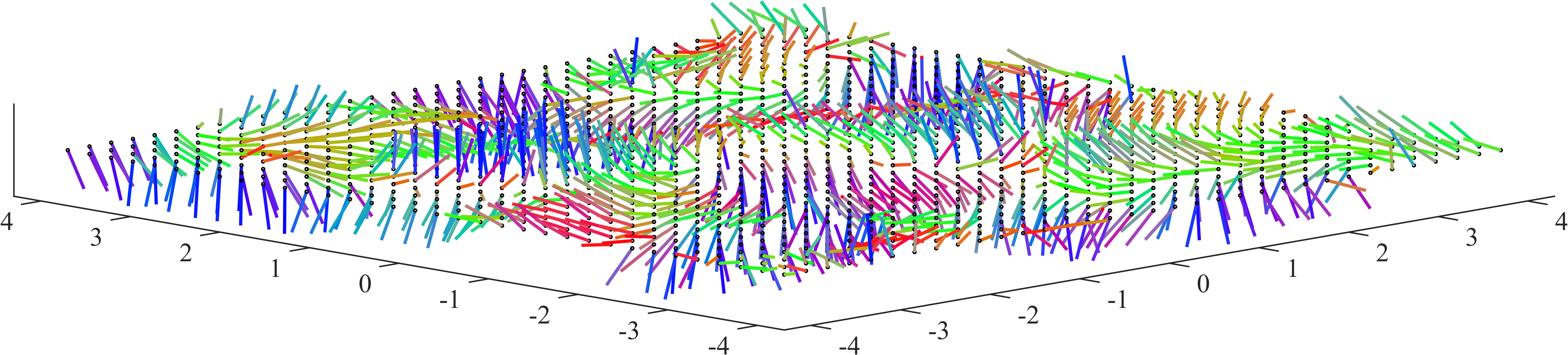}
\includegraphics[width=0.25\textwidth,trim={0.0cm 0cm 0cm 0cm},clip,valign=m]{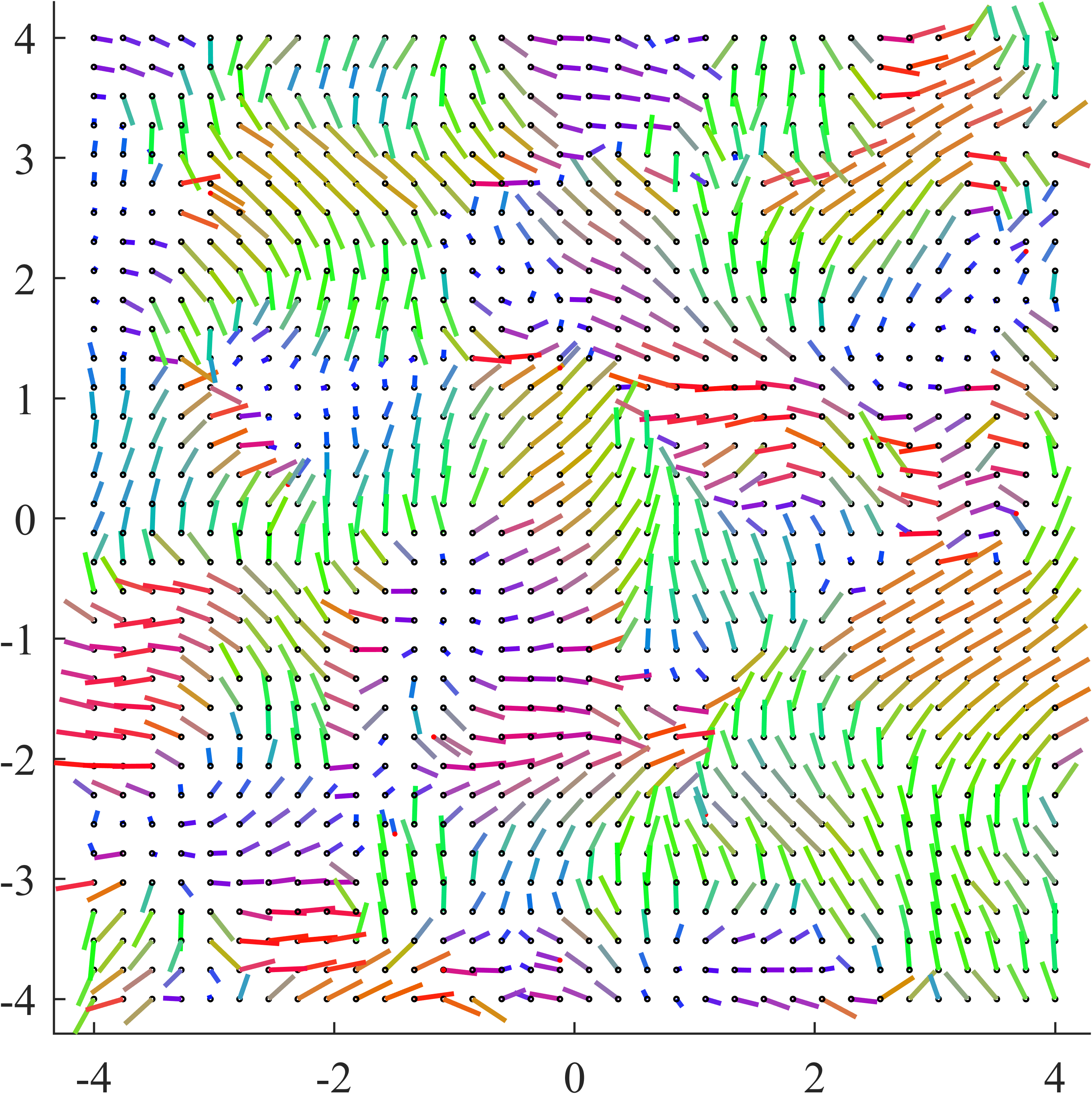}
\tiny{$C_{\S}(\textbf{h})$}\includegraphics[width=0.5\textwidth,trim={0.45cm 0cm 0cm 0cm},clip,valign=t]{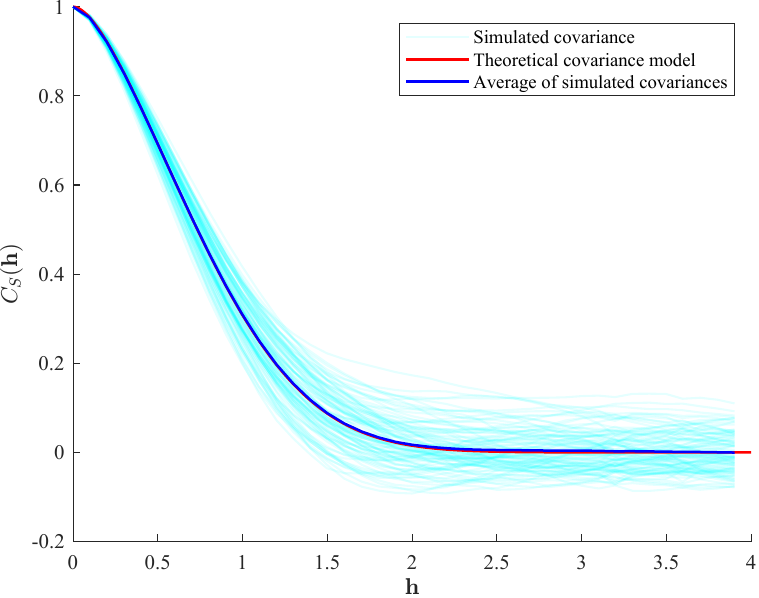}
\caption{Realizations of a three-dimensional orientation process (or $ \S^{2} $-RF) drawn over a two-dimensional grid with size 100 $ \times $ 100. (a) A non-conditional simulation; (b) the conditioning data points, and two conditional simulations (c, d). Bottom, sample covariance for a set of 100 non-conditional realizations; their average and the theoretical model.}
\label{condsim}
\end{figure}

The set of weights is determined from the solution of the simple kriging system, using the $ C_\S(\textbf{h}) $ as the covariance function (\citealt{van2002kriging})
$$
\begin{bmatrix}
1 & C_{\S}(\textbf{u}_1-\textbf{u}_2) & \dots  & C_{\S}(\textbf{u}_1-\textbf{u}_N)\\
C_{\S}(\textbf{u}_2-\textbf{u}_1) & 1& \dots  & C_{\S}(\textbf{u}_2-\textbf{u}_N)\\
\vdots & \vdots & \ddots & \vdots \\
C_{\S}(\textbf{u}_N-\textbf{u}_1) & C_{\S}(\textbf{u}_N-\textbf{u}_2)  & \dots  & 1\\
\end{bmatrix}
\cdot
\begin{bmatrix}
\lambda_{1}^{\textmd{SK}}(\textbf{u})\\
\lambda_{2}^{\textmd{SK}}(\textbf{u})\\
\vdots\\
\lambda_{N}^{\textmd{SK}}(\textbf{u})\\
\end{bmatrix}
=
\begin{bmatrix}
C_{\S}(\textbf{u}-\textbf{u}_1)\\
C_{\S}(\textbf{u}-\textbf{u}_2)\\
\vdots\\
C_{\S}(\textbf{u}-\textbf{u}_N)\\
\end{bmatrix}.$$
An illustration is given in Fig. \ref{condsim}, which
considers a two-dimensional simulation domain and conditioning data uniformly distributed across it.

\subsection{Dealing with arbitrary distributions on \texorpdfstring{$ \S^{p-1} $}{Lg}}
\label{ar2sphe}

When working with regionalized orientations, it is likely that these do not follow a uniform distribution but rather complex, intricate ones. Therefore, a procedure is required to map a uniform PDF to an arbitrary PDF and vice-versa in order to make our methodology functional for the spatial analysis of more general cases. We split the problem into two steps: (\textit{i})
first, we map an arbitrary distribution into a spherical Gaussian distribution, and (\textit{ii}) second, we translate the spherical Gaussian distribution into a uniform PDF on $ \S^{p-1} $.

\subsubsection{Mapping an arbitrary distribution into a  Gaussian distribution on \texorpdfstring{$ \S^{p-1} $}{Lg}}

Given an arbitrary distribution of points $ \textbf{s}_1,\dots, \textbf{s}_N \in \S^{p-1}$, we start by finding the geometric mean of the set, $ \varvec{\mu}$ (Eq. \ref{freme}). (Hereafter, we assume without loss of generality that $ \varvec{\mu}$ points in the ``north-pole'' direction $ (0,\dots,0,1)^T \in \R^p $ as the complete distribution can be carried out through parallel transport (Eq. \ref{partras}) to this point and vice versa.) Then, we apply the logarithmic map to obtain the set of vectors $ \textbf{v}_\alpha = \textmd{Log}_\varvec{\mu}({\textbf{s}_\alpha}), \alpha \in{1,\dots,N} \in T_\varvec{\mu}\S^{p-1}$.

Since the tangent space at $ \varvec{\mu}$ is $ \R^{p-1} $ itself, it is possible to apply any preferred multi-Gaussian transform to map the distribution into a Gaussian distribution on $ \R^{p-1} $. We consider here the use of Projection Pursuit Multivariate Transform (PPMT; \citeauthor{barnett2014projection} \citeyear{barnett2014projection}) for this purpose (other valid alternatives are reviewed and compared in \citealt{abulkhair2023geostatistics}).

PPMT transforms a vector of variables in $ \R^{p-1} $ individually into normal scores through quantile matching, followed by iterative Gaussianization along the direction that maximizes a projection index (\citealt{friedman1987exploratory}). The procedure allows us to obtain a new set of vectors in the tangent space, $ \{\tilde{\textbf{v}}_\alpha\}^N_{\alpha=1} \in T_\varvec{\mu}\S^{p-1}$, following an uncorrelated multivariate Gaussian PDF. Figure illustrates the initial setting and a few steps of the PPMT procedure.

\begin{figure}[htbp]
\includegraphics[width=1\textwidth,trim={0.5cm 0cm 2cm 0cm},clip]{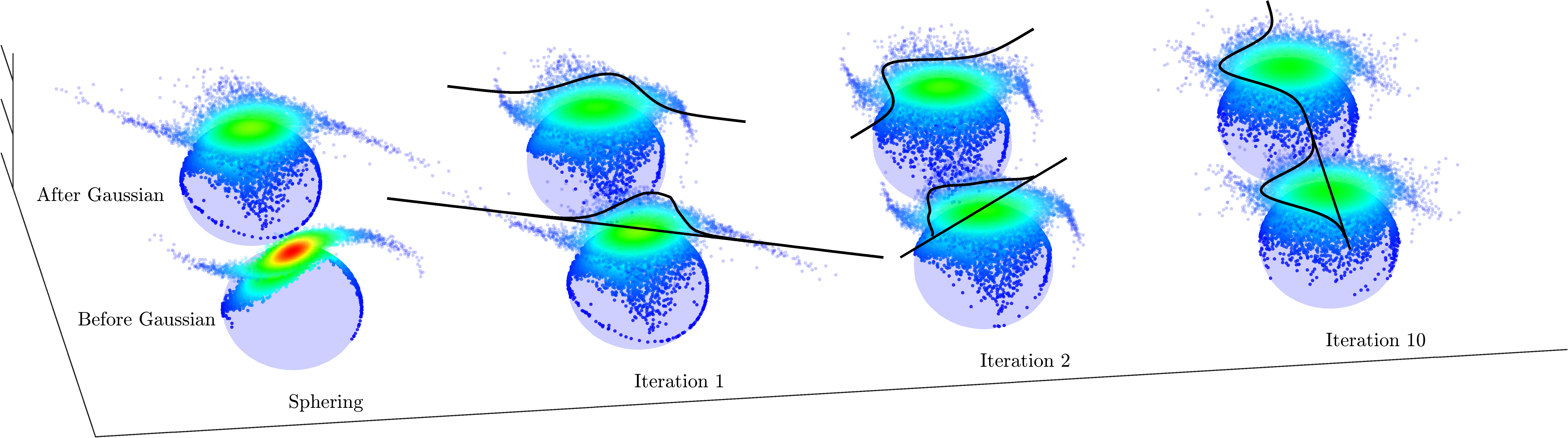}
\caption{Application of PPMT to an arbitrary distribution of points on the sphere. The data is mapped into $ T_\varvec{\mu}\S^{p-1}$ prior to run the Gaussianization procedure.}
\label{ppmt}
\end{figure}

\subsubsection{Mapping a Gaussian distribution into a Uniform distribution on \texorpdfstring{$ \S^{p-1} $}{Lg}}

Mapping of a multivariate Gaussian PDF on $T_\varvec{\mu}\S^{p-1}$ into $ \S^{p-1} $ through the exponential map corresponds to the definition of spherical normal (SN) distribution (\citealt{pennec2006intrinsic}), with a probability density given by
\begin{align}
f_{\textmd{SN}^{}_{\S^{p-1}}}(\textbf{x}|\varvec{\mu},\kappa) & \propto \exp\Big(-\frac{\kappa}{2}\arccos^2\big(\langle \textbf{x},\varvec{\mu} \rangle\big)\Big)\nonumber\\
& =\frac{1}{C_{p-1}(\varvec{\mu},\kappa)}\exp\Big(-\frac{\kappa}{2}\arccos^2\big(\langle \textbf{x},\varvec{\mu} \rangle\big)\Big)\nonumber
\end{align}
for a point $ \textbf{x} \in \S^{p-1} $, a concentration parameter $ \kappa $, and a normalizing constant (\citealt{hauberg2018directional})
\[C_{p-1}(\varvec{\mu},\kappa)=\int_{\S^{p-1}}\exp\Big(-\frac{\kappa}{2}\arccos^2\big(\langle \textbf{x},\varvec{\mu} \rangle\big)\Big)\textmd{d}\textbf{x}.\]
The identity \[ \arccos^2\big(\langle \textbf{x},\varvec{\mu} \rangle\big)={\textmd{Log}_\varvec{\mu}(\textbf{x})}^T{\textmd{Log}_\varvec{\mu}(\textbf{x})}\]
allows us to verify that this distribution corresponds to the projection of a normal distribution into the sphere. The case of a standard multivariate Gaussian PDF on $T_\varvec{\mu}\S^{p-1}$ coincides with the case on which  $ \kappa=1 $.

In order to map this spherical normal distribution into a uniform distribution, we consider a quantile matching procedure (anamorphosis) on the sphere, noticing that both distributions are radially symmetric with respect to the north pole. This consideration allows us to work equivalently with (\textit{i}) the elevation coordinate with respect to the north pole (the last entry $ s_n \in [-1,1] $, given $ \textbf{s} \in \S^{p-1} $) or just $ s $; (\textit{ii}) in the sphere, through arc distances $\theta=\arccos(s)$ from the north pole; or (\textit{iii}) in the tangent space, through a radial vector $ \textbf{r} $ with length $ \theta $.

The cumulative distribution function (CDF) of a uniform distribution on the sphere is given by (\citealt{hotelling1953new})
\begin{align}
\textmd{CDF}_{\textmd{Uniform}^{}_{\S^{p-1}}}(s)&=\frac{A_{p-2}}{A_{p-1}}\int_{s}^{1}(1-t^2)^{(p-3)/2}\textmd{d}t\nonumber\\
&=\frac{1}{2}-\frac{\upGamma\big(\frac{p}{2}\big)}{\pi^{\frac{1}{2}}\upGamma\big(\frac{p-1}{2}\big)}\Bigg(s\cdot{}_{2}{F}_{1}\left\lparen\tfrac{1}{2},\tfrac{3-p}{2};\tfrac{3}{2};s^2\right\rparen\Bigg),\nonumber
\end{align}
or, in terms of a radial vector $ \textbf{r} \in T_\varvec{\mu}\S^{p-1} $, by
\begin{align}
\textmd{CDF}_{\textmd{Uniform}^{}_{\S^{p-1}}}(\textbf{r})=\frac{1}{2}-\frac{\upGamma\big(\frac{p}{2}\big)}{\pi^{\frac{1}{2}}\upGamma\big(\frac{p-1}{2}\big)}\Bigg(\cos(\norm{\textbf{r}})\cdot{}_{2}{F}_{1}\left\lparen\tfrac{1}{2},\tfrac{3-p}{2};\tfrac{3}{2};\cos^2(\norm{\textbf{r}})\right\rparen\Bigg).\nonumber
\end{align}

Similarly, the CDF for the spherical normal distribution can be computed, for a radial vector $ \textbf{r} \in T_\varvec{\mu}\S^{p-1} $, by
\[\textmd{CDF}_{\textmd{SN}^{}_{\S^{p-1}}}(\textbf{r})=\frac{1}{C_{p-1}(\varvec{\mu},\kappa)}\int_{\norm{\textbf{v}}\leq\norm{\textbf{r}}}\exp\Big(-\frac{\kappa}{2}\norm{\textbf{v}}^2\Big)\textmd{det}(\textbf{J})\textmd{d}\textbf{v},\]
where $ \textmd{det}(\textbf{J}) = \sin(\norm{\textbf{v}})/\norm{\textbf{v}} $ denotes the
Jacobian of $ \textmd{Exp}_{\varvec{\mu}}(\textbf{v}) $.
Both the normalizing constant $ {C_{p-1}(\varvec{\mu},\kappa)} $ and the CDF can be computed numerically, and even in a closed expression (we refer to the supplementary material in \cite{hauberg2018directional} as the procedure is long and tedious).

\begin{figure}[htbp]
\includegraphics[width=1\textwidth,trim={0cm 0cm 0cm 0cm},clip]{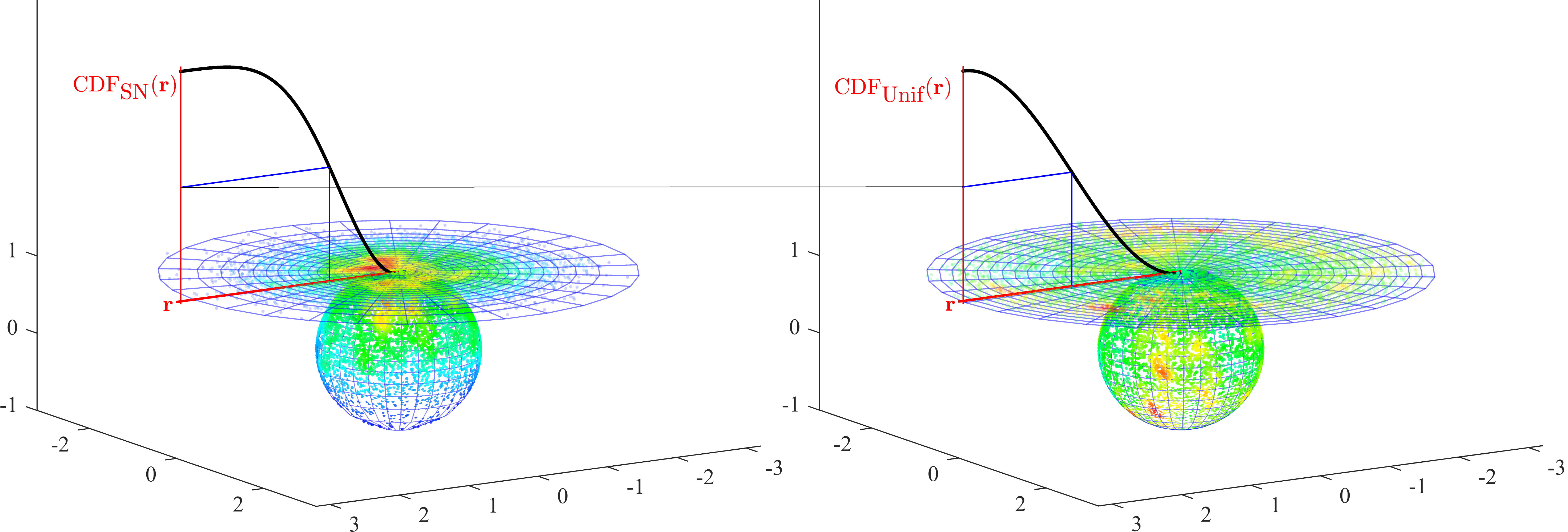}
\caption{Anamorphosis on the sphere. A SN distribution is mapped into a uniform one via quantile matching of CDFs on the tangent space, as these are radially symmetric.}
\label{ansphere}
\end{figure}

Once both CDFs are established, any element $ \textbf{r} \in T_\varvec{\mu}\S^{p-1} $ following a SN distribution can be shifted radially into a new vector $ \textbf{r}' $ belonging to a uniform distribution (Fig. \ref{ansphere}), based on the law of conservation of probability
\[\textmd{CDF}_{\textmd{SN}^{}_{\S^{p-1}}}(\textbf{r})=\textmd{CDF}_{\textmd{Uniform}^{}_{\S^{p-1}}}(\textbf{r}') \implies \textbf{r}'=\textmd{CDF}^{-1}_{\textmd{Uniform}^{}_{\S^{p-1}}}\circ\textmd{CDF}_{\textmd{SN}^{}_{\S^{p-1}}}(\textbf{r}).\]
Anamorphosis of the sphere can be used as an alternative to rejection sampling methods in the drawing of values following a SN distribution (\citealt{hauberg2018directional}; \citealt{you2023spherical}).

\section{Real Case Study}
\label{Real}
The methodology is employed in a real case study using publicly available data, with the aim of identifying potential Nickel-rich areas. The area of study corresponds to the Yilgarn Craton, Western Australia, known for its significant Nickel reserves. Overall, the MPM problem in the Australian case is highly relevant as less than 20\% of the territory is exposed rock, while the remaining fraction of the land is covered \citep{wilford2016regolith}. Simultaneously, MPM methods find in the Australian case an important opportunity for being implemented and tested, due to the vast and varied data from numerous sources across the country, much of which is publicly accessible. We refer to the studies made by \cite{porwal2010introduction} and \cite{gonccalves2024mineral} as alternative MPM studies in the same area and for further details of the geological context of the Yilgarn Craton.

\subsection{Study Area and Dataset}
\label{Study Area}

\begin{figure}[htp!]
\centering
\includegraphics[width=0.45\textwidth,trim={0.0cm 0cm 0.0cm 0cm},clip]{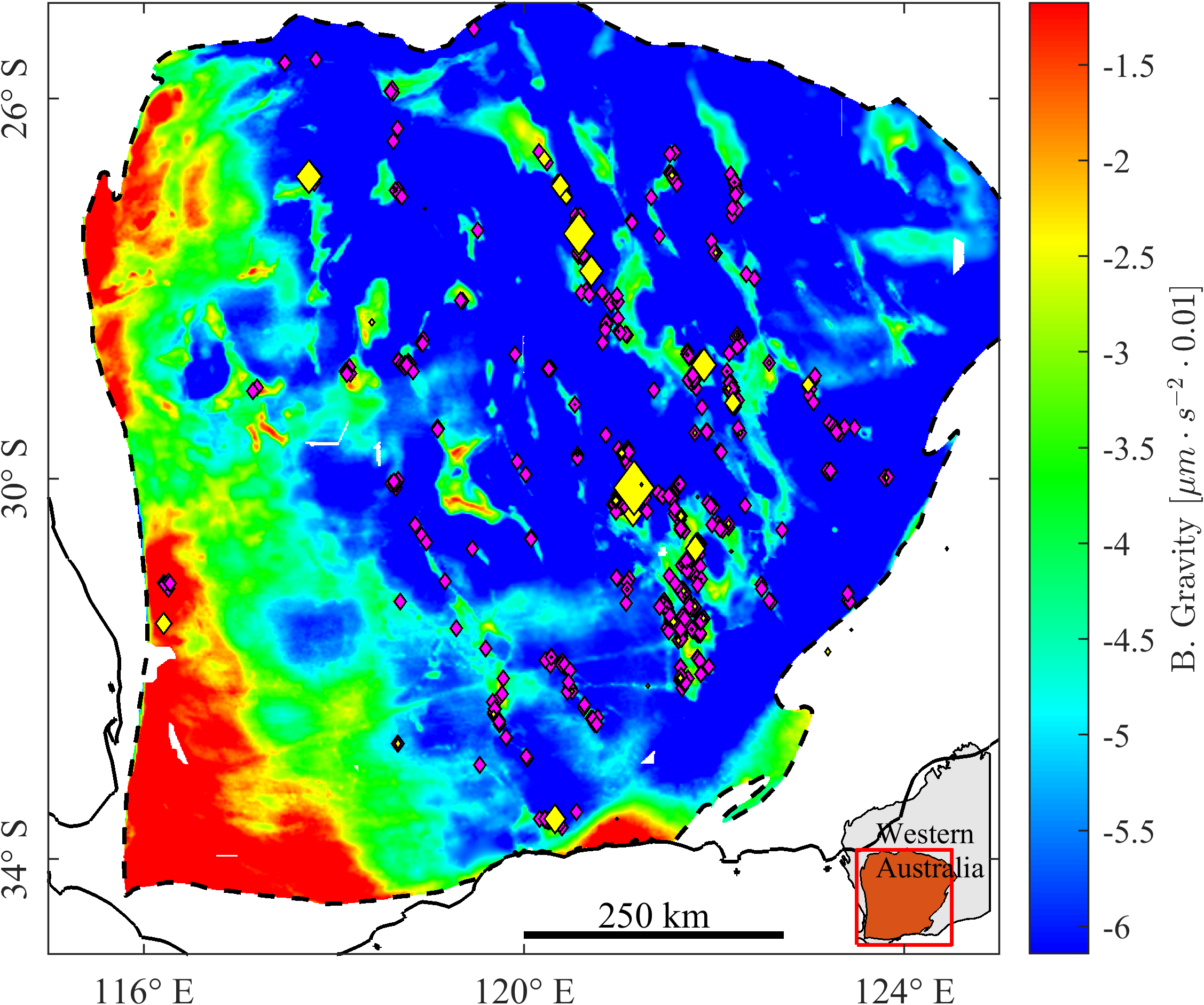}
\includegraphics[width=0.45\textwidth,trim={0.0cm 0cm 0.0cm 0cm},clip]{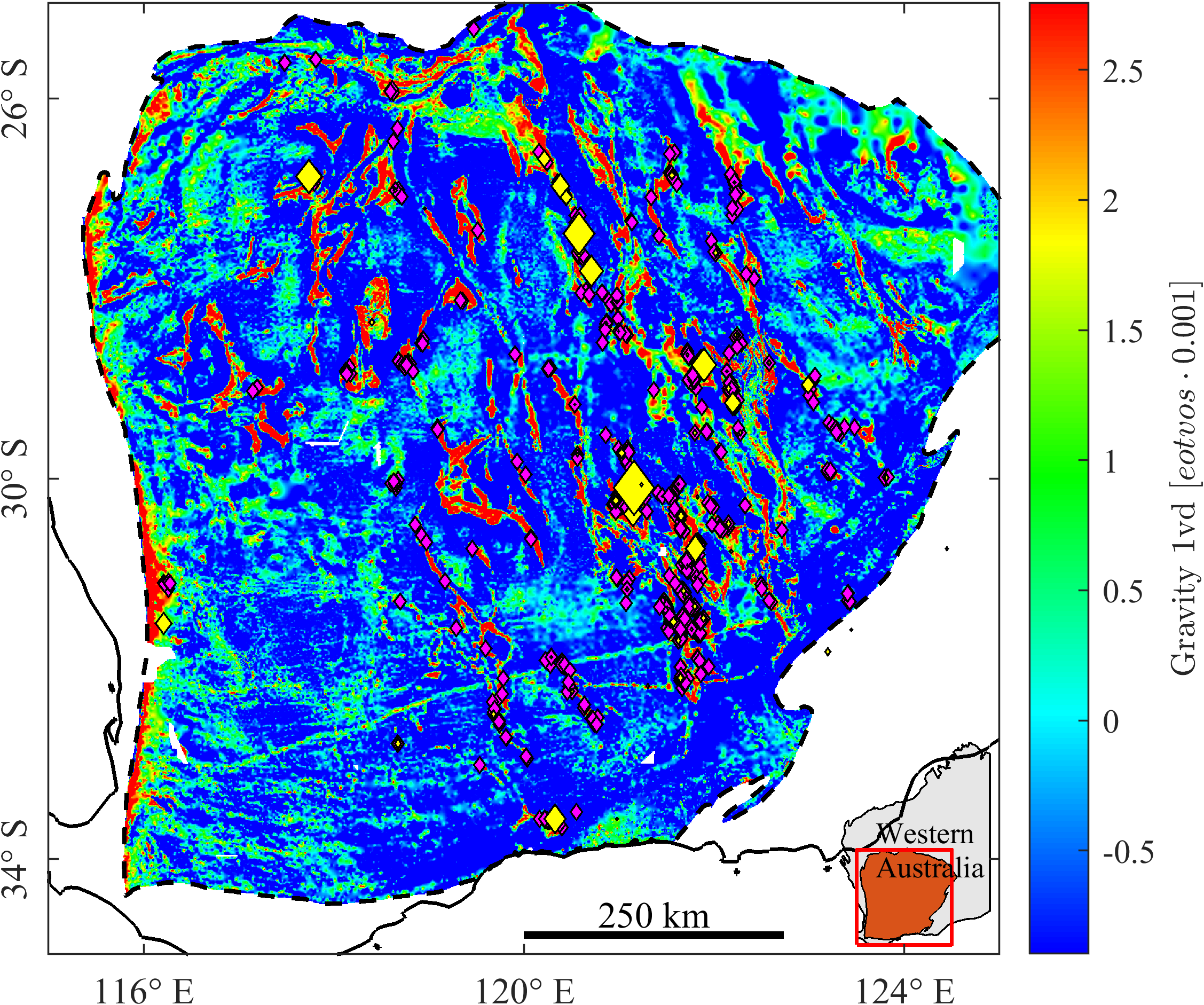}
\includegraphics[width=0.45\textwidth,trim={0.0cm 0cm 0.0cm 0cm},clip]{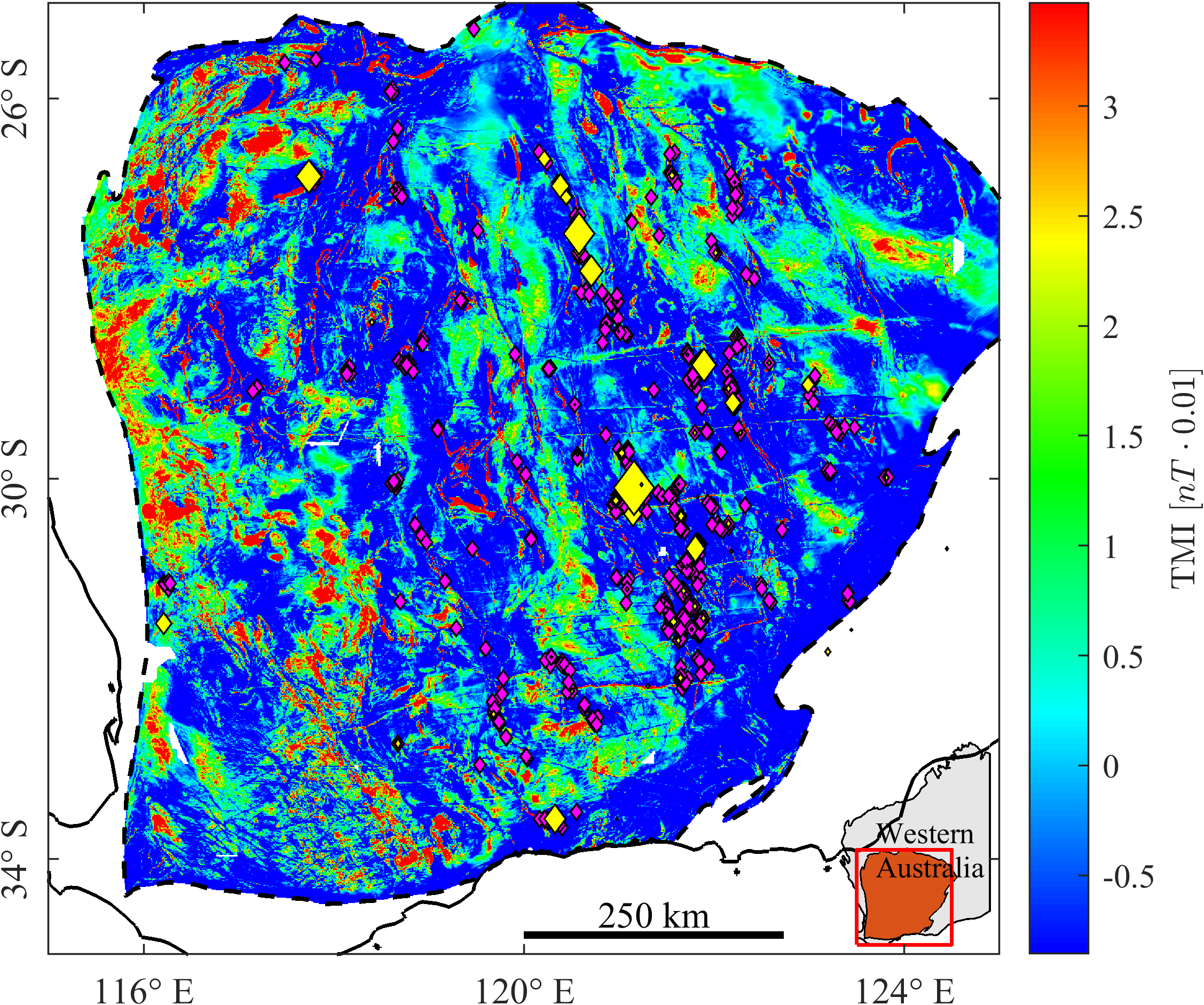}
\includegraphics[width=0.45\textwidth,trim={0.0cm 0cm 0.0cm 0cm},clip]{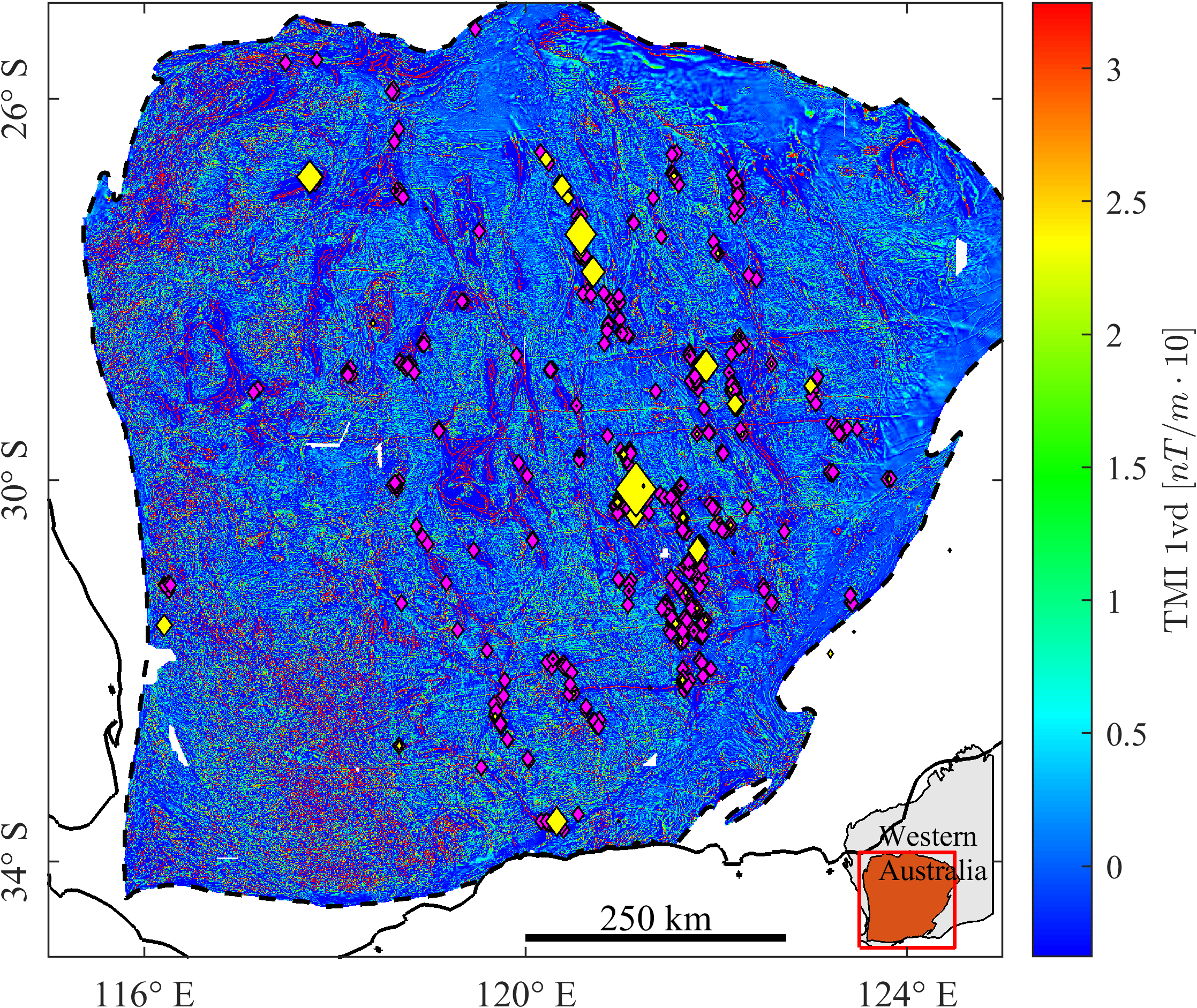}
\includegraphics[width=0.45\textwidth,trim={0.0cm 0cm 0.0cm 0cm},clip]{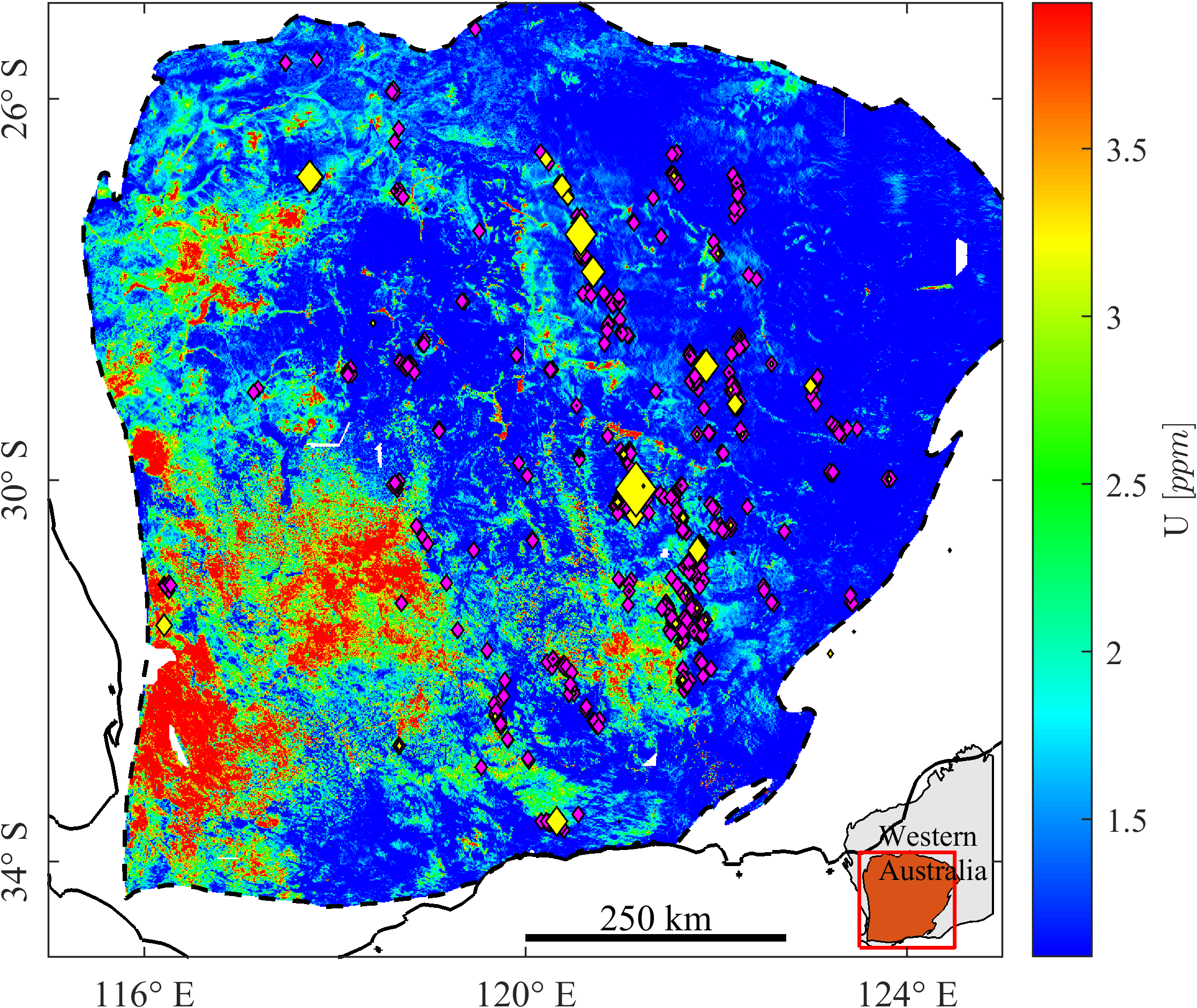}
\includegraphics[width=0.45\textwidth,trim={0.0cm 0cm 0.0cm 0cm},clip]{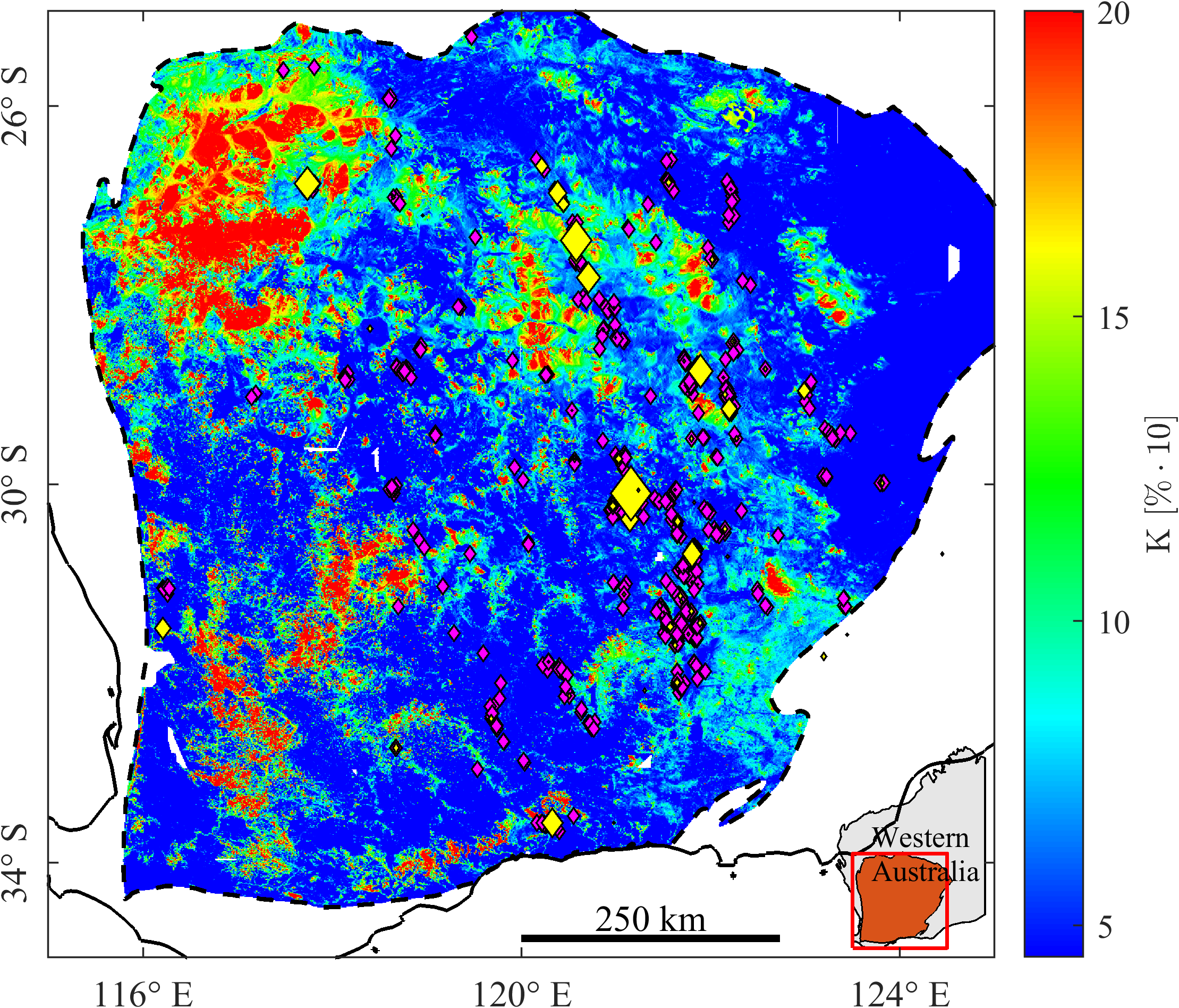}
\includegraphics[width=0.45\textwidth,trim={0.0cm 0cm 0.0cm 0cm},clip]{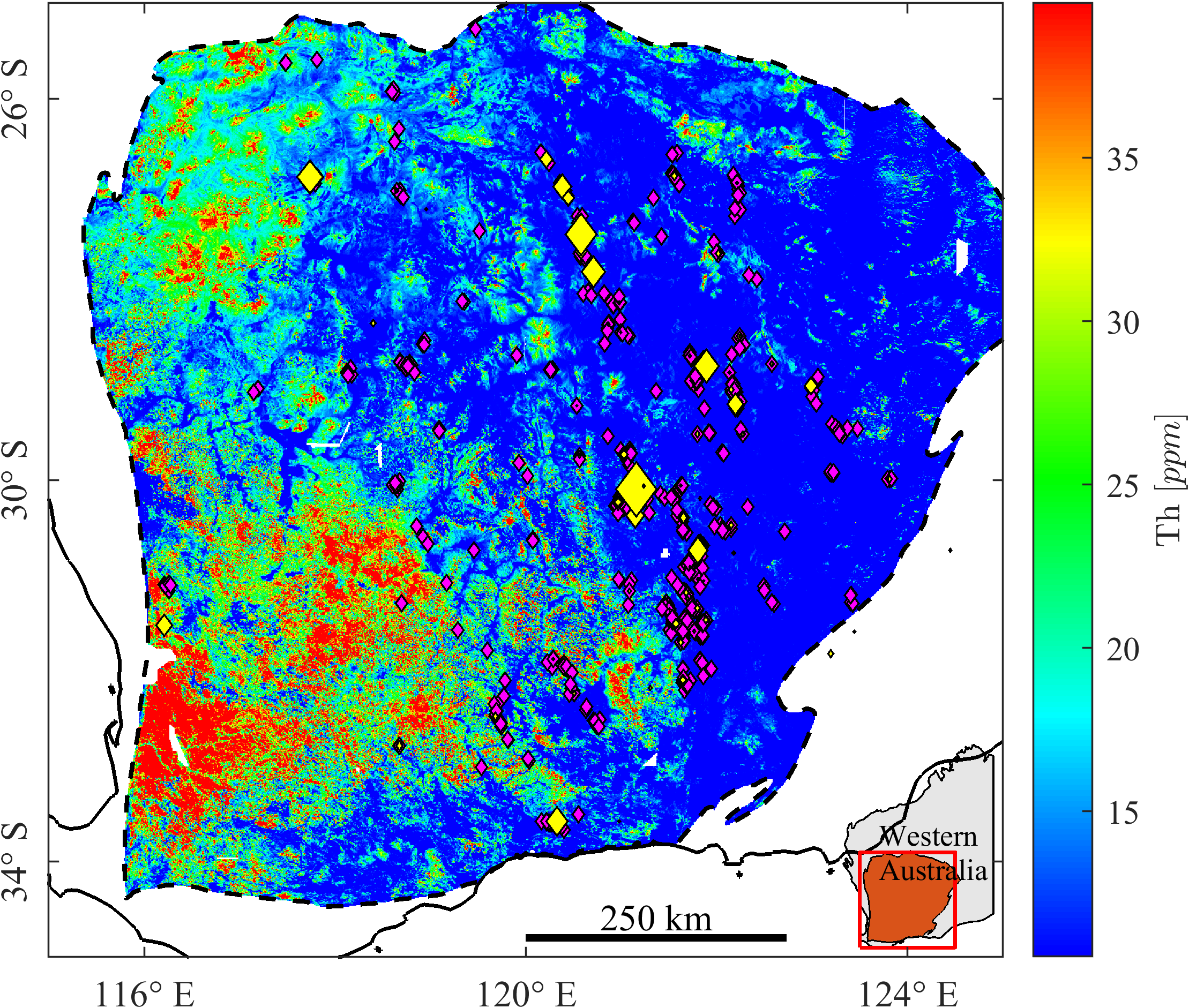}
\includegraphics[width=0.45\textwidth,trim={0.0cm 0cm 0.0cm 0cm},clip]{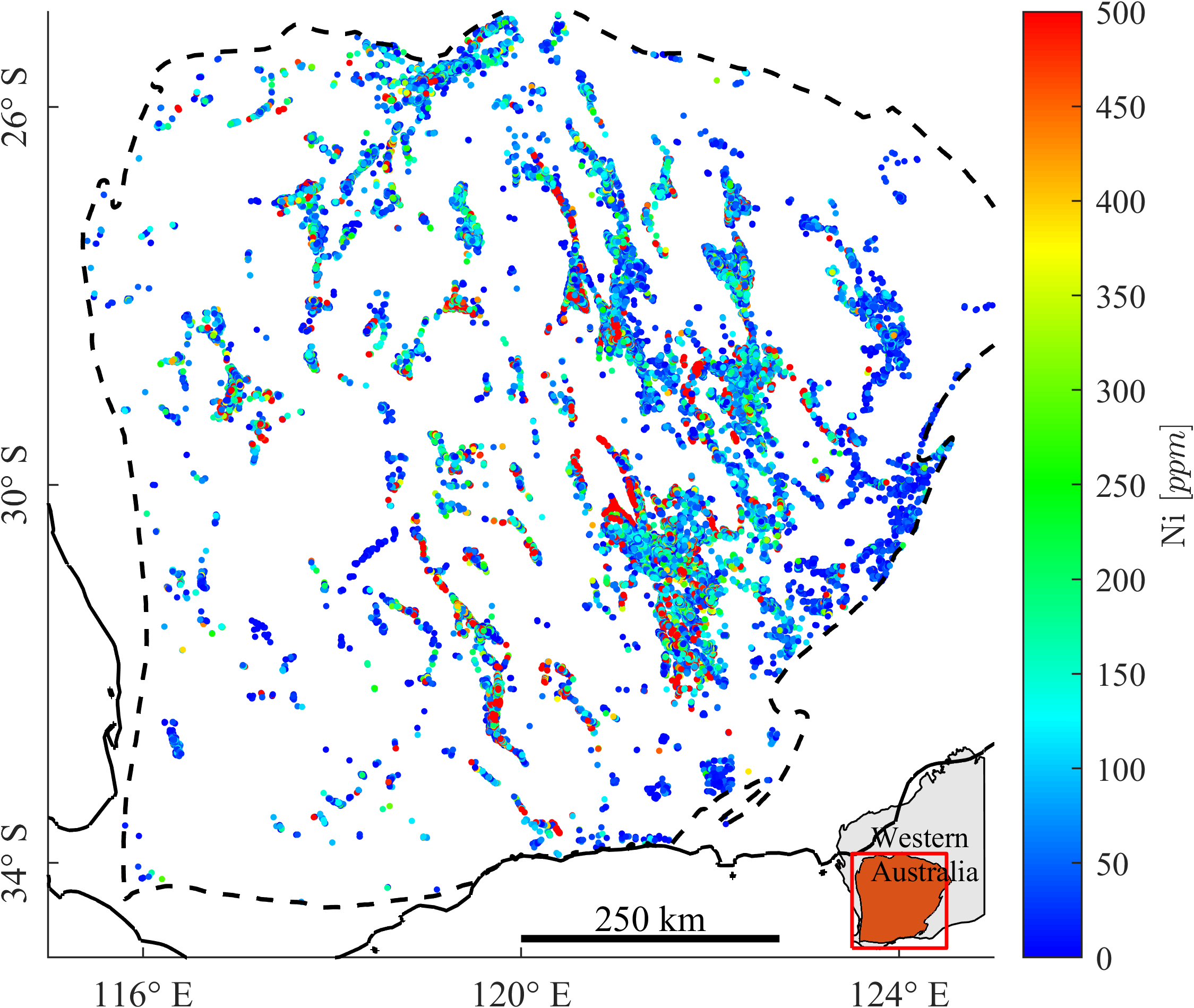}
\caption{Maps displaying the different data layers used in the case study, restricted to the Yilgran Craton area, Western Australia. The selected locations for calibration of models are shown in magenta-colored markers, and locations with known reserves are shown in yellow-colored markers.}
\label{data}
\end{figure}

Western Australia hosts most of the known nickel resources within Australia (around 90\% of Australian demonstrated resources; \citealt{Britt}) and contributes to positioning Australia among the countries with most of the world nickel reserves ($\sim 18.5 \%$; \citealt{usgeo}). The majority of significant deposits in Western Australia are found within the Yilgarn Craton. This area counts with a comprehensive repository of data of varying types and quality, which has been made accessible to the public to encourage additional mineral exploration and data gathering. In this study, we consider four types of datasets:
\begin{description}[leftmargin=0cm,itemsep=4pt]
    \item[\textbf{Input data features}$ \quad $] We consider the following 7 input features for training purposes, consisting of geophysical datasets in grid format: Bouguer gravity anomaly (or just B. gravity; \citealt{lane20202019}); first vertical derivative of the Bouguer anomaly (gravity 1vd; \citealt{1VDgrav}); total magnetic intensity (TMI; \citealt{Djomani}); first vertical derivative of the total magnetic intensity (TMI 1vd; \citealt{Djomani2}); and radiometric data for Potassium (K; \citealt{DjomaniK}), Thorium (Th; \citealt{DjomaniTh}), and Uranium (U; \citealt{DjomaniU}). The data can be directly accessed through the Geophysical Archive Data Delivery System (GADDS) on the Australian Government portal (\url{http://www.geoscience.gov.au/gadds}). These input layers have been built from exhaustive sampling data acquired both at ground and airborne levels, with differences in the sampling spacing reflected on their respective grid resolution: TMI data have a grid cell spacing of $\sim80$  m; gravity has a grid spacing of $\sim800$ m; and the grid spacing of radiometric data (K, Th, and U) is $\sim100$ m. The same input layers are used later to inquiring the simulated response models obtained at unexplored locations, with the purpose of determining the local mineral potential. Figure \ref{data} shows the values attained by these input variables and Tab. \ref{tab:becn} summarizes some basic statistics restricted to the area of interest for this case study. These 7 datasets compose the regionalized vector $\textbf{z}(\textbf{u})$ used later for evaluating a classification SVM RF.
    
    \item[\textbf{Response dataset}$ \quad $] A single outcome variable is considered in this case study, derived from a Nickel content variable (in ppm; Fig. \ref{data}) sourced from the Company Mineral Drillhole Database \citep{riganti2015125}, a dataset comprising more than 3 million historic drill holes and more than 9 million surface samples obtained from exploration companies' open-file reports. The dataset is available for download in SQL format at the Western Australia Exploration Geochemistry portal (\url{https://wamexgeochem.net.au}). It is important to note that no quality control has been carried out on the dataset values. Some pre-processing considerations on the dataset were taken, such as the removal of those samples within the first 50 meters in depth, an attribute included in the database as well. A thresholding in the Nickel values is taken in order to obtain a binary variable, by using the following rules: Nickel values below the 100 ppm limit are assigned a 0 value; Nickel values above the 500 ppm limit are assigned a 1 value; any Nickel value in between the 100 ppm and the 500 ppm limit is deleted from the database. This last consideration is taken to create a buffer area between the two classes, aiding the process of finding the classification boundary in the multivariate attribute space via SVM. After applying these considerations, the total amount of training geochemical samples in the area of study is reduced from $\sim8.5$ million samples to $\sim108$ thousand ($\sim27$ thousand of positively-classified samples and $\sim81$ thousand of negatively-classified samples).
    
    \item[\textbf{Training locations}$ \quad $] Relationships between input and output variables can be obtained and assigned at every single location counting with a geochemical sample, for which it is enough to consider a neighborhood of samples at the location of interest from where evaluating this correspondence. Since not every relationship is necessarily linked to the mineral potential of the variable of interest in this case study, we seek a way of discriminating among the set of available locations, looking only for those relationships of higher significance Nickel mineral potential. The selected mineralization sites of Western Australia database \cite{DMIRSNi} provide a set of locations on which some type of evidence associated with Nickel mineralization (and other commodities) has been recorded (plotted on the following maps as magenta-colored diamonds). The initial number of locations restricted to the area of study is 973. The adjustment of a classification model is then restricted only to this set of locations, illustrated throughout the figures in magenta-colored diamond markers. 
    
    \item[\textbf{Validation dataset}$ \quad $] In order to validate our mineral potential results, we consider a last dataset gathered from Western Australia's Minedex database \cite{DMIRS}. The database contains information related to Nickel reserves in Western Australia. The database is used to check the amount of resources unlocked when searching the area of interest, sorting the grid cells from higher to lower potential. Locations with known resources are illustrated throughout the figures in yellow-colored diamond markers, with varying sizes proportional to the resources. 
    
\end{description}

\renewcommand{\arraystretch}{1.5}
\begin{table}[htp]
\centering
\caption{Descriptive statistics of input features gathered at geochemical data locations within the study area.}
\label{tab:becn}
\resizebox{1\textwidth}{!}{
\begin{tabular}{lrrrrrrrrl}
& Mean  & Std   & Min     & 0.25   & 0.5    & 0.75   & Max & Kurt    &  \\ \cline{1-9}
B. Gravity & -5.13 & 1.43 & -11.11  & -10.34 & -10.11 & -9.96  & 1.43   & 4.8 & \\
Gravity 1vd & 1.34  & 2.5 & -12.04  & -5.71  & -4.88  & -4.5   & 16.59  & 3.7  & \\
TMI & 2.9   & 19.81 & -106.06 & -20.3  & -13.4  & -11.11 & 361.43 & 93.2 & \\
TMI 1vd & 6.03  & 45.9  & -438.83 & -91.61 & -65.01 & -50.83 & 1,494.4 & 165.6 & \\
U & 1.3   & 0.67  & 0 & 0.08   & 0.13   & 0.18   & 43.38  & 203.5  &  \\
K & 5.5   & 2.75  & 0 & 0.26   & 0.42   & 0.54   & 40.29  & 11.6  &  \\
Th  & 11.23 & 6.14  & 0.01    & 1.02   & 1.37   & 1.75   & 170.46 & 30.5 &  \\ \cline{1-9}
\end{tabular}}
\end{table}
\renewcommand{\arraystretch}{1}

\subsection{Results}
\label{Results}

\begin{figure}[htp!]
\centering
\includegraphics[width=0.32\textwidth,trim={0.0cm 0cm 1.8cm 0cm},clip]{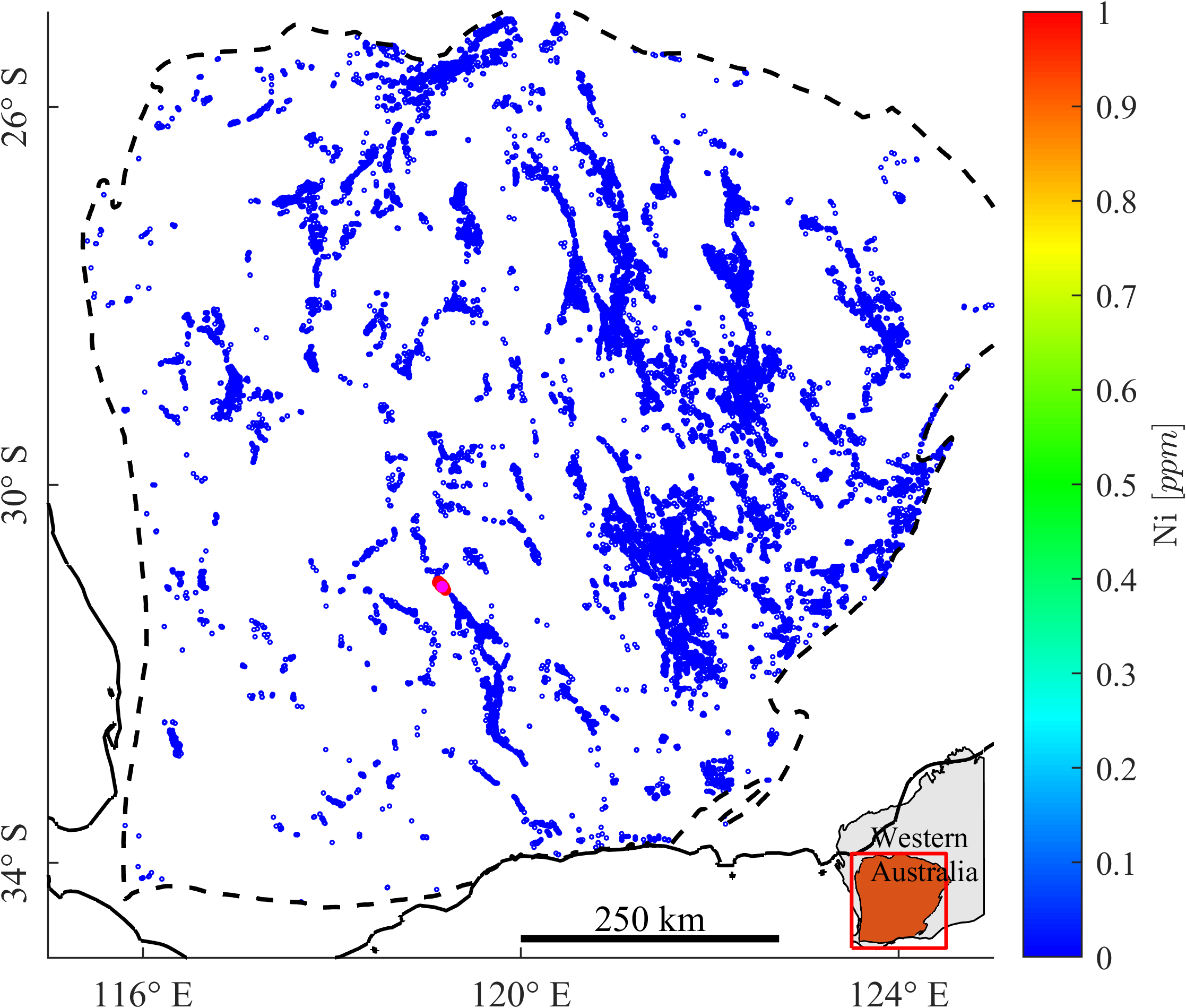}
\includegraphics[width=0.32\textwidth,trim={0.0cm 0cm 1.8cm 0cm},clip]{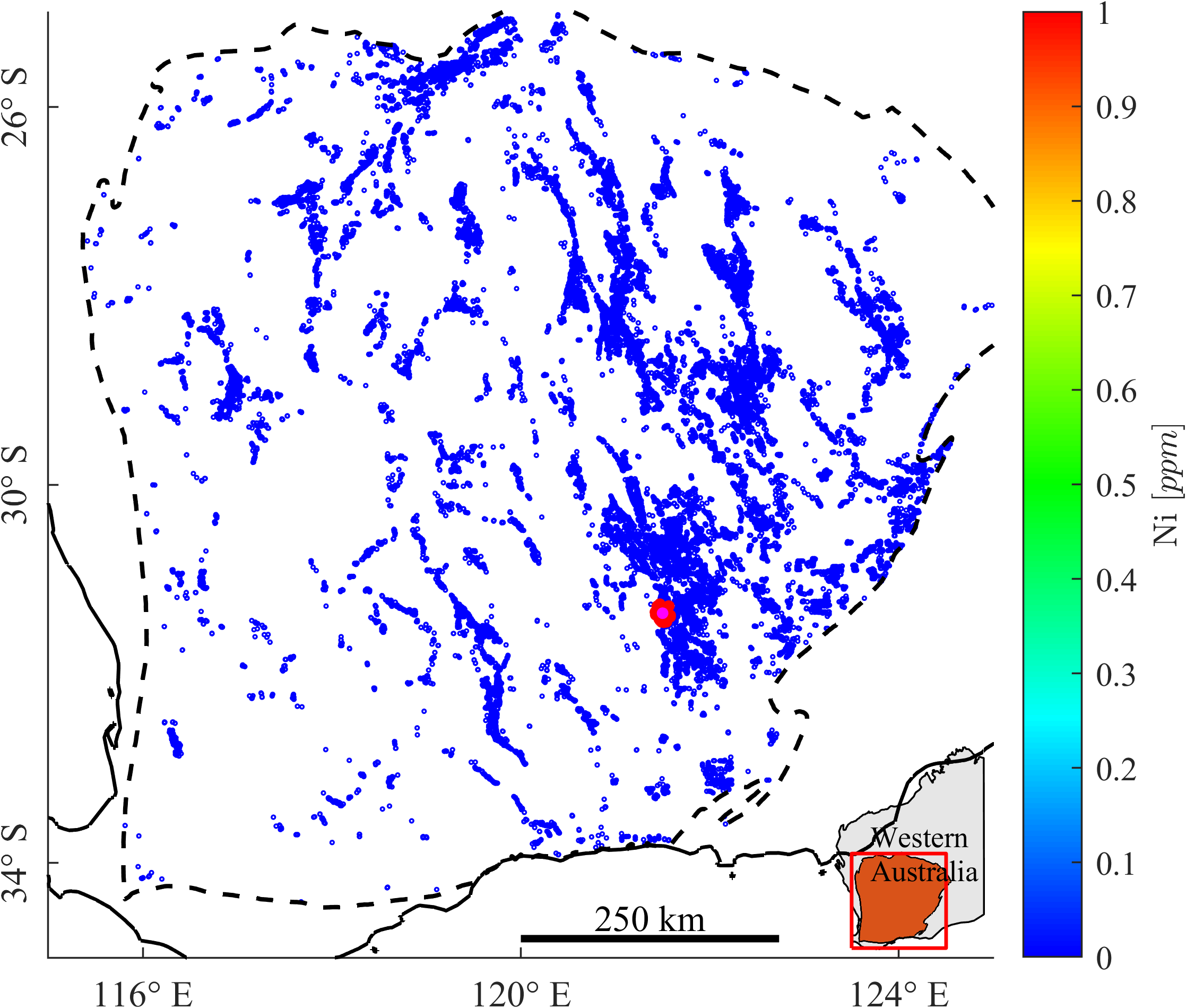}
\includegraphics[width=0.32\textwidth,trim={0.0cm 0cm 1.8cm 0cm},clip]{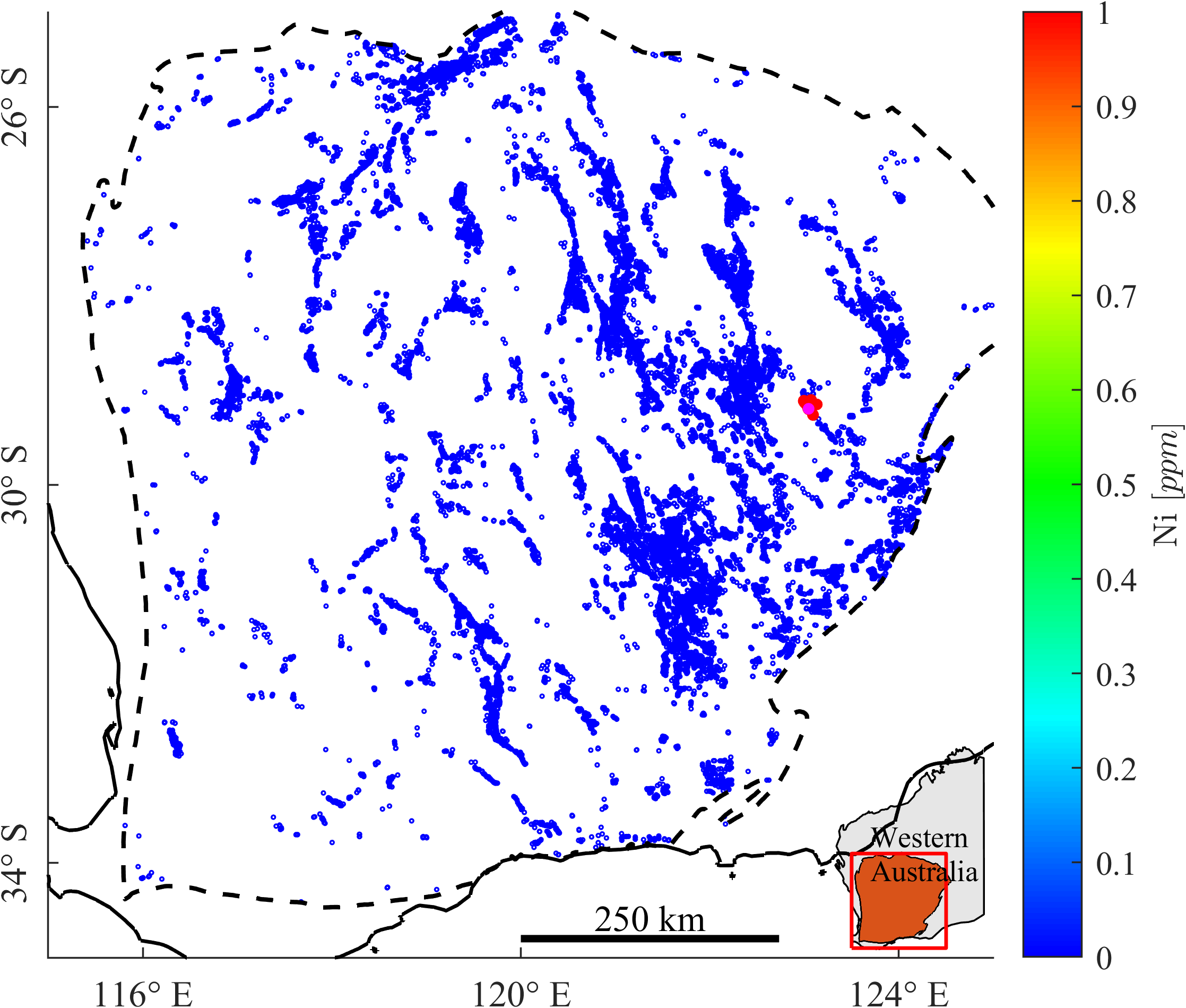}\\
\centering
\includegraphics[width=0.32\textwidth,trim={0.0cm 0cm 0cm 0cm},clip]{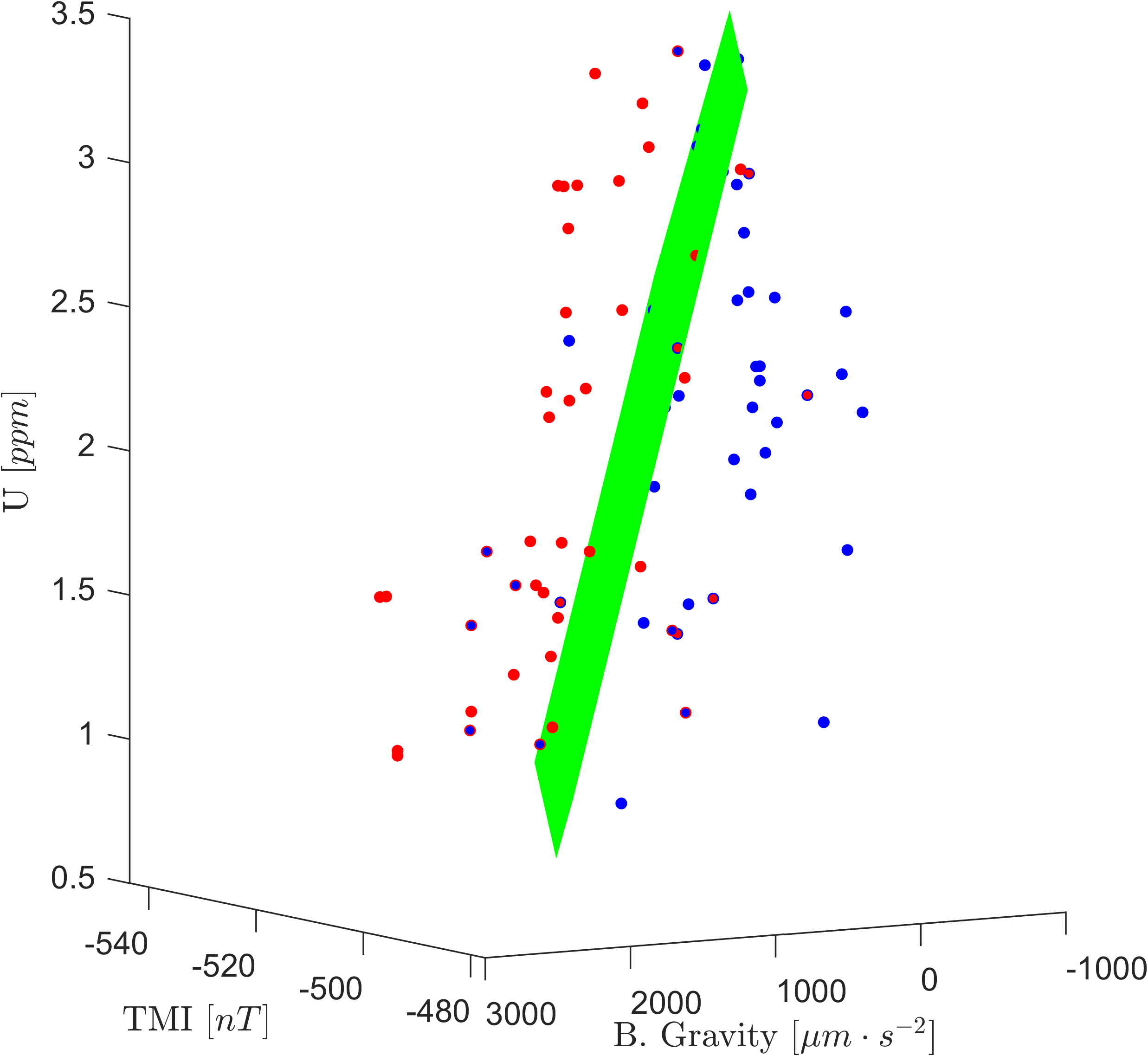}
\includegraphics[width=0.32\textwidth,trim={0.0cm 0cm 0cm 0cm},clip]{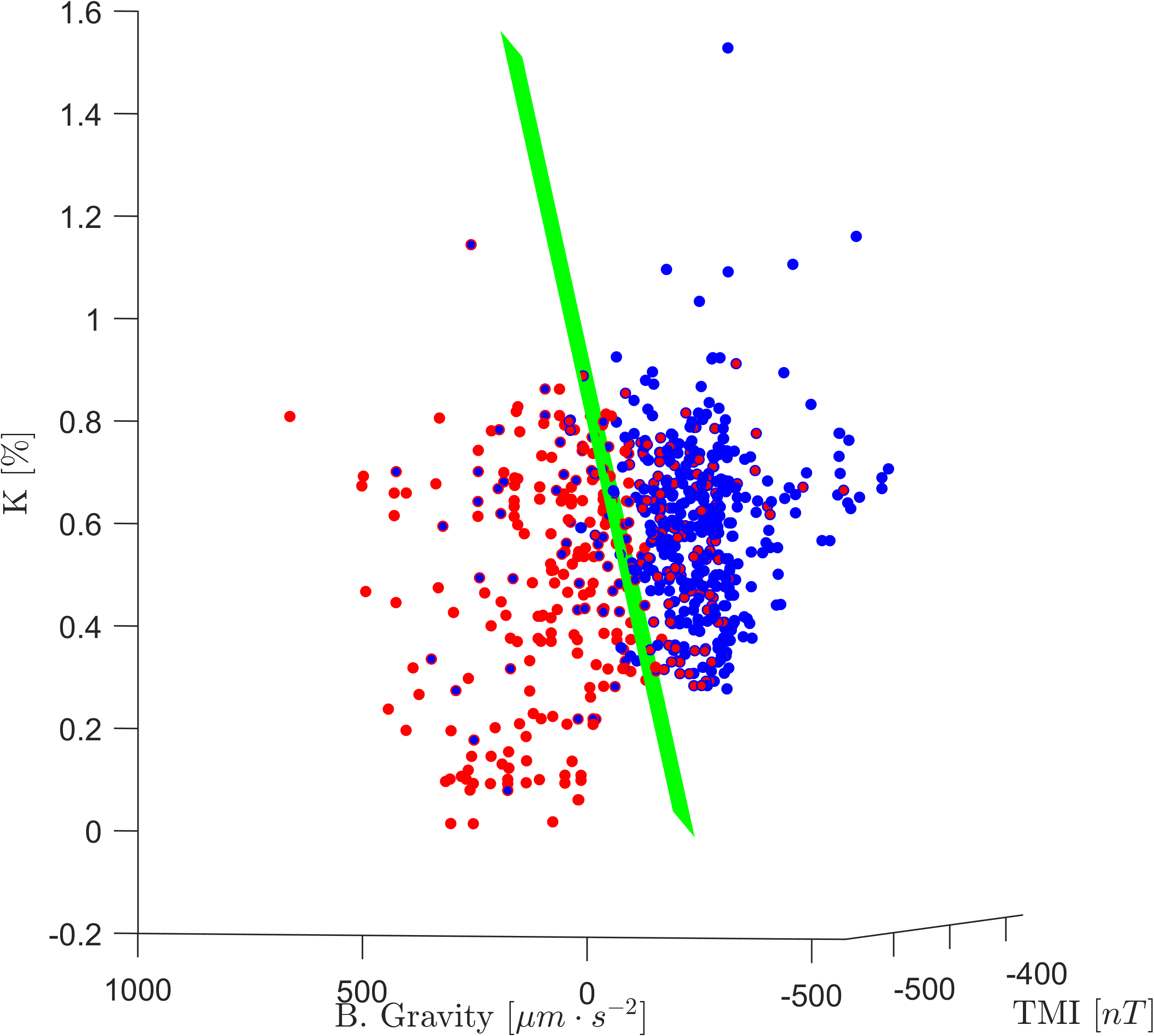}
\includegraphics[width=0.32\textwidth,trim={0.0cm 0cm 0cm 0cm},clip]{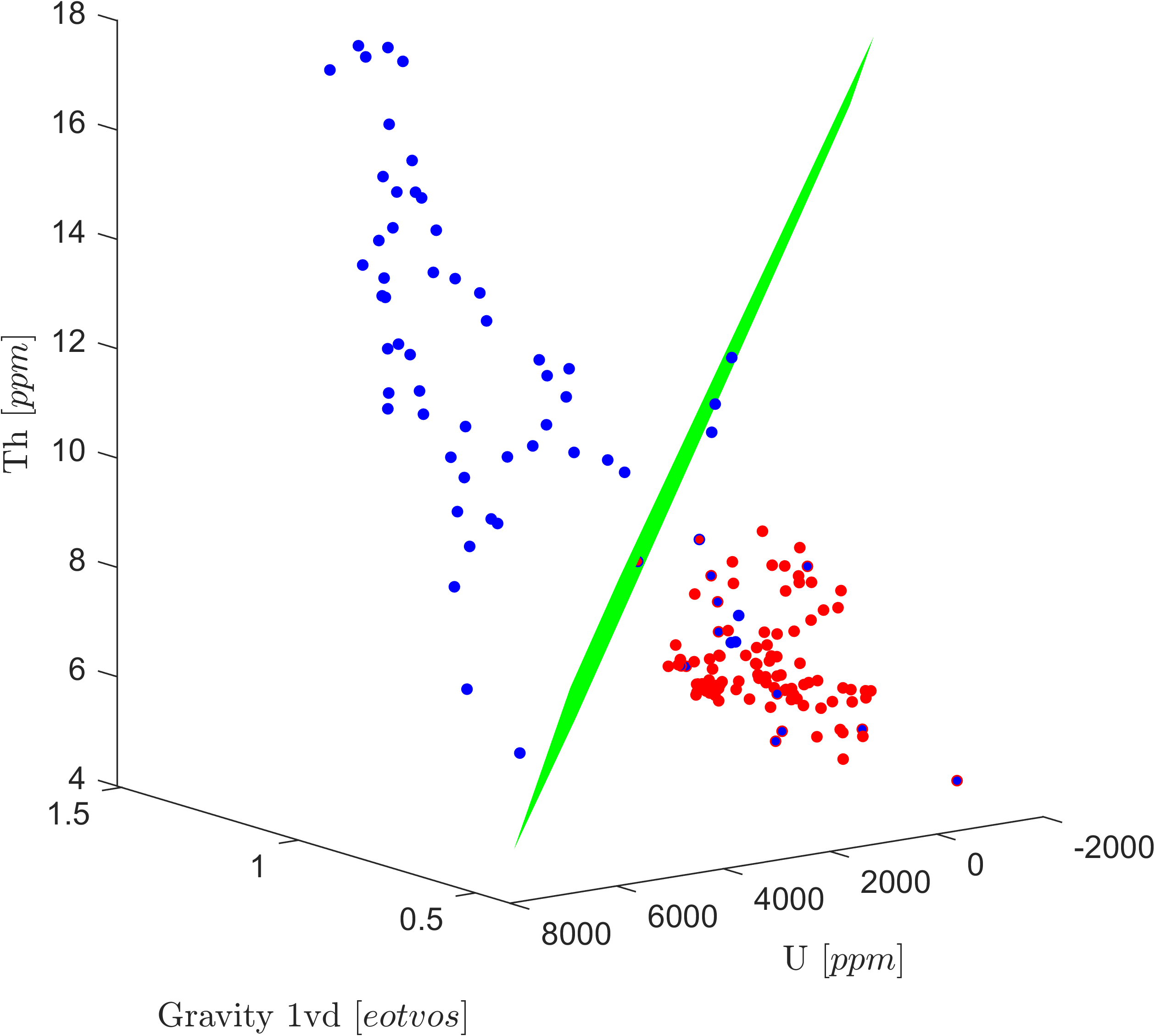}
\caption{SVMs adjusted at different training locations. Top: three different training location (magenta diamonds) and their corresponding neighborhoods of selected training geochemistry data (red circles). Bottom: separating boundary (green) projected into a lower three-dimensional feature space, considering the highest relevant features. Training data is shown in circles and predicted labels are shown on the edges of the markers, identifying misclassified points when colors do not match.}
\label{neigh}
\end{figure}

Verifying the central hypothesis raised in this work, that is, that relationships among input features and the binary response change across the area of study, is the first step accomplished. The verification is done through the inspection of calibrated SVM classifications at each training location, restricting the search neighborhood of training samples fed in the modeling by setting a maximum of 1,500 samples and a maximum search radius of 10 km. A minimum requirement of 5 positive training samples is additionally included in order to certify a representative classification, reducing the number of training locations to 889. Figure \ref{neigh} illustrates different search locations and corresponding neighborhoods, as well as the respective classification boundaries projected from the initial 7-dimensional feature space into a lower 3-dimensional space. The projection is made only for illustrative purposes and varies according to the importance of the input layer in the local classification, recalling that the squared of the $ i $-th component of classification-plane unit-normal can be used as a feature ranking criterion for the $ i $-th feature \citep{guyon2002gene}.

\begin{figure}[htp!]
\centering
\includegraphics[width=0.49\textwidth,trim={0.0cm 0cm 0.0cm 0cm},clip]{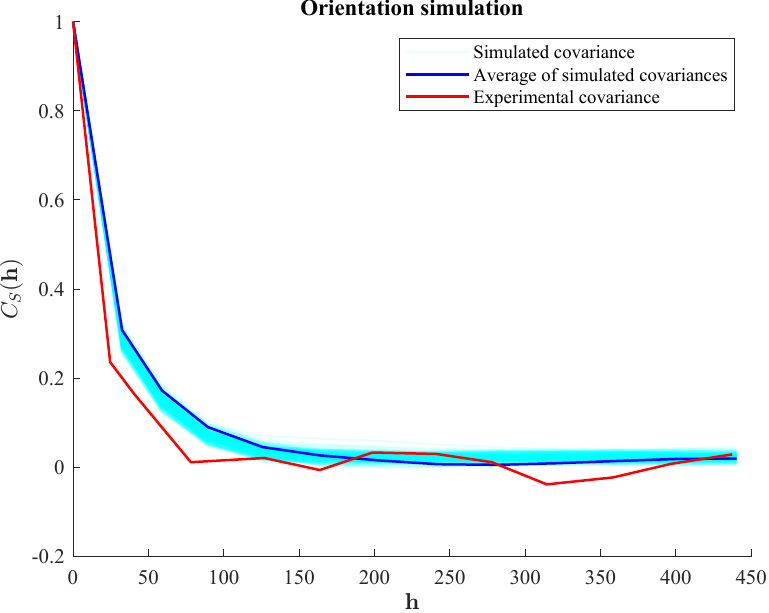}
\includegraphics[width=0.49\textwidth,trim={0.0cm 0cm 0.0cm 0cm},clip]{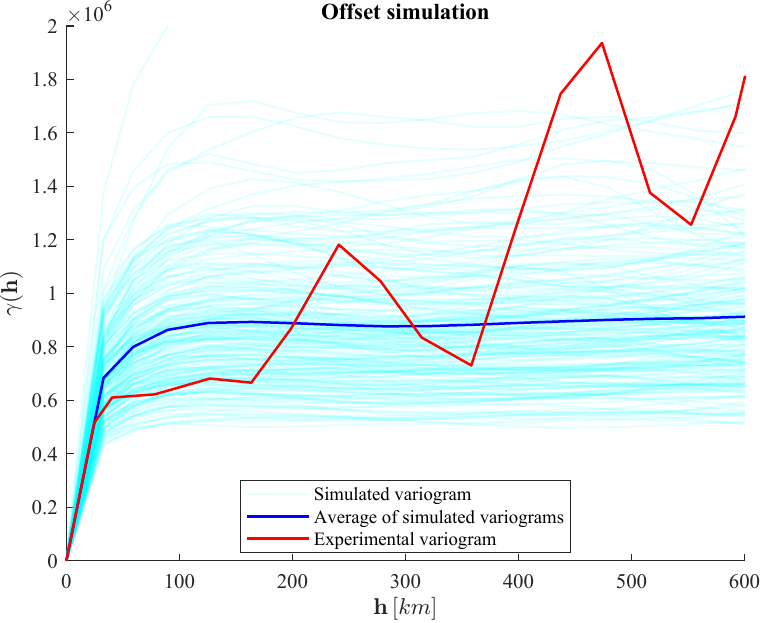}
\caption{Verifying the statistical reproduction of spatial continuity in spatially simulated response models.}
\label{stats}
\end{figure}

A collection of unit normal and offset pairs $(\textbf{s}_\alpha,{b}_\alpha)$ describing the adjusted SVM classification boundaries at training locations $ \textbf{u}_\alpha, \alpha\in \{1,\dots,N = 889\} $ is then obtained. Variography analysis of these parameters is carried out independently. Covariance analysis of $\textbf{s}_\alpha$ values show a very short effective range in relation to the area under analysis, close to 100 km, and with most of the covariance ($\sim 80 \%$) lost within the first 50 km. This fact indicates that any response model using the features selected in this study can be generalized only in a narrow area around any given training location. Regarding the ${b}_\alpha$ offset values, these show a higher spatial continuity, with a variogram partly stabilized initially at a range $\sim 120$ km, but then continuing with a persistent trend after such distance. This trend is related to the both-sided heavy tails displayed in some of the input features probability distributions (verified by the kurtosis values shown earlier, in Tab. \ref{tab:becn}). Therefore, when the classification boundary is calibrated at locations displaying extreme values on their input features, a high value in the offset is likely to be obtained. The experimental covariance of orientations and the offset variogram are shown in Fig. \ref{stats}. In the same figure, we include the CDFs of the orientation vectors decomposed into their feature components. Both the variography of orientations and the CDFs are key to validate our proposed workflow.

\begin{figure}[htp!]
\centering
\includegraphics[width=1\textwidth,trim={0.0cm 0cm 0.0cm 0cm},clip]{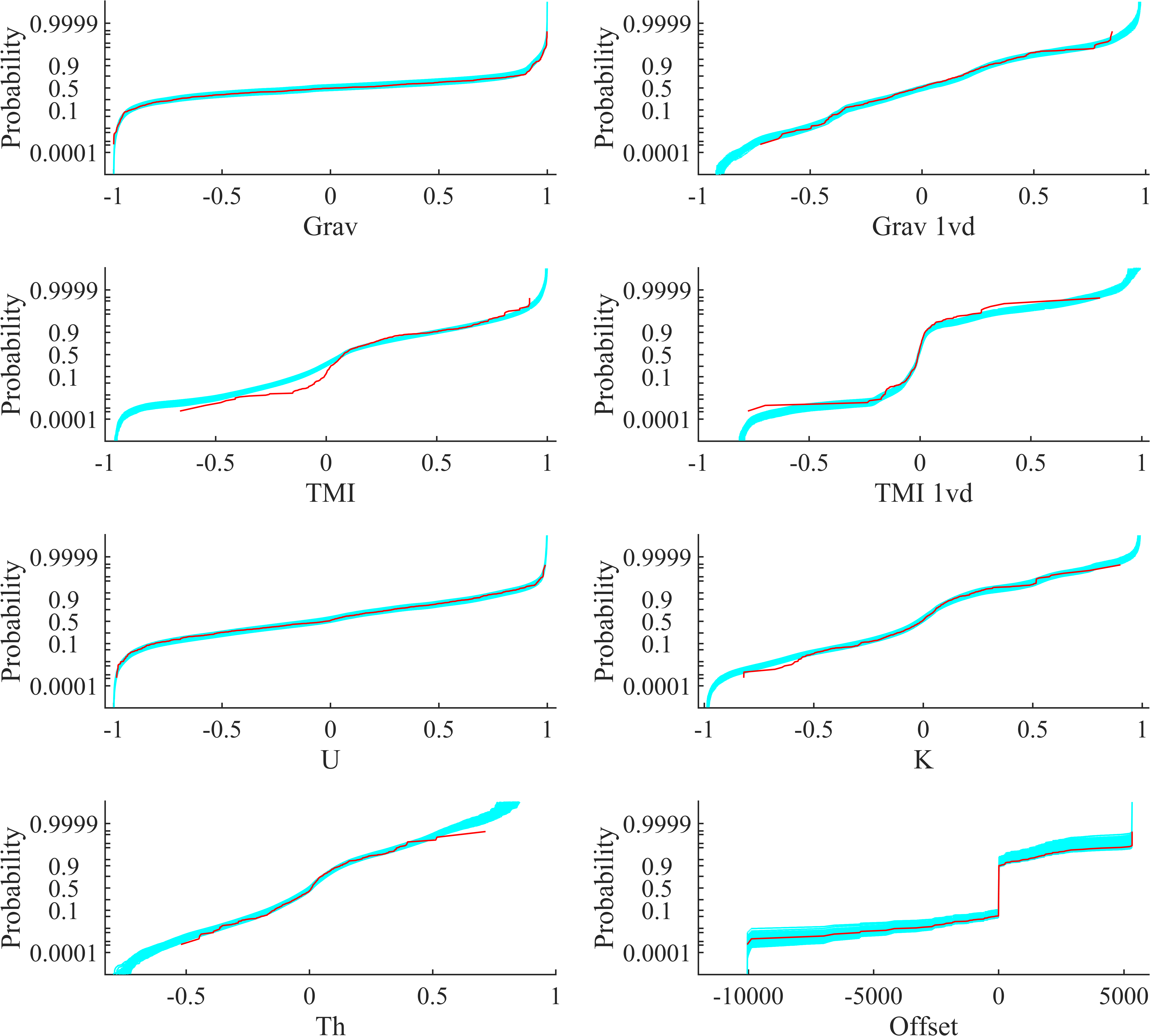}
\caption{Verifying the statistical reproduction of CDFs in spatially simulated response models.}
\label{stats2}
\end{figure}

The $\textbf{s}_\alpha$ orientation values are mapped into a uniform distribution on the 6-dimensional hyper-sphere ($p=7$) using the procedure described in Subsect. \ref{ar2sphe}, and non-conditional orientation fields were generated starting from a 7-dimensional vector of Gaussian RFs. Each independent component on this Gaussian vector was modeled using an exponential covariance with a theoretical range equal to 40 km. Every single realization is conditioned after normalizing the Gaussian vector, using the geometrical procedure proposed in Subsect. \ref{condi}, obtaining then a realization of the $S(\textbf{u})$ field. On the other hand, the Gaussian simulation of the $B(\textbf{u})$ RF is carried out through the standard procedure, Gaussianizing the probability distribution of the offset values followed by conditional simulation using an exponential covariance of 100 km range. 200 conditional realizations of the $\big(\textbf{S}(\textbf{u}),B(\textbf{u})\big)$ pair are drawn. Figure \ref{stats} includes the covariance and variogram of each $\textbf{S}(\textbf{u})$ and $B(\textbf{u})$ realization, respectively, and their probability plots. The spatial covariance of the $\textbf{S}(\textbf{u})$ simulations follow closely the experimental one and could be further improved to follow closely the experimental values at smaller lags just by decreasing the range imposed on the Gaussian vector components, at the expense of sacrificing spatial conditioning at unexplored locations. The simulation modeling of $B(\textbf{u})$ follows closely the experimental variography near the origin besides the presented challenge posed due to the trend displayed in the variogram, and can be further improved in order to display fewer fluctuations in the variography of simulations by modeling such a trend. 

\begin{figure}[htp!]
\centering
\includegraphics[width=0.45\textwidth,trim={0.0cm 0cm 0.0cm 0cm},clip]{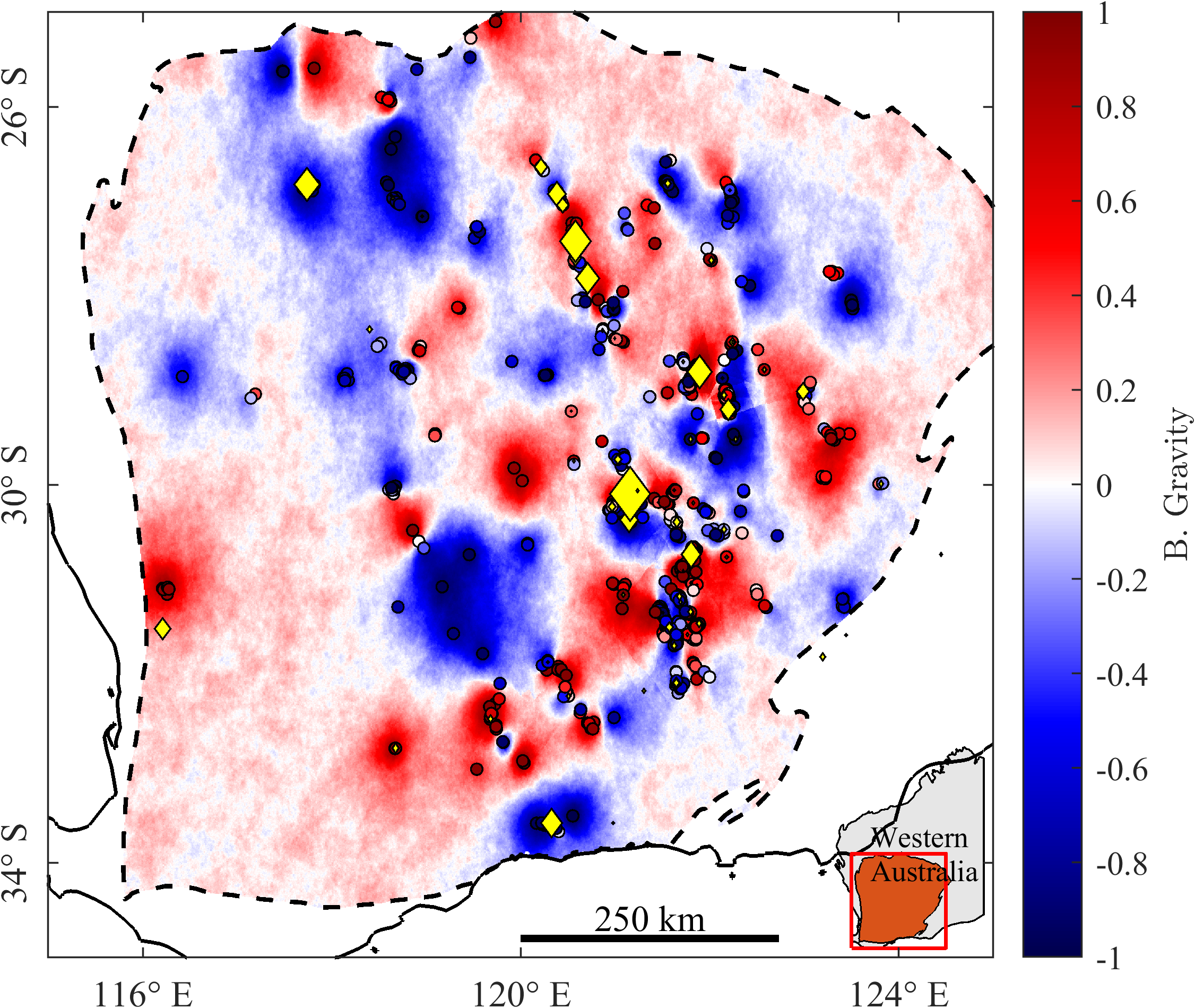}
\includegraphics[width=0.45\textwidth,trim={0.0cm 0cm 0.0cm 0cm},clip]{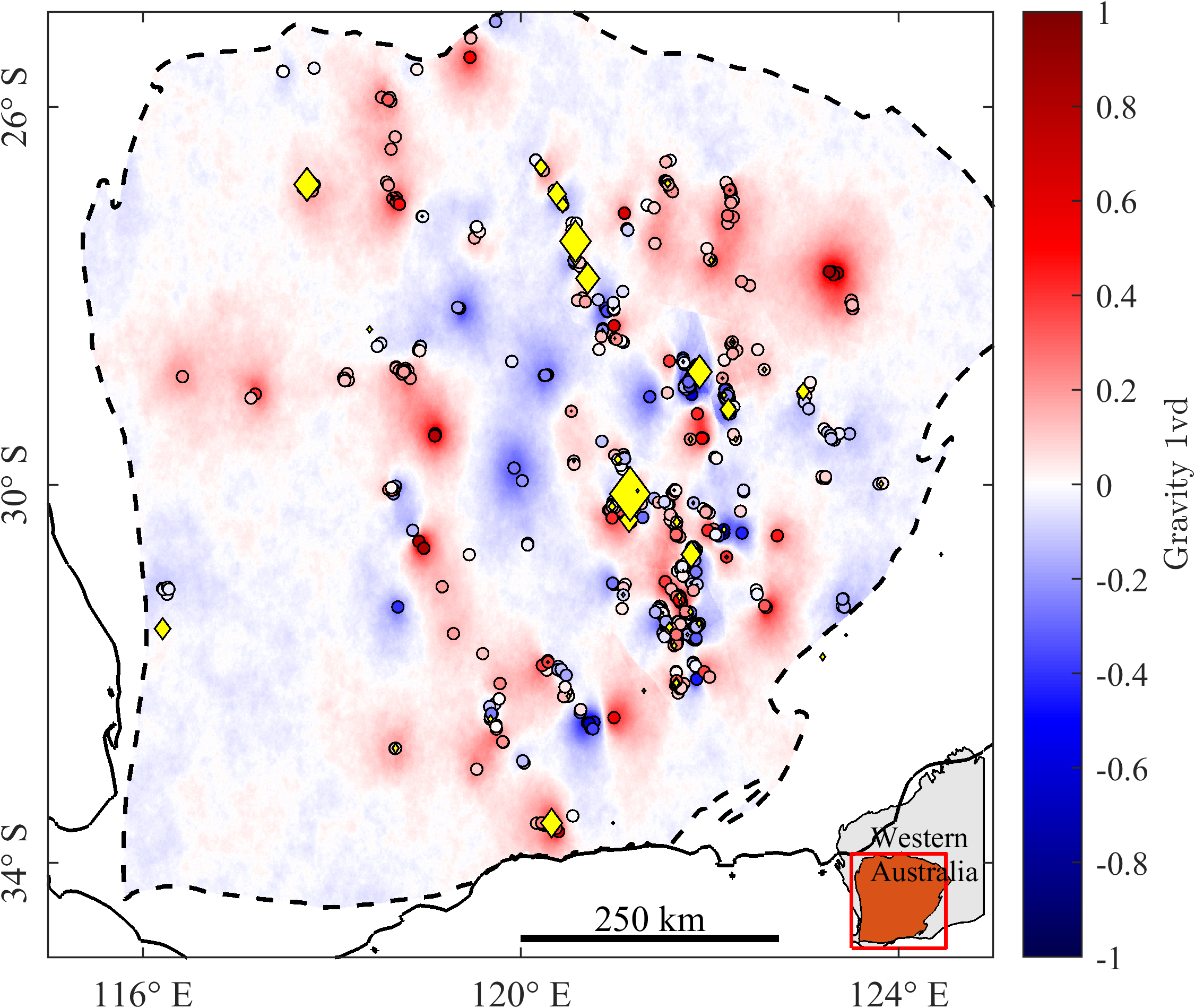}
\includegraphics[width=0.45\textwidth,trim={0.0cm 0cm 0.0cm 0cm},clip]{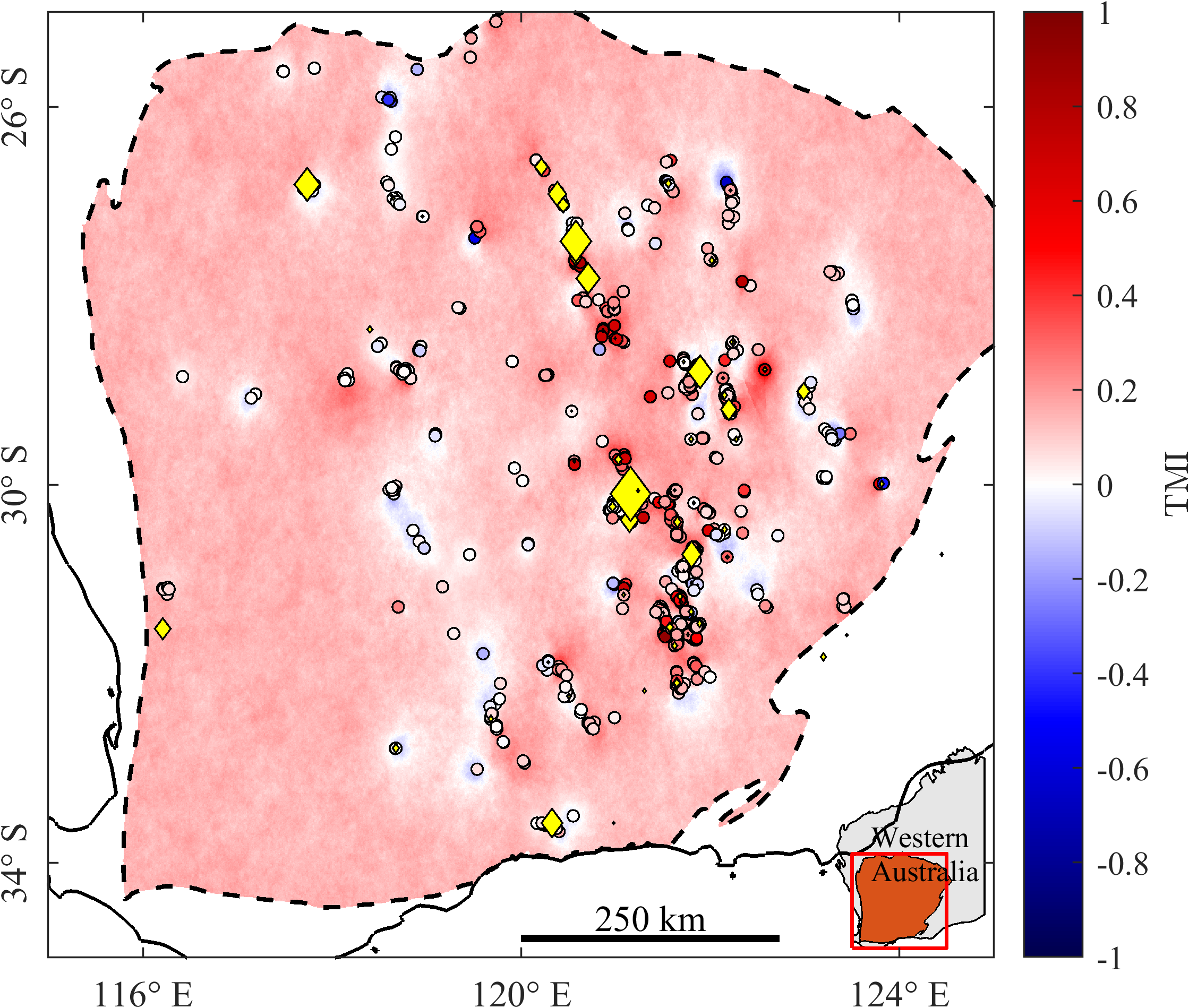}
\includegraphics[width=0.45\textwidth,trim={0.0cm 0cm 0.0cm 0cm},clip]{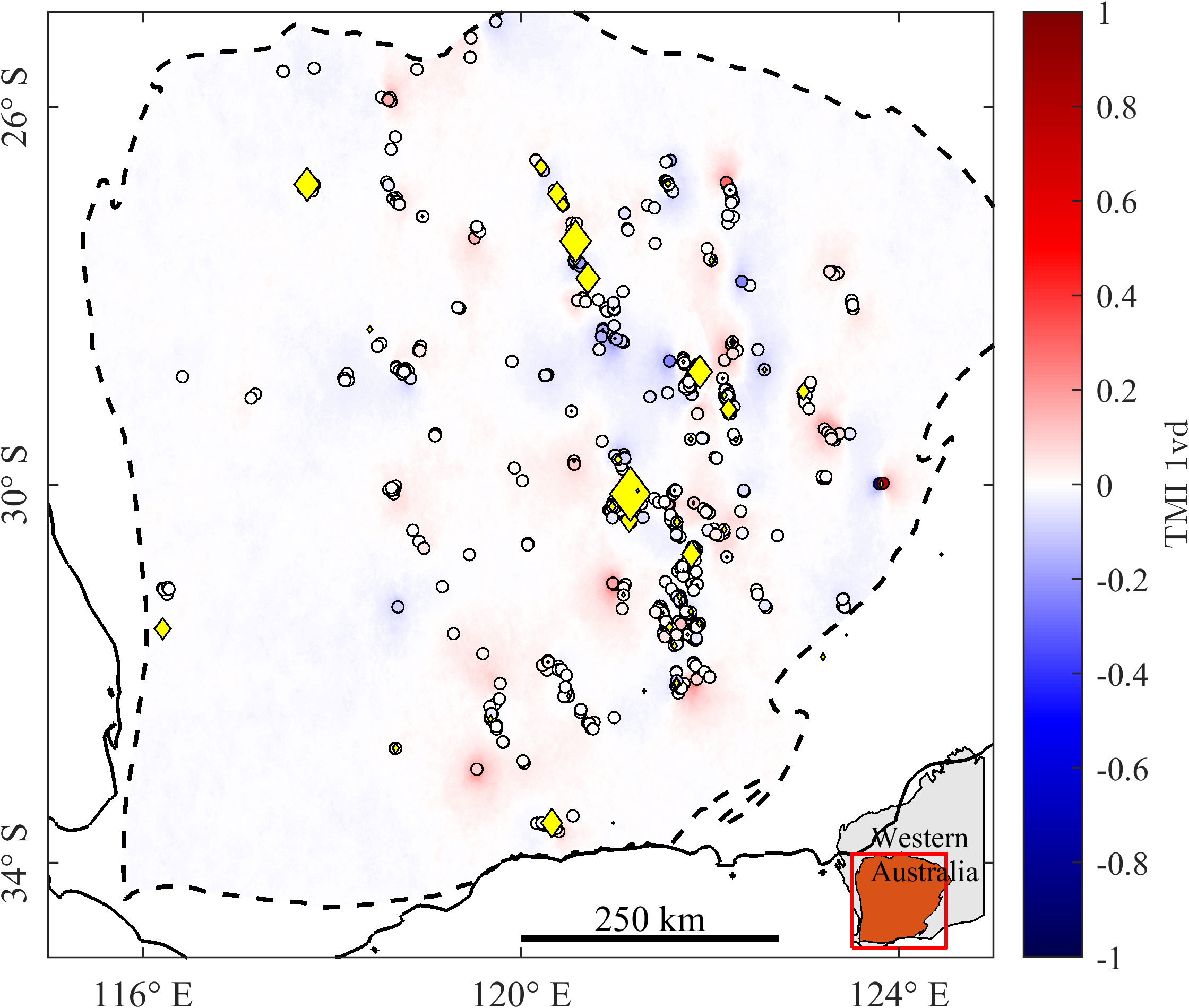}
\includegraphics[width=0.45\textwidth,trim={0.0cm 0cm 0.0cm 0cm},clip]{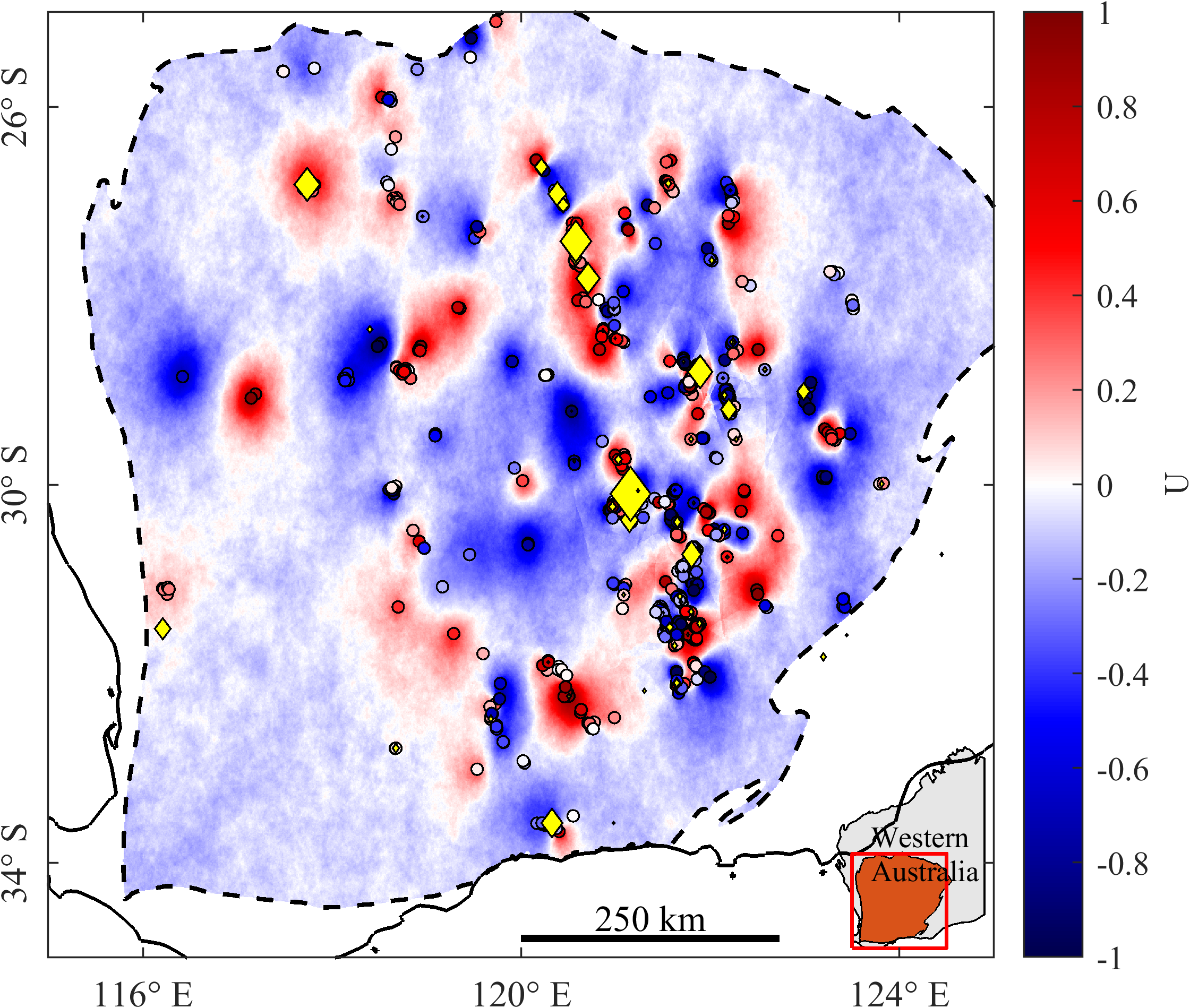}
\includegraphics[width=0.45\textwidth,trim={0.0cm 0cm 0.0cm 0cm},clip]{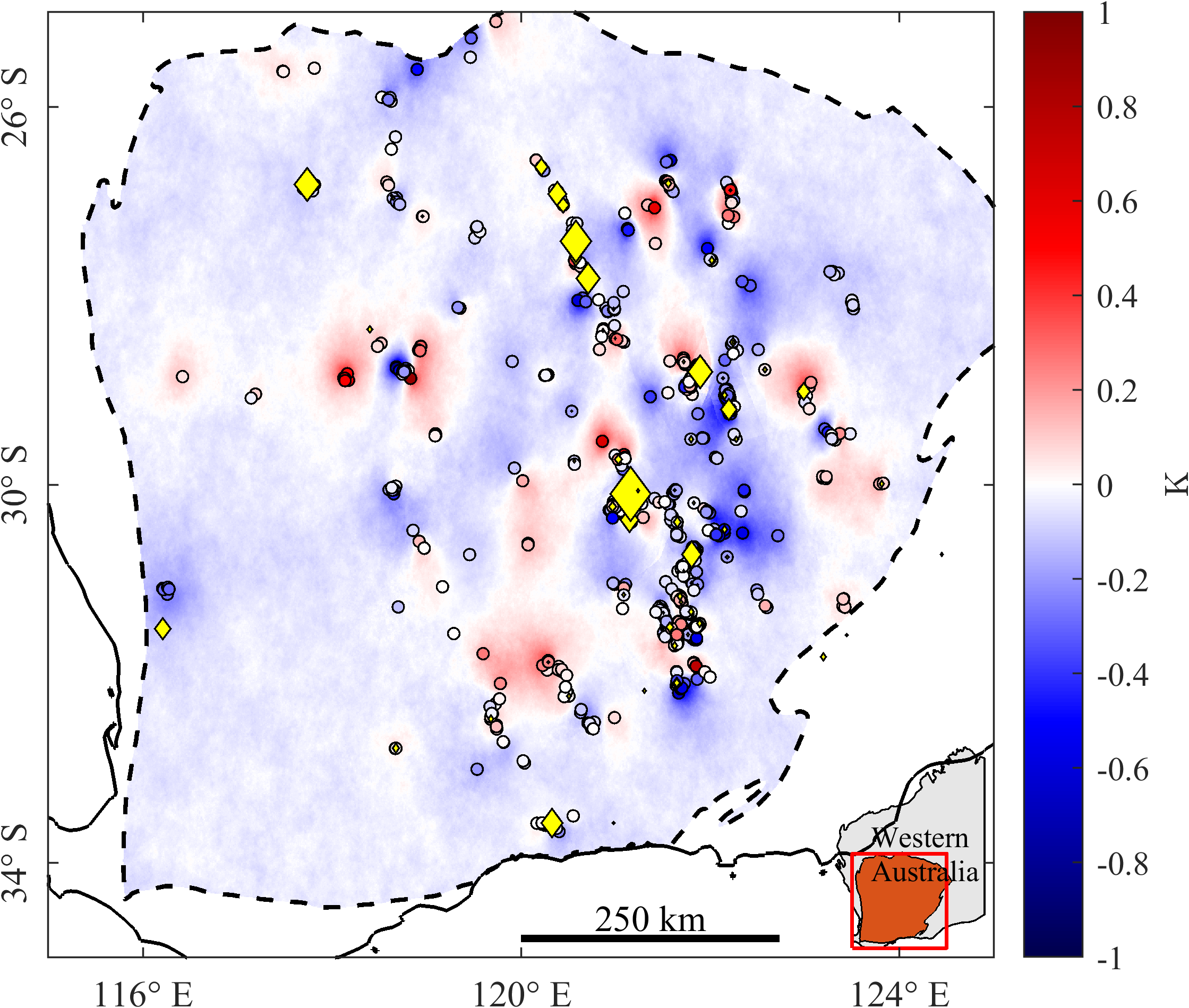}
\includegraphics[width=0.45\textwidth,trim={0.0cm 0cm 0.0cm 0cm},clip]{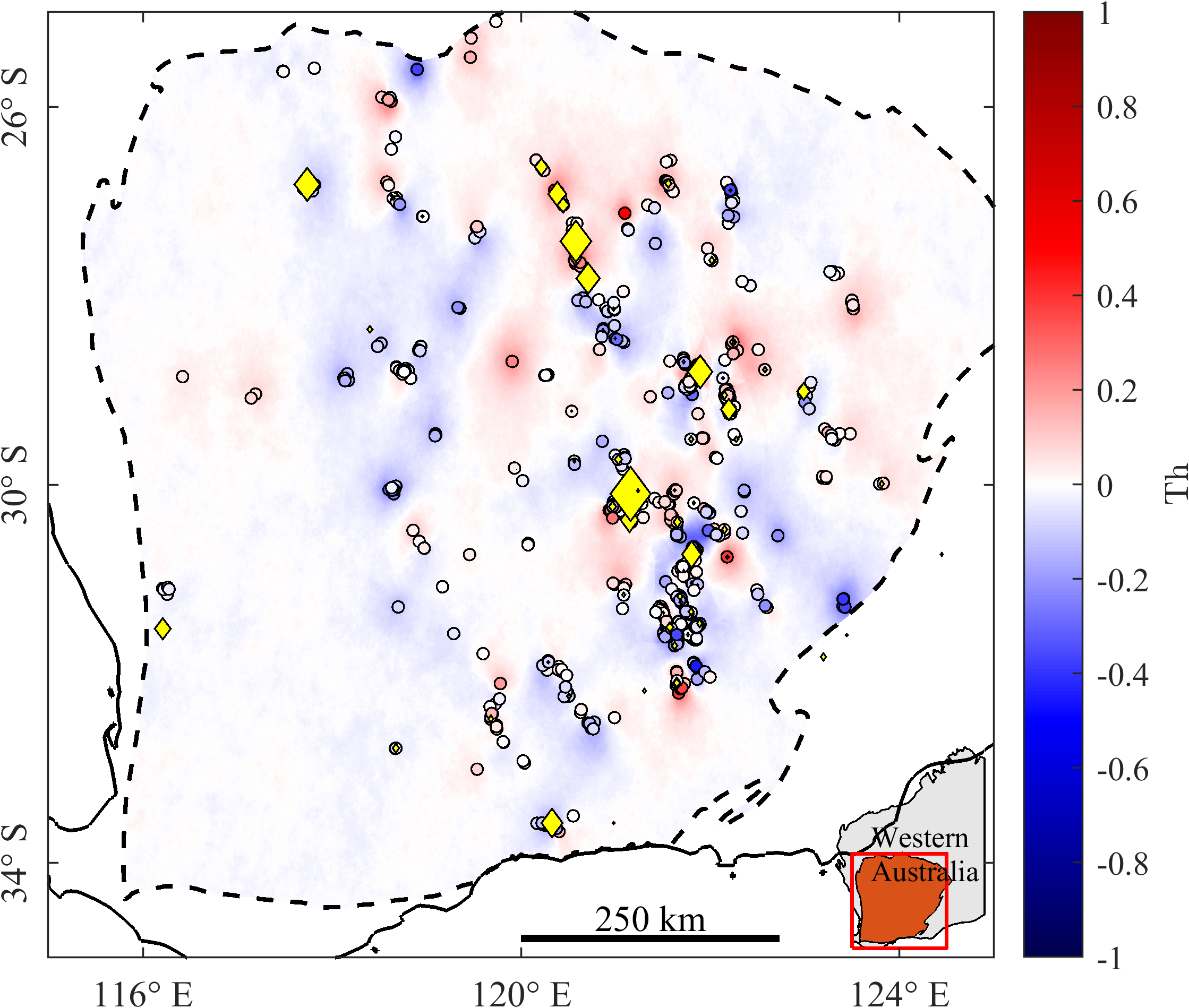}
\includegraphics[width=0.45\textwidth,trim={0.0cm 0cm 0.0cm 0cm},clip]{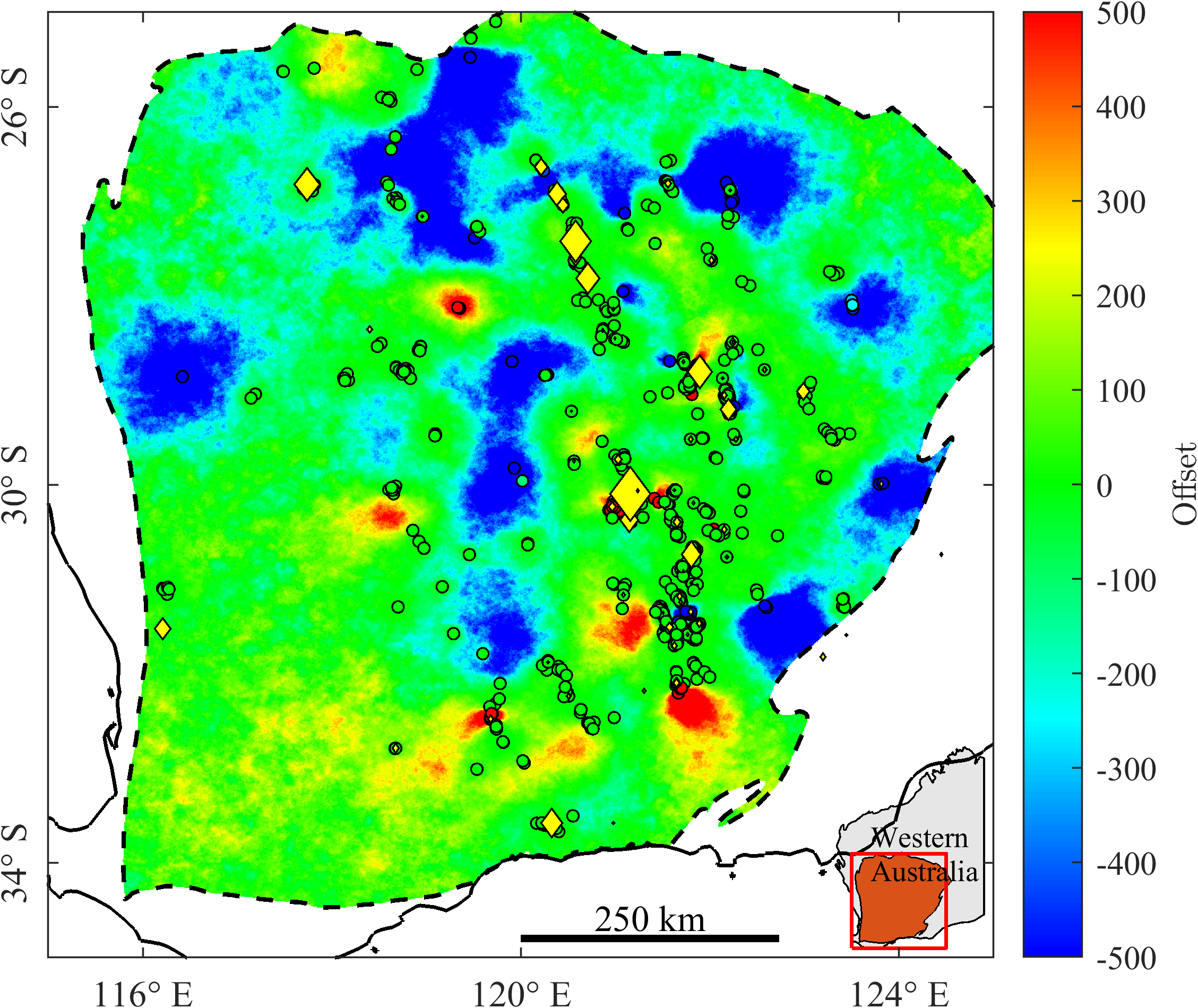}
\caption{Average feature component value. This average value is related to the feature's importance: darker areas (either in red or in blue) reflect higher importance than whiter values. The offset average values are also included.}
\label{modelpar}
\end{figure}

As mentioned earlier, the realizations of $\textbf{S}(\textbf{u})$ and $B(\textbf{u})$ give us valuable information by itself, such as the importance of each feature in predicting the response variable, as a function of the location \textbf{u}. As the inference of this information does not require an exhaustive knowledge of input features on the domain but only at training locations, this predictive output takes relevance as guiding any further acquisition of features in the geographical domain. Figure \ref{modelpar} includes the average feature value importance on the domain and the average offset value, taken from the 200 simulations.

\begin{figure}[htp]
\centering
\includegraphics[width=0.321\textwidth,trim={0.0cm 0cm 1.8cm 0cm},clip]{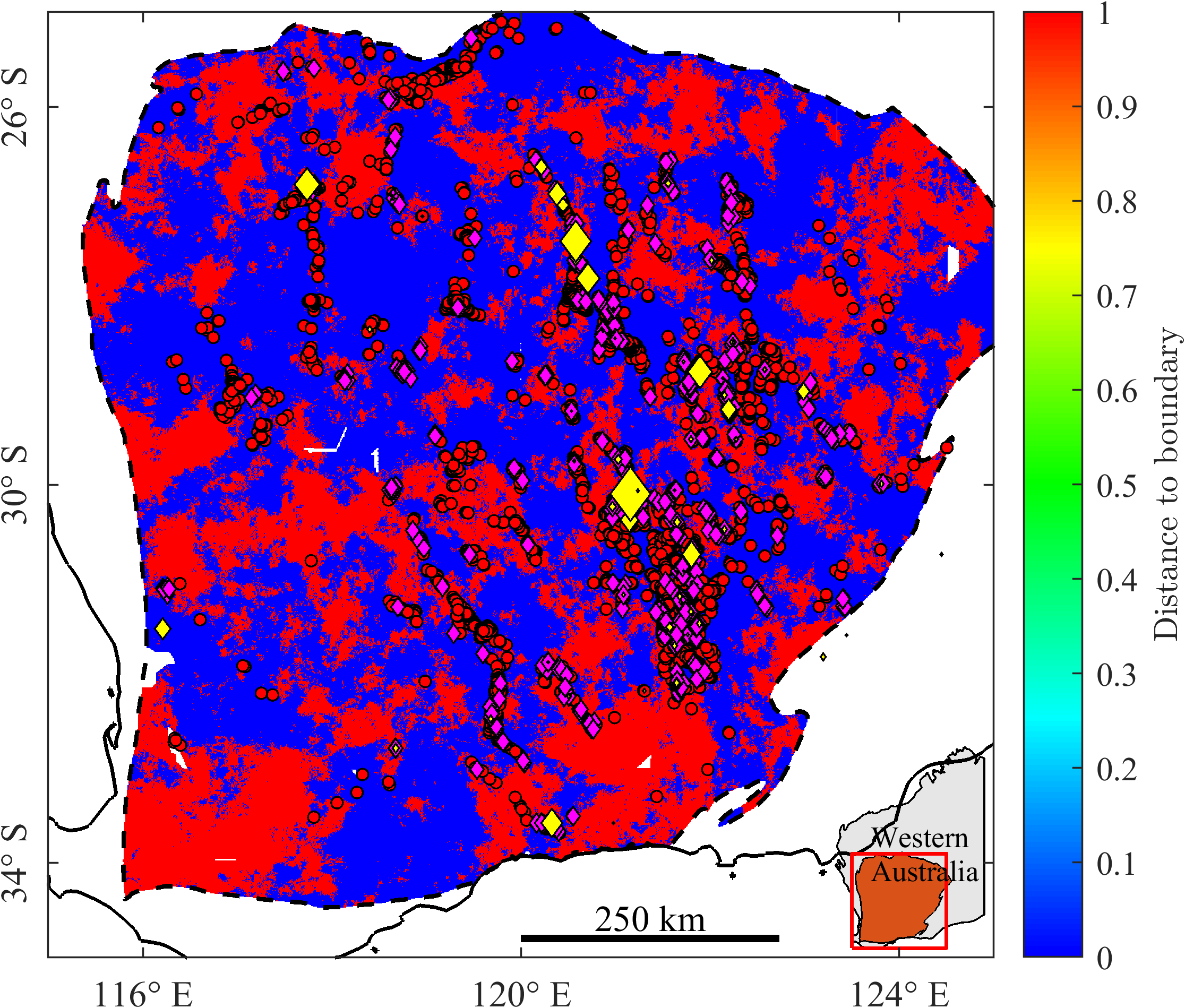}
\includegraphics[width=0.321\textwidth,trim={0.0cm 0cm 1.8cm 0cm},clip]{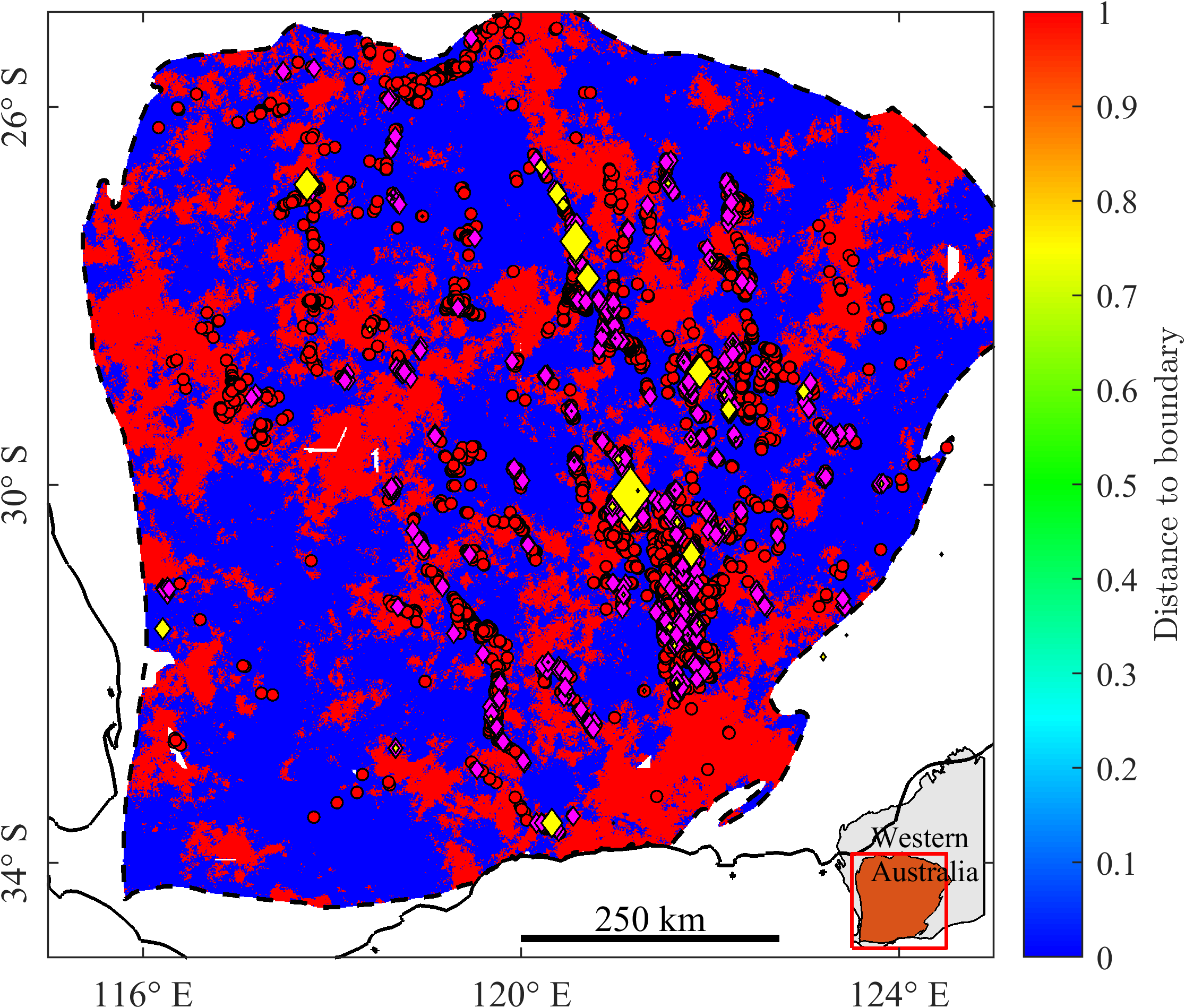}
\includegraphics[width=0.321\textwidth,trim={0.0cm 0cm 1.8cm 0cm},clip]{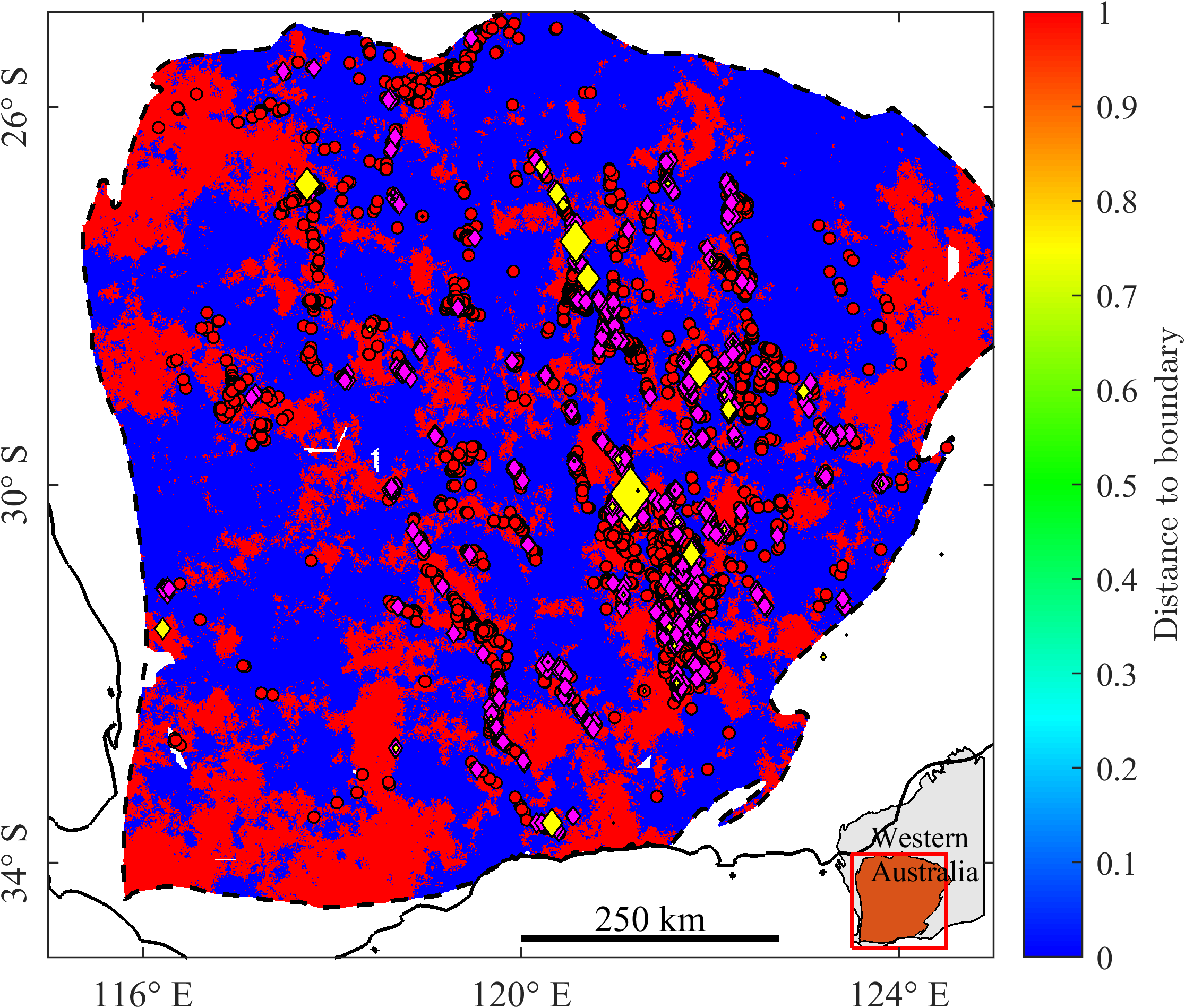}
\includegraphics[width=0.49\textwidth,trim={0.0cm 0cm 0.0cm 0cm},clip]{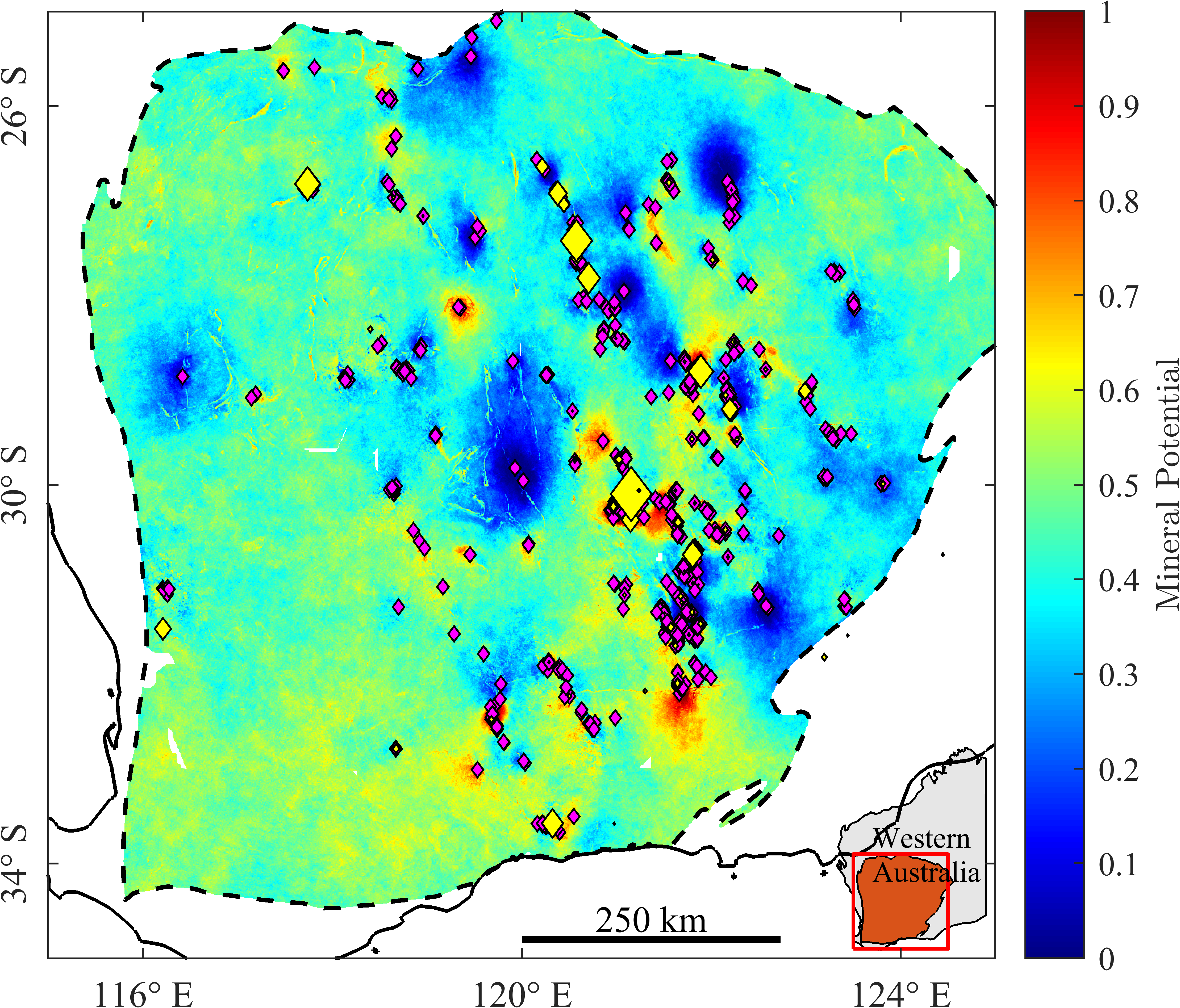}
\includegraphics[width=0.49\textwidth,trim={0.0cm 0cm 0.0cm 0cm},clip]{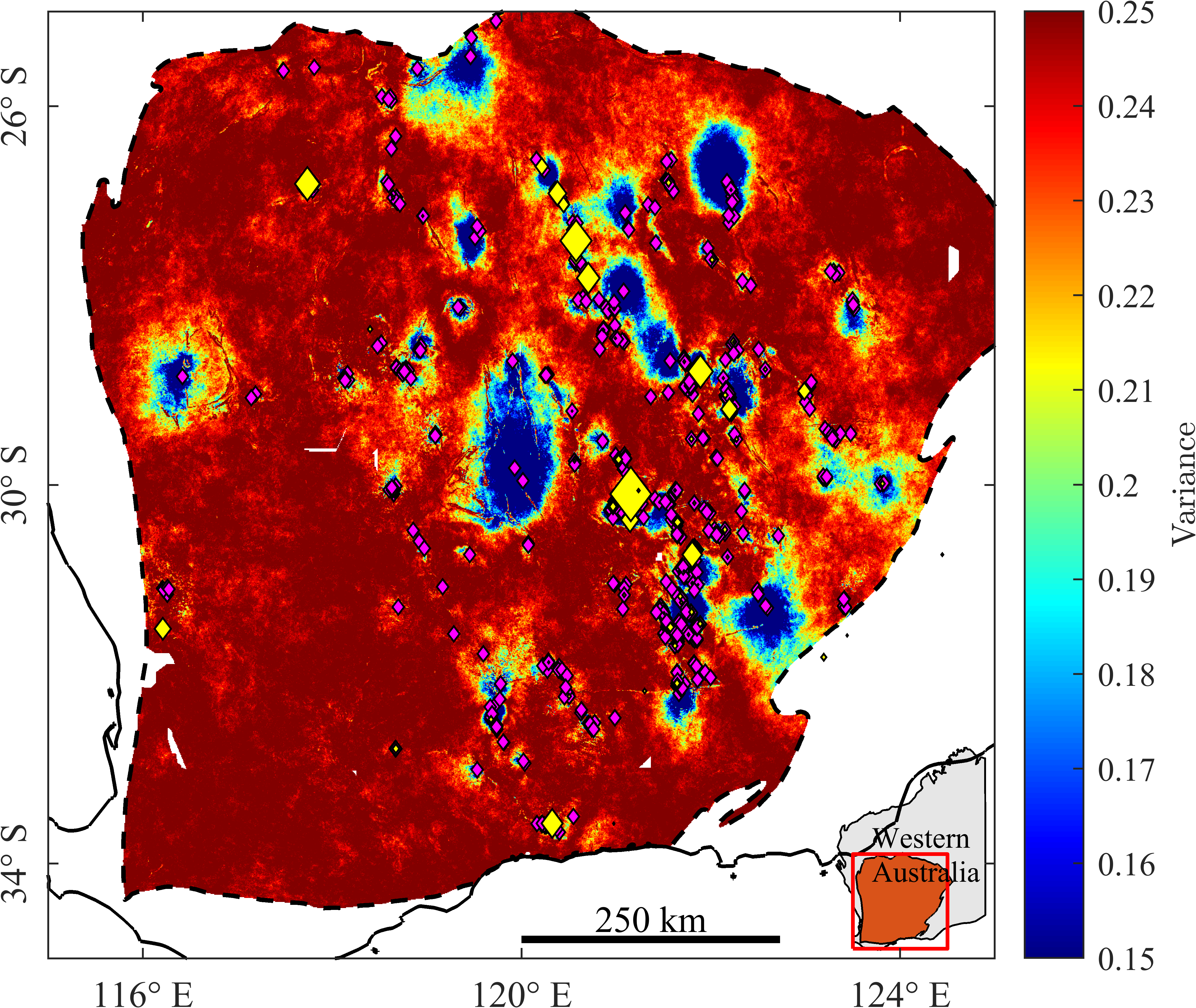}
\caption{The $\mathbbcal{SVM}(\textbf{u})$ RF. Top: three realizations classifying the study area as prospective or not (red or blue, respectively), together with the positively classified training sampled used in the SVM calibration (negative samples are omitted in this plot). Bottom: the E-Type and variance for 200 realizations, representing the mineral potential map an its uncertainty.}
\label{results}
\end{figure}

As pointed out earlier, when input features are known exhaustively in the spatial domain, it is possible to obtain the local response by inquiring the simulated models. In this case study, the regionalized vector $\textbf{z}(\textbf{u})$ of input features is used to evaluate each realization of the SVM RF $\mathbbcal{SVM}(\textbf{u})$, explicitly obtaining a response by $\mathbbcal{SVM}(\textbf{u}) = \textmd{sign}\big(\left\langle\textbf{S}(\textbf{u}),\textbf{z}(\textbf{u})\right\rangle + B(\textbf{u})\big)$. A few realizations of this classification response field are displayed in Fig. \ref{results}. It is important to note that the spatial field emulates a classification process within the geographical space, and not a process of mineralization. The mineralization outcomes are not used directly in the simulation workflow; they are only used indirectly to calibrate a classification boundary (often fitted by using non-separable data). 

The E-type and the variance for the 200 $\mathbbcal{SVM}(\textbf{u})$ realizations are included in the same Fig. \ref{results}, which can be interpreted as the mineral potential map and its uncertainty, respectively. On the one hand, the proposed methodology applied to this study area provides decisive results toward explored areas, assigning extended areas with a negative classification and some minor zones with a positive classification. On the other hand, away from the conditioning data, the classification tends to be highly cautious. Outside the $~100$ km range imposed in the modeling of RFs, most locations are assigned with a uniform binary distribution, which can be observed from the $0.5$ values in the E-type and the $0.25$ values in the variance map (maximum variance for a binary classification). However, some scattered locations are assigned with an E-type value above the $0.5$ threshold, which may indicate opportunities for mineral exploration away from well-explored areas.

The proposed mineral potential map, based on the use of dRFs, is validated through the use of a success rate curve, constructed by plotting the proportion
of mineral deposits (or resources) used for validation on the $y$-axis against the proportion of the study area classified as prospective on the $x$-axis. Overall, the graph indicates that nearly $60 \%$ of known locations having resource information are found within the first $10 \%$ of the area predicted as prospective. Such area accounts for nearly $85 \%$ of total resources. 

\begin{figure}[htp]
\centering
\includegraphics[width=0.65\textwidth,trim={0.0cm 0cm 0.0cm 0cm},clip]{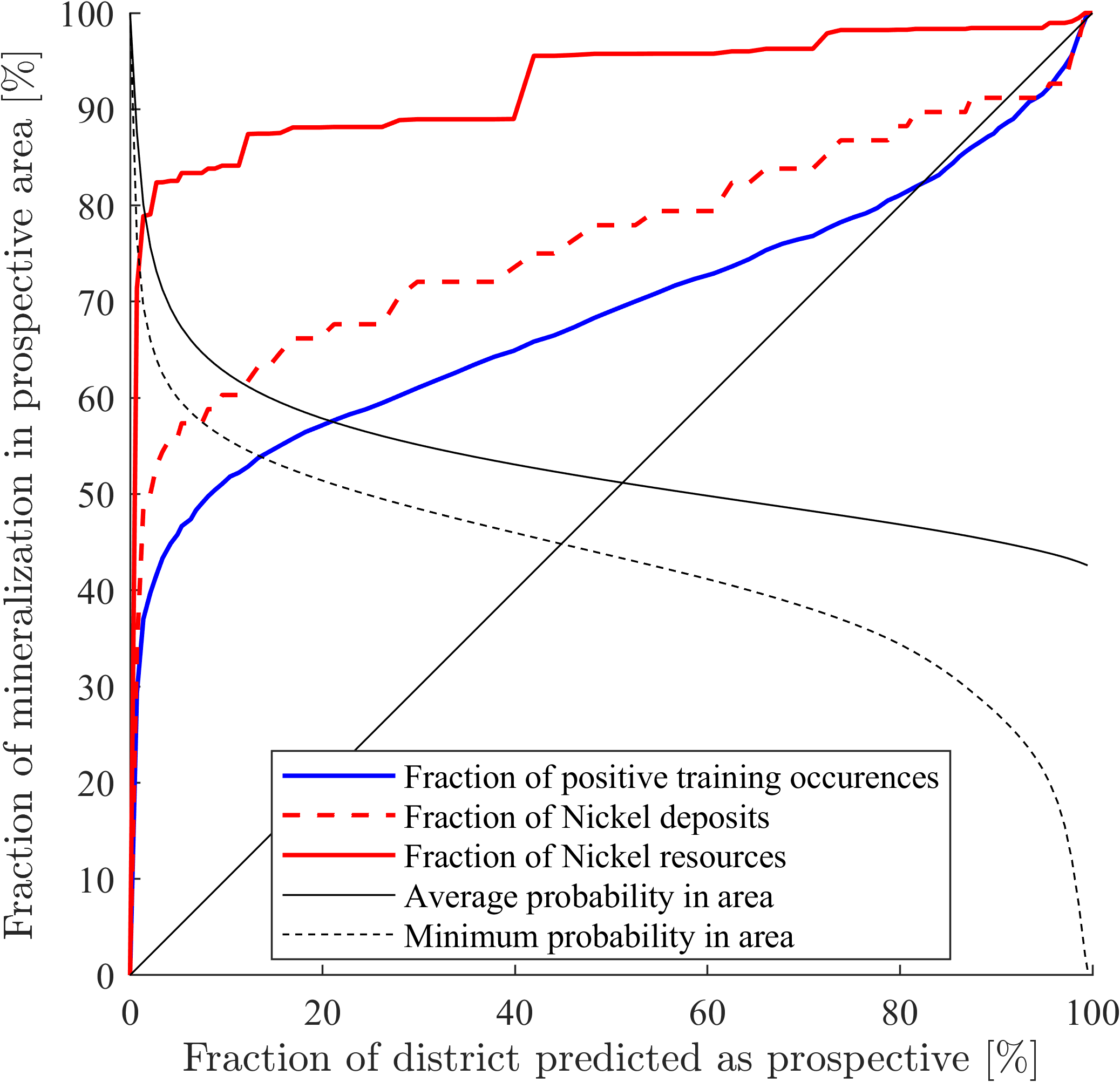}
\caption{Success rate curve for the proposed mineral potential map, whose model is based on the use of dRFs, together with probability measures (as a percentage) on the prospective area. Overall, the graph indicates that nearly $60 \%$ of known locations with resource information, accounting for nearly $85 \%$ of total resources, are found within the first $10 \%$ of the area predicted as prospective.}
\label{frac}
\end{figure}

\subsection{Discussion}
\label{Discussion} 
This section is closed with a short discussion on two pertinent points concerning the implementation of the methodology in the case study, focusing on the selection of hyperparameters throughout the process.

The selection of the vicinity is a crucial aspect when calibrating local SVMs, defining how input attributes correlate with outputs. Therefore, we investigate how the vicinity size affects the inference results. Following numerous tests, we have determined that utilizing a broad enough vicinity to encompass a diverse range of input features and sufficient binary results is preferred to a smaller vicinity size, prone to inadequate generalizing in the feature space. An ellipsoidal search pattern, if implemented, needs to be aligned perpendicular to the main direction of the input-outcome correlation. For example, in our case study, we find an alignment of certain occurrences with zones where gravity values are higher, predominantly in a north-south orientation. Therefore, choosing samples from regions within and adjacent to the anomalies gives a better representation of the behavior to be interpolated. Still, it is important for the vicinity not to be excessively large as this translates into different models having too many sample pairs in common, reducing the variability observed among the models in the study area.

A second aspect to consider is related to the variographic analysis and modeling, and its practical challenges after the decision of carrying the initial probability density of orientations into a uniform distribution in the hyper-sphere.
This transformation introduces some spatial issues that are difficult to manage theoretically. These complications primarily emerge due to the application of PPMT on preliminary data, where the spatial behavior of data is hardly traceable throughout the iterative Gaussianizing procedure. However, the main issue lies in the quantile matching from the SN distribution into a uniform distribution. This step induces a shrinking of the range in the experimental covariance of transformed orientations, as any initial distribution of orientation vectors concentrated around a certain mean direction will display a correlated behavior that is partially lost after applying the mapping procedure. In practice, we chose to iteratively adjust by hand the covariance $ C(\textbf{h}) $ of the independent Gaussian vector components $ X_1(\textbf{u}),\dots, X_p(\textbf{u})$ to match the experimental covariance of the input raw observations.

\section{Conclusions}
\label{Conclusions} 

The concept of dRFs is introduced in this research, integrating it into a methodology designed for mineral potential mapping. The main purpose of the presented methodology is to combine various mineral response models, calibrated in areas with extensive exploration data. The information contained within these models is then transferred into less studied regions by the inference of a new, best-estimated model in the geostatistical sense. Key to this approach is the notion of dRFs, as it enables the examination of the spatial relationships among models, quantifying their spatial correlation, and later allowing their spatial simulation through a procedure analogous to classical Gaussian simulation of RFs. In this fashion, we are able to generate multiple scenarios that honor both the conditioning response models on space and the probability distribution of their parameters, all while avoiding any data augmentation strategies. The outcomes in the case study have proven to be promising, unlocking a significant amount of resources within a small portion of the exploration area. Furthermore, in situations where there is a lack of exhaustive knowledge regarding input features, the approach provides insights into the significance of input features within a geographic context, thus benefiting future exploration endeavors.

\section{Acknowledgements}
The author acknowledges the joint support provided by CSIRO, Australia, and by the Mineral Exploration Cooperative Research Centre (MinEx CRC), whose activities are funded by the Australian Government's Cooperative Research Centre Programme. %

\section{Conflict of Interest}
The author declares that no conflict of interest could influence the work reported in this paper.

\small
\bibliographystyle{spbasic}      %
\bibliography{SVMRF}
\end{document}